\documentclass[dissertation, copyright, approved, final]{uothesis}
\usepackage[USenglish,UKenglish]{babel}
\usepackage{mathtools,amsmath,amsfonts,amsthm,amssymb,bm, bbm}
\usepackage{dcolumn,booktabs,longtable}
\usepackage[singlelinecheck=false]{caption}
\usepackage{placeins}
\usepackage{graphicx}
\usepackage[space]{grffile} 
\usepackage{rotating}
\usepackage{enumitem}
\usepackage{epstopdf}
\usepackage{ctable}
\usepackage{setspace}
\usepackage[normalem]{ulem}
\usepackage{tabularx,xspace,multirow}
\usepackage{array}

\usepackage[utf8]{inputenc}
\usepackage{hyperref}
\usepackage{etoolbox}

\newcommand{\zerodisplayskips}{%
  \setlength{\abovedisplayskip}{3pt}
  \setlength{\belowdisplayskip}{3pt}
  \setlength{\abovedisplayshortskip}{3pt}
  \setlength{\belowdisplayshortskip}{3pt}}
\appto{\normalsize}{\zerodisplayskips}
\appto{\small}{\zerodisplayskips}
\appto{\footnotesize}{\zerodisplayskips}

\usepackage[utf8]{inputenc}
\usepackage[square,numbers]{natbib}

\usepackage{Includes/imports}  
\usepackage{Includes/definitions} 
\usepackage{Includes/commands} 

\begin{document}
\IfFileExists{Preface/preface.tex}{%
\covertitle{\DissertationTitle}
\abstracttitle{Smaller, Faster, Cheaper: Architectural Designs for Efficient
Machine Learning}
\author{Steven Walton}
\major{Computer Science}
\department{Department of Computer Science} 
\narrowdepartment{Department of Computer Science}
\degreetype{Doctor of Philosophy}
\degreemonth{Summer}
\degreeyear{2025}
\chair{Hank Childs}
\committee{Humphrey Shi, Core Member \\*
           Daniel Lowd, Core Member \\*
           Thien Nguyen, Core Member \\*
           Edward Rubin, Institutional Representative \\*}

\abstract{%
    Major advancements in the capabilities of computer vision models have been
primarily fueled by rapid expansion of datasets, model parameters, and
computational budgets, leading to ever-increasing demands on computational
infrastructure.
However, as these models are deployed in increasingly diverse and
resource-constrained environments, there is a pressing need for architectures
that can deliver high performance while requiring fewer computational resources.

This dissertation focuses on architectural principles through which
models can achieve increased performance while reducing their computational
demands.
We discuss strides towards this goal through three directions.
First, we focus on data ingress and egress, investigating how information may be
passed into and retrieved from our core neural processing units.
This ensures that our models make the most of available data, allowing
smaller architectures to become more performant.
Second, we investigate modifications to the core neural architecture, applied to
restricted attention in vision transformers.
This section explores how removing uniform context windows in
restricted attention increases the expressivity of the underlying neural
architecture.
Third, we explore the natural structures of Normalizing Flows and how we can
leverage these properties to better distill model knowledge.

These contributions demonstrate that careful design of neural architectures can
increase the efficiency of machine learning algorithms, allowing them to become
smaller, faster, and cheaper.


    This dissertation includes previously published and unpublished co-authored
    material.
}

\school{University of Oregon, Eugene, OR, USA}
\school{Embry-Riddle Aeronautical University, Prescott, AZ, USA}

\degree{Doctor of Philosophy in Computer Science, 2025, University of Oregon}
\degree{Master of Science in Computer Science, 2023, University of Oregon}
\degree{Bachelor of Science in Space Physics, 2014, Embry-Riddle Aeronautical University}

\interests{Computer Vision}
\interests{Machine Learning}
\interests{Artificial Intelligence}
\interests{Generative Modeling}
\position{Graduate Researcher, University of Oregon, Eugene, OR, Aug. 2018 - Jun. 2025}
\position{Metropolis Intern, Nvidia, Sep. 2023 - Mar. 2024}
\position{Ph.D. Research Intern, Picsart AI Research, Eugene, OR, Jun. 2021 - Nov. 2022}
\position{Computer Science Intern, Lawrence Livermore National Labratory, Livermore, CA, Jun. - Sept. 2020}
\position{Computer Science Intern, Lawrence Livermore National Labratory, Livermore, CA, Jun. - Sept. 2019}
\position{ASTRO Intern, Oak Ridge National Labratory, Oak Ridge, TN, Jun. - Aug. 2018}
\position{}
\award{Outstanding Reviewer, CVPR 2025} 

\publication{Steven Walton, Ali Hassani, Xingqian Xu, Zhangyang Wang, and Humphrey Shi.
Efficient image generation with variadic attention heads. In Proceedings of the
IEEE/CVF Conference on Computer Vision and Pattern Recognition (CVPR)
Workshops, 2025}
\publication{Steven Walton, Valeriy Klyukin, Maksim Artemev, Denis Derkach, Nikita Orlov,
and Humphrey Shi. Distilling normalizing flows. In Proceedings of the IEEE/CVF
Conference on Computer Vision and Pattern Recognition (CVPR) Workshops,
2025.}
\publication{Ali Hassani, Fengzhe Zhou, Aditya Kane, Jiannan Huang, Chieh-Yun Chen, Min
Shi, Steven Walton, Markus Hoehnerbach, Vijay Thakkar, Michael Isaev, Qinsheng
Zhang, Bing Xu, Haicheng Wu, Wen mei Hwu, Ming-Yu Liu, and Humphrey Shi. Generalized neighborhood attention: Multi-dimensional sparse attention at the
speed of light, 2025. arXiv:2504.16922}
\publication{Jonathan Roberts, Mohammad Reza Taesiri, Ansh Sharma, Akash Gupta, Samuel
Roberts, Ioana Croitoru, Simion-Vlad Bogolin, Jialu Tang, Florian Langer, Vyas
Raina, Vatsal Raina, Hanyi Xiong, Vishaal Udandarao, Jingyi Lu, Shiyang Chen,
Sam Purkis, Tianshuo Yan, Wenye Lin, Gyungin Shin, Qiaochu Yang, Anh Totti
Nguyen, David I. Atkinson, Aaditya Baranwal, Alexandru Coca, Mikah Dang, Sebastian Dziadzio, Jakob D. Kunz, Kaiqu Liang, Alexander Lo, Brian Pulfer,
Steven Walton, Charig Yang, Kai Han, and Samuel Albanie. Zerobench: An
impossible visual benchmark for contemporary large multimodal models, 2025. arXiv:2502.09696}
\publication{Noble Kennamer, Steven Walton, and Alexander Ihler. Design amortization for
bayesian optimal experimental design. Proceedings of the AAAI Conference on
Artificial Intelligence, 37(7):8220–8227, 2023.}
\publication{Ali Hassani, Steven Walton, Jiachen Li, Shen Li, and Humphrey Shi. Neighborhood
attention transformer. In Proceedings of the IEEE/CVF Conference on Computer
Vision and Pattern Recognition (CVPR), pages 6185–6194, 2023.}
\publication{Steven Walton. Isomorphism, normalizing flows, and density estimation: Preserving
relationships between data, 2022. https://www.cs.uoregon.edu/Reports/AREA-202307-Walton.pdf}
\publication{Jitesh Jain, Anukriti Singh, Nikita Orlov, Zilong Huang, Jiachen Li, Steven Walton,
and Humphrey Shi. Semask: Semantically masked transformers for semantic
segmentation. In Proceedings of the IEEE/CVF International Conference on
Computer Vision (ICCV) Workshops, pages 752–761, 2023.}
\publication{Jiachen Li, Ali Hassani, Steven Walton, and Humphrey Shi. Convmlp: Hierarchical
convolutional mlps for vision. In Proceedings of the IEEE/CVF Conference on
Computer Vision and Pattern Recognition (CVPR) Workshops, pages 6307–6316,
2023.}
\publication{David Pugmire, James Kress, Jieyang Chen, Hank Childs, Jong Choi, Dmitry
Ganyushin, Berk Geveci, Mark Kim, Scott Klasky, Xin Liang, Jeremy Logan, Nicole
Marsaglia, Kshitij Mehta, Norbert Podhorszki, Caitlin Ross, Eric Suchyta, Nick
Thompson, Steven Walton, Lipeng Wan, Matthew Wolf, Jeffrey Nichols, Becky
Verastegui, Arthur ‘Barney’ Maccabe, Oscar Hernandez, Suzanne Parete-Koon,
and Theresa Ahearn. ``Visualization as a Service for Scientific Data''. In ``Driving
Scientific and Engineering Discoveries Through the Convergence of HPC, Big Data
and AI'', pages ``157–174'', ``Cham'', ``2020''. ``Springer International
Publishing''.}
\publication{Steven Walton, Ali Hassani, Abulikemu Abuduweili, and Humphrey Shi.
Training compact transformers from scratch in 30 minutes with pytorch.
medium.com/pytorch, 2021. arXiv:2104.05704}
\publication{Ali Hassani, Steven Walton, Nikhil Shah, Abulikemu Abuduweili, Jiachen Li, and
Humphrey Shi. Escaping the big data paradigm with compact transformers, 2022.}
\publication{Steven Walton. Datum: Dotted attention temporal upscaling method.
2020. https://www.cs.uoregon.edu/Reports/DRP-202006-Walton.pdf}
\publication{}

\acknowledge{
    I'd like to thank my mentors and professors from my Universities for helping
    get me to where I am today.
    Thank you Jeff Spear, for being the first to show me how to be creative with
    math.
    To Karla Westphal, for helping me find passion and dedication to the
    subject.
    To my undergraduate professors: Timothy Callahan, Andri Gretarsson, Edward
    Poon, Hisaya Tsutsui, and Darrel Smith who taught me my passion for math,
    physics, and providing me the tools to understand the world around me.
    To my graduate professors and advisors, who helped get me through these
    difficult times.
    I especially want to thank Hank Childs for encouraging me to pursue Machine
    Learning and to be my acting advisor after Humphrey moved to Georgia Tech.
    I want to thank Humphrey Shi for being my advisor and helping me make all
    the connections and pushing me to become a better researcher.

    I'd like to thank my friends and family for helping get through this.
    It was a journey that I could not have made alone.
    Noble, 
    you've been a close friend for so many years and your insights helped shape
    my research and encouraged me to go to graduate school.
    You constantly challenge my ideas, often frustratingly so,
    but they always end up better and more refined for it.
    Never change.
    Ali, I couldn't ask for a better co-author nor friend.
    Your intelligence and work ethic have always pushed me to better myself, and
    I look forward to calling you ``doctor''.
    I want to thank my cat Hypatia, who has been my best friend for the last
    decade.
    She's had to listen to many explinations and I'm sorry you have not received
    formal recognition for your contributions despite frequent appearances in my works (including this one).
    Lastly, I want to thank my wonderful girlfriend: Jaichung Lee.
    We have been through so much and I could not have crossed the finish line
    without you.
    I know it was as much of a challenge for you as it was for me, and this PhD
    would not have been possible without your many efforts. Thank you.
}

\dedication{
    \centering
    My mom, and the many years of watching Star Trek together.

    My dad, and the many years of reading Asimov together.

    Jaichung, and the many years to build the future together.
}

    }{
}
\maketitle

\IfFileExists{Chapters/chapter_magic.tex}{%
    \chapter{Introduction}\label{ch:introduction}
\chapterquote{
    I don't believe in empirical science. I only believe in a priori truth.
}{Kurt G\"{o}del}

\section{Motivation}

This thesis focuses on the development of efficiently training machine learning 
algorithms, primarily applied to Computer Vision.
Our focus is to develop methods which allow for a reduction in computational
resources required to train and deploy models.

Machine Learning is a subfield of Artificial Intelligence which aims to process
data and automate the discovery of structures within the data.
This process reduces the burden of needing to derive explicit formulations,
instead allowing automation through optimization.
This process allows algorithms to ``learn'' by ``training'' on the
data.

Computer Vision applies to a wide range of problems related to perception.
Traditionally associated with image and video processing, the field extends to
processing of other data, such as LIDAR, radio, depth
estimation, and other forms of signal processing.
The domain involves a broad range of tasks, including: regression, which models
quantitative relationships between variables; discrimination, the processing
distinguishing relevant objects or patterns; and generation, or data synthesis.
The primary focus of this thesis revolves around discrimination and generation
of images.

Image processing presents unique challenges, often due to the high
dimensionality of the embeddings.
This high dimensionality causes difficulties in formulating explicit
descriptions of our data and the underlying structures within it.
The goal of computer vision is to create the machinery necessary to automate
this process for us, as efficiently as possible.
While we may not be able to create fully formulate descriptions, the
descriptions we provide our algorithms can both help and hinder them.
For example, images usually have spatial relationships, with pixels that are
local spatially having high probabilities of being related to one another.
This has led to the use Convolutional Neural Networks, as their architecture is
able to exploit this natural bias.
But such relationships may not always hold.
For example, a QR code contains sharp transitions, where neighboring pixels do 
not aid the prediction of one another.
More flexible architectures, such as attention, can better process such imagery
by reducing the importance of locality.
Therefore, to efficiently process data we must consider the biases
implicit to the neural architectures that we use.

The modern success of these algorithms has presented additional challenges.
It has been found that many of these methods can be improved through simple
means: making them larger and providing them with more training
data~\cite{suttonbitter}.
While this has led to dramatic improvements, it has similarly led to dramatic
increases in the computational resources necessary to train and deploy these
models.
Once trained, these models may still be quite difficult to deploy, with their
high computational demands, greatly limiting where they can be used.
This has led many researchers to consider how these models can be more
efficiently trained, requiring: less data, less time to train, and fewer
computational resources.
Similar challenges exist with respect to the deployment of these models.

\section{Research Goals and Approaches}

The focus of this thesis revolves around two primary questions:
\begin{itemize}
    \item How do we reduce the model's data dependence?
    \item How do we reduce the model's computational demands?
\end{itemize}
These questions are fundamentally intertwined, necessitating solutions which
address the problems simultaneously.
Naturally, by reducing the amount of data that a model must ingest reduces the
amount of time that a model must be trained for.
Conversely, by making a model more efficiently extract information from its
data, the less data it will need to achieve a given performance level.
This is because model parameters do not just determine its information capacity, 
but also play an integral role in the solution space during
training~\cite{NEURIPS2023_adc98a26}.
Many works have found that once trained models are often significantly 
over-parametrized, meaning only a subset of their parameters are being used to
model the data~\cite{frankle2018the,lee2018snip}.
These findings are further evidenced by the continued increasing performance of 
smaller models~\cite{hooker2024limitationscomputethresholdsgovernance}, and 
strongly suggest our models can be trained more efficiently.

Our motivation to reduce a model's data depends exists beyond our desire to be
cost effective.
Real world large datasets provides two primary challenges which require our
models to be data efficient.
First, many important structures within the data are subtle and difficult to
recover.
Second, data is often \emph{heavy-tailed}, meaning we do not have many samples.
Fundamentally, these require our models to generalize relationships with minimal
examples.
While we may focus on explicitly constrained data to aid the interpretation of
our work, it provides benefits as our models and data expand in size.

These feats are primarily accomplished the development the development of 
neural architectures and optimization methods.
This thesis focuses on the former, specifically, studying the design of 
Computer Vision architectures which reduce: parameters, data dependence, and 
system resources.
These goals must be simultaneously optimized. 
Our objective is not to develop models with a small number of parameters if 
they also require substantially greater costs during training or deployment.
Similarly, this would undermine our own goals if we reduce a model's data
dependence with significant cost to its performance.

This thesis investigates three critical aspects of our neural
architectures and structure it to follow a natural progression in complexity.
The first work focuses on the understanding how our core neural architecture 
takes in data and how to efficiently extract the relationships it uncovers.
Without efficiently providing and extracting data to/from a model, they become
wasteful and this hinders the ability to develop more efficient core 
architectures. 
The second work focus on the core architecture, which perform the majority of
the data processing.
This section studies these two aspects as applied to vision transformers, 
directly building off one another.
The third work revolves around knowledge distillation of Normalizing Flows.
These models are structurally aware, explicitly designed to preserve the
structures within the data.
From these three lenses this thesis seeks to better understand how to build
neural architectures that are smaller, faster, and cheaper.

\section{Dissertation Outline}\label{sec:intro-outline}

This dissertation is organized as follows:

\Cref{ch:background} provides the necessary background and foundational
information necessary to understand the research objectives.
This background is necessary for understanding how the works are connected and
the ways we seek to resolve underlying issues.

\Cref{ch:escaping} presents the work \emph{Escaping the Big Data Paradigm with
Compact Transformers}~\cite{hassani2022escapingbigdataparadigm}, and focuses on 
efficiently embedding and extracting data from Vision Transformers.

\Cref{ch:stylenat} presents the work \emph{Efficient Image Generation with
Variadic Attention Heads}~\cite{WaltonStyleNAT2025CVPR}, as well as the
works it builds upon: 
\emph{Neighborhood Attention Transformer}~\cite{Hassani_2023_CVPR}.

\Cref{ch:flows} presents the work \emph{Distilling Normalizing Flows}, which
provides a framework for knowledge distillation with Normalizing Flow
architectures and studies the categorical distillation methods.

\Cref{ch:conclusion} provides an overview of the findings and recommendations
for future work.

\section{Co-Authored Material}\label{sec:intro-coauth}
The research presented herein involves previously published material.
Below is a listing of the prior works in relation to the chapter material.
Details of division of labor is provided in the preface to each chapter.

\begin{itemize}
    \item \Cref{ch:background}: This \lcnamecref{ch:background} includes material that was part of Steven Walton's Area Exam~\cite{walton2023isomorphism}.
    \item \Cref{ch:escaping}: This work was contains materials from
        \emph{Escaping the Big Data Paradigm with Compact
        Transformers}~\cite{hassani2022escapingbigdataparadigm}. This work was a
        collaboration with Ali Hassani, Nikhil Shah, Abulikemu Abuduweili, 
        Jiachen Li, Humphrey Shi, and myself.
    \item \Cref{ch:stylenat}: This \lcnamecref{ch:stylenat} contains materials from both
        \emph{Neighborhood Attention
        Transformer}~\cite{Hassani_2023_CVPR} and \emph{Efficient Image
        Generation with Variadic Attention Heads}~\cite{WaltonStyleNAT2025CVPR},
        with focus around the latter. The former is a collaboration between Ali
        Hassani, Jaichen Li, Shen Li, Humphrey Shi, and myself. The latter was a
        collaboration between Ali Hassani, Xingqian Xu, Zhangyang Wang, Humphrey
        Shi, and myself.
    \item \Cref{ch:flows}: This \lcnamecref{ch:flows} contains material from a collaboration
        between Valeriy Klyukin, Maksim Artemev, Denis Derkach, Nikita Orlov,
        Humphrey Shi, and myself.
\end{itemize}

\chapter{Background}\label{ch:background}
\chapterquote{%
    Mathematicians do not deal in objects, but in the relationships among objects.
}{Henri Poincar\'{e}}
\textbf{Nota Bene:}
\emph{Some of the text and figures from this section were part of Steven
Walton's Area Exam~\cite{walton2023isomorphism}, which has been publicly
released by The University of Oregon. Steven was the sole author of
this work.
}

This section covers the background necessary for understanding the
motivation and purpose of the work performed.
There is includes some necessary discussion about how machine learning
algorithms work, how data is processed, and the inherent biases of different
learning architectures.
The latter of which is the main focus of this thesis.
While subsequent chapters will have lower mathematical notation and formulation,
those herein provide important context and intuition for the work ahead.
To reach our goal of making our machine learning models smaller, faster, and 
cheaper, we need to have some core understandings as to how these models work.
It is not enough to treat them as black boxes; rather we have to look inside.
Much of machine learning terminology has not been standardized, thus this
section may be used to contextualize these terminologies and the usage within
this thesis.

\section{Learned Data Mappings}\label{ch:bg-data-maps}

The procedure can be understood through mapping between two sets, where our
neural network is a learned mapping, $f(x)$.
Deep neural networks are \emph{Universal
Approximators}~\cite{csaji2001approximation,hornik1989multilayer,liu2024kan},
where every multivariate continuous function can, in principle, be approximated 
by the superposition of a sequence of continuous functions.

With this in mind, it helps to revisit some of the basics of functions and set
theory.
We can view the \emph{Domain}, $D$, as all valid inputs to 
the neural network.
In the study of Computer Vision this is \emph{any valid image}, regardless of
whether this image is meaningful to humans or not.
We can then define our \emph{Range}, $R$, as all possible outputs that our
neural net can produce.
In the case of Image Classification this would be all labels that we are trying
to learn.
Our \emph{Codomain}, $C$, is a super-set to our Range, $R \subseteq C$, and may
include elements that our map cannot reach.
In our example of Image Classification our Codomain would represent 
\emph{all} possible labels.

\begin{table}
    \begin{center}
        \begin{tabular}{|l|c|c|}
            \hline
            \textbf{Name} & \textbf{Relation} & \textbf{Set}\\
            \hline
            Domain & $D$ & $\{\forall\; x \in D\}$ \\
            Codomain & $C$ & $\{\forall\; y \in C\}$ \\
            Range & $\widetilde{C}\subseteq C$ & $\{y\in C\;|\; \exists\; x\in D :
            f(x) = y\}$\\
            Image & $\widetilde{C}\subseteq C$ & $\{y \in C\;|\; \exists\;
            x\in\widetilde{D} : f(x) = y\}$ \\
            Preimage & $\widetilde{D}\subseteq D$ & $\{x \in D \;|\; \exists\;
            y\in\widetilde{R} : f(x) = y\}$\\
            \hline
        \end{tabular}
        \caption[Terminology of Mathematical Sets]{%
            Explanation of important set terms denoting their relationships and what
            elements are in their set.
        }\label{tab:bg-sets}
    \end{center}
\end{table}

\begin{figure}
    \centering
    \includegraphics[width=\textwidth]{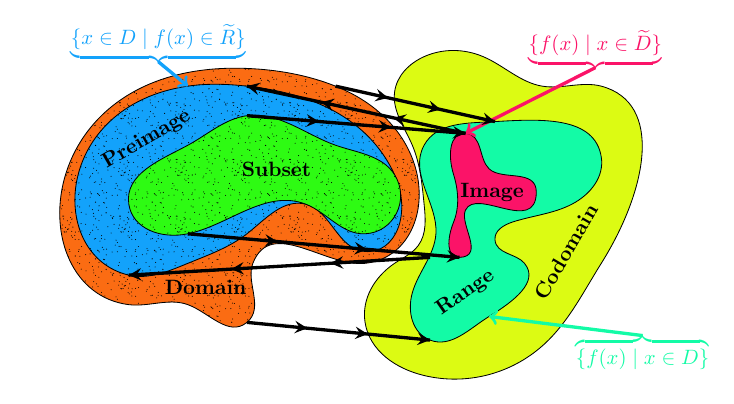}
    \caption[Domain, Range, Image, Preimage diagram]{The diagram illustrating concepts from Set Theory, explaining the
    \emph{Domain} ($D$), \emph{Codomain}, \emph{Range} ($R$), \emph{Image}
    ($\widetilde{R}$), and \emph{Preimage}.
    }
    \label{fig:bg-manifold}
\end{figure}

In practice, we are likely only interested in studying some subset of our
domain, $\widetilde{D}\subseteq D$.
This subset can be arbitrary and may be something like our training set, the set
of images interesting to humans, or even some subset of our training data.
Regardless of what this subset is, when they are passed through our mapping
function then we call the outputs an \emph{Image}, $\widetilde{R}\subseteq R$.
A ``reverse'' of this function may then be defined, called the \emph{Preimage},
$f^*[\widetilde{R}]$.
The Preimage is defined as the set of elements in the domain that map to some
image in the codomain.
It is important to note that the Preimage is not the \emph{inverse} of the image.
Many texts use the notation $f^{-1}$, but we will use $f^*$ to avoid
confusion.\footnote{This notation is used for a pullback, which is a nearly 
identical concept.}
\Cref{tab:bg-sets} and \Cref{fig:bg-manifold} are included to help explain these
concepts. 

We will define a \emph{Target}, \Target, as the set of data we \emph{intend} to 
model.
Unfortunately, this distribution may be unobtainable and is often intractable.
That is, we are unable to provide a formal description of the distribution.
An example, which we will use in \Cref{ch:stylenat} and \Cref{ch:flows}, is 
``the set of all possible human faces.''
We do not have a proper mathematical description this set, making it 
\emph{intractable}, nor is it possible for us to completely sample from this 
set as it would require
infinite time\footnote{This set would include all faces that were and all faces
that will be.}.
We instead collect a set of sample data $\Samplespace\subseteq\Target$,
 which may be used to train the model (e.g. FFHQ~\cite{Karras_2019_CVPR} or
 CelebA~\cite{liu2015faceattributes}).
It is important to note that we may not know how well \Samplespace approximates
\Target, especially when \Target is intractable.
Our model processes data from the \Samplespace to generate output, \Output.
When performing Classification/Discrimination tasks, our output may be a (or a
list of) label(s) but in generative tasks we instead seek to approximate the
target distribution, \Learned{\Target}.
We should keep this model in mind when evaluating our work, so we can best
understand what our models can and cannot do.
Our data are discrete and sampled from the distributions we are trying to
approximate, and great care must be taken to determine what is in our
distribution or not.

\section{Scale Is Not All You Need}\label{sec:bg-SINAYN}

In March of 2019 Richard Sutton wrote a short article titled The
Bitter Lesson~\cite{suttonbitter}.
This article had a large impact on the machine learning community.
Sutton makes the argument that methods based predominantly on leveraging human
knowledge are ill-founded and that our historical progress has shown us that
focusing on search has resulted in success.
Sutton acknowledges the benefits of leveraging human knowledge as well as how in
practice this can often be constraining, preventing our machines from leveraging
more general computation.
Either through misinterpretation by Sutton or through readers, a popular belief
rose through the community: ``\emph{Scale Is All You Need}''.
This notion need be addressed, for if the belief is true to face then the only
work need be done is that of scaling compute and data gathering.
Some will interpret this in that scaling is sufficient, and that there may be
more efficient methods, but we will show that scaling alone is insufficient.
We do not disagree that scale is a necessary and essential component, but that
it alone is insufficient to both explain recent progress as well as provide
direction for further advancement.
These claims let critical conditions remain implicit, assuming shared
assumptions among readers.
These subtle details are consequential to generating efficient machine learning
models, as understanding what data increases performance allows us to also
better design algorithms to maximally incorporate information.

Two aspects of scaling must be addressed: that of scaling data
(\Cref{sec:bg-scale-data}) and 
that of scaling compute (\Cref{sec:bg-size}).

\subsection{Scaling Data}\label{sec:bg-scale-data}
Undeniably one of the reasons for major advances has resulted with scaling of
data.
There is a simple argument that may suggest scaling data will be sufficient.
We need to look at this to understand where it works and doesn't.

Our goal in machine learning is to learn some distribution, which we will call
our \emph{Target Distribution}, \Target.
If we uniformly and randomly sample from our target distribution, one can 
conclude that with scale we will also increase our covering of the distribution.
We may view this another way: if we select some arbitrary point in our target
distribution, as we continue to sample then the distance between it and some
data point, $d_i$, in our set of sampled points will decrease.
$ \exists\; \varepsilon\in\mathbb{R} \text{ s.t. } ||d_i - d_j||_p^p <
\varepsilon\; |\;\forall d_i,d_j\in \mathcal{T} $.
Where $|| \cdot ||_p^p$
represents an arbitrary $L_p$ distance.

We can refine this more generally, which will better help us as we increase
complexity. 
We can partition our distribution \Target into disjoint continuous
partitions $\{P_0,\ldots,P_n\}$.
That is: $ P_i\cap P_j = \{\emptyset\}\; |\; \forall i\neq j$ and $\bigcup
P_i = \mathcal{T}$.
We can reach a similar natural conclusion: as the number of samples increases,
the probability that there does not exist a sample belonging to partition $P_i$
goes to zero.
$\lim_{n\rightarrow\inf} Pr(s\in P_i) = 0$.


This generalization helps us in two ways.
Our partitions can be of arbitrary size and shape, allowing us to use them as
abstractions, such as semantic representations.\footnote{We still need to maintain
care to ensure our semantic representations are disjoint. This does not allow us
to pick arbitrary semantic representations.}
Where a semantic representation may represent categories of our data.
For example, if our model is generating human faces we may consider hair color
as a semantic representation.
This formulation can also be repeated for each partition, which allows us to extend
the notion to a more realistic setting where data is discrete (i.e.
discretization).

While this logic may be natural, it relies on assumptions that are not true in
practice.
Notably, it assumes that both the data is independent and identically
distributed (\emph{i.i.d}) and that our sampling process is unbiased.
These assumptions are not representative of the real world data, nor of the way
in which we sample.
In practice, as we increase the number of samples we increase the diversity of
our data.
This diversity, or variance, in data has a large impact on our models' ability
to generalize.
We will see in \Cref{ch:escaping} that introducing data augmentation to our
models results in a significant improvement in their performance.
These augmentations create additional variance in the data and help the model to
not overfit.

Scaling of data in the way we typically gather data can grow the variance to a
greater degree than our typical data augmentation methods can.
But this represents a fundamental limitation as well.
We cannot scale infinitely, and as we gather more data inevitably we turn from
increasing variance to contracting the variance.
There are only so many unique things in the world.
To understand this, we may think about randomly throwing a dart at a dartboard.
As we start, every new dart likely lands with a high distance from one another.
But as we continue we increase our coverage over the dartboard and our new darts
land close to an existing dart.
This variance contraction means that we cannot rely on scaling data
indefinitely.

Additionally, an extra challenge comes from scaling data.
Once the data is so large, we are unable to properly investigate it.
This means we will not be able to properly verify that our model is not trained
on the data it is being tested on.
In this manner, we want to use the minimum amount of data required to train our
models, to reduce our burden of verification.

In practice, our data is heavy tailed, with many samples being underrepresented.
Ultimately, despite high amounts of data, subsets exists in a low data
regime.
Our models may benefit from shared similarities, via a superposition of
representations, but we are still motivated to develop models which work better
when data is sparse.
By better understanding how to make our models efficiently learn in limited data
regimes we hope to build techniques that allow our larger models to efficiently
model data that is within the long tail.

\subsection{Model Size}\label{sec:bg-size}
We face similar complexities when it comes to the scaling of our models.
Inherently our model parameters change our loss
landscape~\cite{NEURIPS2018_a41b3bb3}, with larger models
providing more ways for data to be 
disentangled~\cite{lee2003smooth,grenander_pattern_2006,dupont_augmented_2019}.
It can be shown that different by using different loss functions that we may
even trick ourselves into believing our models have found emergent
capabilities~\cite{wei2022emergent} when they may have
not~\cite{NEURIPS2023_adc98a26}.

With increased model parameters our models are more likely to overfit our data,
making it difficult to generalize.
With such sizes in terms of data and parameters it becomes difficult to
distinguish between our models memorizing the data vs modeling the data.
In practice, we benefit from physical limitations, which also puts pressure on
making our models as small as possible.
The larger our models are, the more expensive they are to run.

\section{The Foundations That Shape Us}\label{sec:bg-architectures}
To cost effectively train our models we want them to both be parameter efficient
and data efficient.
With too much data, we are may spend disproportionate times loading from disk
and simply ingesting the data.
With too many parameters, we must split, or shard, our model across large
supercomputing infrastructures.

Key to Sutton's Bitter Lesson was that models should be powerful and flexible.
With our trend in scaling, we have also seen tremendous improvements in
the algorithms that we use, such as the advent of the
transformer~\cite{vaswani2017attention}.
Scale cannot be enough to explain our progress, as we have found that as
research progresses, many smaller models end up significantly outperforming
larger models~\cite{hooker2024limitationscomputethresholdsgovernance}, and this
thesis is further demonstration of that.

These algorithms may be referred to as our neural architectures, as we build
them to work together.
In the following sections we introduce some of the key architectures that will
be used throughout this work.
There exist far more frameworks
methods~\cite{gu2024mambalineartimesequencemodeling,liu2024kan} and we focus
only on what are used herein.

\subsection{Transformers}\label{sec:bg-transformers}
The transformer model has become the backbone of modern machine learning models.
This is due to its high flexibility, being able to form a relationship between
all elements it attends to.
Unlike many other architectures, the transformer is not limited by the locality
of the data, with it being able to discover relationships between data
regardless of its position in a sequence.
This greater flexibility comes at an increased computational complexity, but
enables the model to form relationships that could not be efficiently formed
through other previous architectures.

These models are fairly simple in construction, having two main components:
attention~\cite{graves2014neuralturingmachines,luong2015effectiveapproachesattentionbasedneural,vaswani2017attention},
and a feed-forward layer.

\begin{figure}[htbp]
    \centering
    \includegraphics[width=0.25\linewidth]{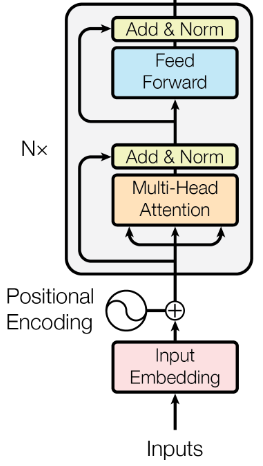}
    \caption[Vision Transformer Architecture]{%
        The Transformer model architecture from Vaswani et. al. 
        Diagram depicts dot-product self attention.%
    }\label{fig:bg-vas-attention}
\end{figure}

In \Cref{fig:bg-vas-attention} depicts part of the transformer model from
Vaswani~\etal's work, showing the dot-product self attention (DPSA) variant,
which is used throughout this work.
The figure depicts a ``post-norm'' configuration, with the normalization layers
appearing after the attnetion and feed-forward units, but modern 
configurations usually use ``pre-norm'' due to increased stability.
The core of the transformer model is attention, defined as:

\begin{equation}
    \text{Softmax}\left(\frac{QK^T}{\sqrt{d_k}}\right)V
\end{equation}

Where Q, K, and V represent queries, keys, and values, respectively.
These are learnable parameters, most usually parameterized by a single layer 
feed-forward network.
In the DPSA configuration, these networks share the same input.
$d_k$ in this equation is a softmax temperature scale, which is the inverse
square root of the embedding dimension (a user defined hyperparameter).
The queries and keys are multiplied together, learning a similarity matrix.
The softmax of this is then referred to as the ``score'', as its values are
defined by a probability distribution.
The value tensor is then weighted by the score, defining our attention function.

Commonly, this configuration is done in a ``multi-headed'' manner.
Instead of performing a single attention we may instead project our Q,K, and V 
tensors into an embedding so that we may process multiple attention 
calculations in parallel. 
The conclusion of the attention mechanism concatenates these tensors.
This tends to make our models more efficient as each head is independent and can
learn unique representations, as we will see in \Cref{ch:stylenat}.

The transformer model typically includes the usage of positional encoding, which
adds extra data to the model to indicate the position of tokens, or data, in a
sequence.

\begin{figure}
    \includegraphics[width=0.8\textwidth]{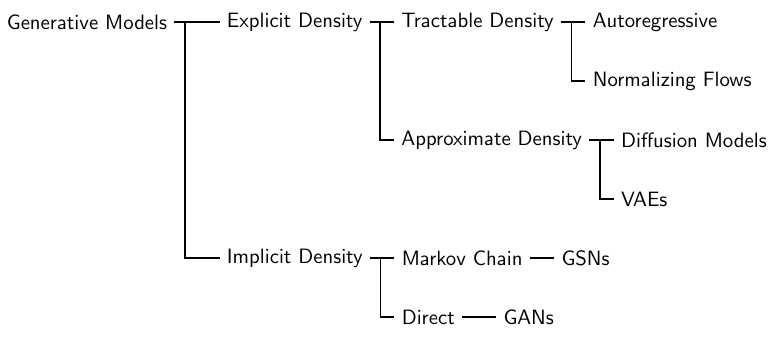}
    \caption[Taxonomy of Generative Models]{%
    Taxonomy of Generative Models, based on Goodfellow's
    Taxonomy~\cite{goodfellow_deep_2016}%
    }\label{fig:goodfellow_taxonomy}
\end{figure}

\subsection{Adversarial Generation}\label{sec:bg-gan}
Generative Adversarial Networks are a form of generative models introduced by
Goodfellow~\etal\cite{NIPS2014_f033ed80} which first enabled the generation of
high quality synthetic imagery.
Not necessarily restricted to image synthesis, these models enable unsupervised
learning by simultaneously training two models at once.
If our goal is to train an image generator, we both a model to generate images
and a model to discriminate real and fake images.
The discriminator model requires labeled data, but only the binary distinction
of real data or synthesized data.
These models then competitively train, being able to play a minimax game, which
often leads to high quality generation.
\begin{equation}
    \min_{G}\,\max_{D}\;
    \mathbb{E}_{x\sim p_{\text{data}}}\!\bigl[\log D(x)\bigr]
    +
    \mathbb{E}_{z\sim p_{z}}\!\bigl[\log\!\bigl(1-D\bigl(G(z)\bigr)\bigr)\bigr].
\end{equation}

$G$ learns a differentiable map $z\!\mapsto\!x$ that pushes forward a
simple prior (usually spherical Gaussian) toward the data manifold,
while $D$ learns to spot discrepancies.
While these models have shown great success and pushed the bounds of what is
possible, they are not without problems.
Training is notoriously unstable—mode collapse, vanishing gradients, and 
catastrophic forgetting are common.

In addition, many generative models have greatly increased in size.
These size increases have resulted in more impressive images but also become
harder to train, costlier to train, and become slower in throughput.
There then must be a trade-off of capabilities and performance, depending on the
applications.
In \Cref{ch:stylenat} we will use a GAN to demonstrate an improved variant of an 
attention mechanism, improving throughput and quality while decreasing the total
number of parameters.

\subsection{Normalizing Flows}\label{sec:bg-nflows}

Normalising flows provide \emph{exact} log-likelihoods by composing a
sequence of bijective, differentiable transforms
$f=f_{1}\cdot\dots\cdot f_{k}$:
\begin{equation}\label{eq:cov}
    p_x(\mathbf{x}) = p_u(\mathbf{u})\left|\text{det}\;J_f(\mathbf{u})\right|^{-1}
\end{equation}

Here $p_{u}$ is a tractable base distribution and $\text{det}\;J_f$ denotes the
Jacobian determinant.
The Jacobian determinant allows for a change of variable, allowing data from
one distribution ($u\in U$) to be expressed in another coordinate system ($x\in
X$).
A simplified example that many readers may be more familiar with is the change 
of coordinates from a Cartesian space into Polar coordinates
\begin{align}
    \begin{split}
        J &= \text{det}\frac{\partial(x,y)}{\partial(r,\theta)} \\
          &= \begin{bmatrix}
              \frac{\partial x}{\partial r} & \frac{\partial x}{\partial\theta}\\
              \frac{\partial y}{\partial r} & \frac{\partial y}{\partial\theta}
          \end{bmatrix}\\
          &= \begin{bmatrix}
              \cos\theta & -r\sin\theta\\
              \sin\theta & r\cos\theta
          \end{bmatrix} \\
          &= r\cos^2\theta + r\sin^2\theta\\
          &= r
    \end{split}
\end{align}
Given the Jacobian determinant it becomes trivial to convert from a Cartesian 
coordinate to Polar by the equation: 
$\iint f(x,y)dxdy = \iint f(r,\theta)rdrd\theta$

This idea extends greatly, with far more complex formulations of coordinate
transforms.
The importance of these transforms is that they generate an isomorphic mapping
from one space to another, where every element in one coordinate precisely maps
to a unique element in the other.
Through the composition of these transformations we can then define a nice
tractable distribution, such as a Gaussian, and learn a coordinate transform
that maps our data.
This, in effect, allows us to turn our intractable distribution into a tractable
one.
We should remain careful, as there are still some pitfalls and our distribution
is still only an approximation.

What makes this different from Approximate Density models, such as VAEs and
Diffusion models, is that those models do not generate isomorphic functions.
Like flows, they are able to generate a probability density function, making
them ``explicit'' (\Cref{fig:goodfellow_taxonomy}), but these models are by
nature lossy.
Where Flows are bijective, diffusion and VAEs are not.

\begin{figure*}[ht!]
    \input{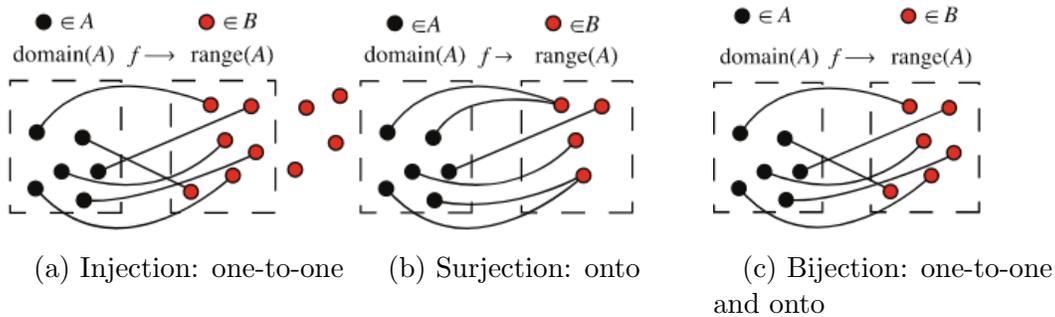}
    \caption[Diagram of Injection, Surjection, and Bijection]{%
        Visual representation of injections, surjections , and bijections. 
        Source: \emph{Wolfram  Mathworld}
    }\label{fig:functions}
\end{figure*}

The two most common forms of Normalizing Flows, which are also used within this
thesis, are:
\begin{description}
\item[Affine coupling flows.]:
      Partition input $x$ into two units, $(x_{0},x_{1})$, such that
        $f(x_{0},x_{1})=(x_{0},x_{1}\odot e^{s(x_{0})+t(x_{1}}))$, which make 
      computationally inexpensive triangular Jacobians
      (e.g.\ RealNVP~\cite{dinh2017densityestimationusingreal}, 
        Glow~\cite{kingma_glow_2018}).

\item[Autoregressive flows.]:
      Parameterise each dimension conditioned on previous ones,
      yielding a composition of triangular Jacobians
        (MAF/IAF~\cite{papamakarios2017masked,kingma2017improvingvariationalinferenceinverse}).
\end{description}

Unlike transformer models, the architecture to Normalizing flows are highly
restrictive.
These restrictions come with the benefits of increased interpretability, but at
the cost of additional computation and less flexibility.
Where to make these trade-offs is difficult but it remains a challenge in
determining the capability of these models.
Unfortunately these models tend to be greatly under studied, with only a
handful of models having been trained with $>100$M parameters, which is fairly
small by modern standards.

\section{The Tyranny of Measurements}\label{sec:bg-measures}
As a final note, we must be ever vigilant of the metrics that we use.
Qualitative metrics are a critical part of the scientific method, evidencing our
hypotheses and theories.
Yet, metrics are only guides, proxying the things we wish to measure.
We must stress the importance of this distinction as it is necessary to properly
evaluate our models and interpret what they are doing.
Within this thesis several of our works face the challenges of interpreting our
metrics and the absence of them.
In \Cref{ch:stylenat,ch:flows} perform image synthesis tasks, where our models
create new data that is representative of what they trained on.
There are no metrics that properly convey what is a good image or not.

For example, a common metric is for measuring the capabilities of image models
is the Fr\'{e}chet Inception Distance (FID)~\cite{NIPS2017_8a1d6947}.
This metric was shown to correlate with human judgement of image quality, but
was developed when image quality was much worse.
For comparison, the paper that introduced FID demonstrated models with an FID
around $12.5$ on the CelebA dataset, while the current state of the art is
3.15~\cite{teng2023relaydiffusionunifyingdiffusion}.
These correlations are helped improve the state of art systems, but not being
perfectly aligned with an actual measurement of realism the discrepancies grow
as our models improve.

The rapid success of machine learning is double edged sword. 
Our approximations that helped us make our progress may no longer be sufficient.
With all metrics, we must constantly check their alignment, to ensure that we
are progressing in the directions we intend.
This is quite similar to the gradient decent process we use in machine learning,
where early on we may make large improvements with highly suboptimal steps
towards the optima.
Yet, as our model becomes better, we tend to make smaller steps to ensure we are
progressing in the right direction.

\chapter{Escaping the Big Data Paradigm}\label{ch:escaping}
\chapterquote{
     The first principle is that you must not fool yourself and you are the easiest person to fool.
}{Richard Feynman}
\textbf{Nota Bene:}
This chapter is based on the previously published co-authored work 
\emph{Escaping the Big Data Paradigm with Compact
Transformers}~\cite{hassani2022escapingbigdataparadigm} and the associated blog
post published through PyTorch's Medium page~\cite{walton2021Escaping}.
\begin{itemize}
    \item Ali Hassani and Steven Walton are joint primary authors of this work.
    Together they wrote the majority of the code, performed the majority of
    experiments and writing of the paper.
    The majority of code was written during pair-programming sessions between the
    two.
    \item Steven Walton worked a bit more on designing the experiments and 
        developing the theory, ensuring claims were thoroughly evidenced and 
        finding relevant literature.
    \item Ali Hassani worked a bit more on code and launching experiments, 
        increasing code quality and ensuring experiments were launched 
        effectively, maximizing machine utilization. 
    \item Nikhil Shah helped manage launching experiments and contributed to the
        paper writing.
    \item Abulikemu Abuduweili provided code and feedback for the NLP experiments.
    \item Humphrey Shi was the advisor, contributing overall 
        guidance on the research as well as funding for the work. Humphrey also
        contributed to the writing of the paper and ensuring research stayed on
        track.
\end{itemize}

Critical to any data analysis is the preparation of that data. The ways
in which we encode our data has significant impacts on the way that data
is processed. It is not sufficient to simply apply the right modeling tools to
the data, but one first needs to ensure that the data is properly processed. 
In machine learning systems, this processing is typically done by both man and 
machine. The ingress and egress of data is critical, and will influence what 
structures in the data can ultimately be recovered.

In this chapter we introduce the work \emph{Escaping the Big Data 
Paradigm with Compact Transformers}~\cite{hassani2022escapingbigdataparadigm}.
This work demonstrates that Vision Transformers do not need large amounts of
data to be performant, instead being able to be trained from scratch and be
effective in limited data regimes.
Our results run counter to conventional wisdom around scaling, demonstrating
that scale may \emph{decrease} performance, rather than increase.
On small datasets, like CIFAR-10, our small models are able to achieve
comparable performance to much larger ViT models that also have large
pretraining.
On medium datasets, like ImageNet, we are able to outperform ViTs of comparable
sizes, and achieve accuracies only slightly lower than large models with large
pretraining.

\begin{figure}[hbtp]
    \centering
    \includegraphics[width=\linewidth]{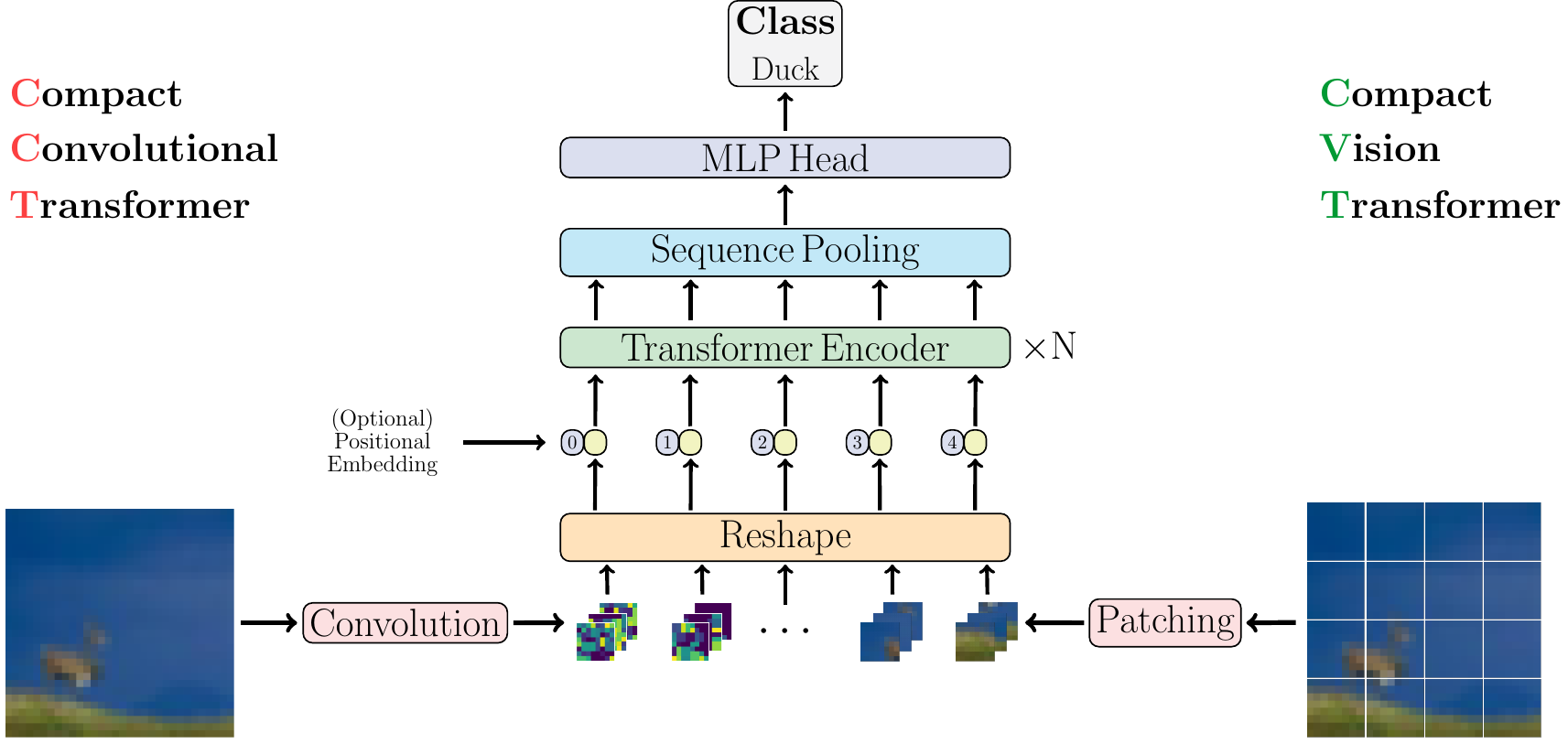}
    \caption{Architectural design of Compact Transformers}
    \label{fig:cct-sym}
\end{figure}

\section{Vision Transformers}\label{ch:escaping-vit}
With Vaswani~\etal's\cite{vaswani2017attention} demonstration of a dot-product
self-attention based transformer architectures in language, there were several 
attempts to integrate them into vision
models~\cite{Bello_2019_ICCV,NEURIPS2019_3416a75f,Hu_2018_CVPR,Hu_2019_ICCV}.
Cordonnier~\etal\cite{Cordonnier2020On} first showed that by downsampling and
adding a positional encoding layer, that a Bert~\cite{devlin2019bert} style
Transformer architecture could learn convolutional filters, given a sufficient
number of attention heads.
Unfortunately, these researchers were memory bound and were using $2\times2$
invertible down-sampling.
Dosovitskiy~\etal\cite{dosovitskiy2021an} improved upon this work, claiming
``An Image is Worth $16\times16$ Words'', introducing the Vision Transformer.
Instead of using a $2\times2$ down-sampling, they used larger $16\times16$
patches, giving the paper it's name.
Additionally, Dosovitskiy~\etal significantly increased scaled both data and
compute.
While Cordonnier~\etal's network was ${\approx}12$M parameters,
Dosovitskiy~\etal used 3 networks, 86M, 307M, and 632M.
While Cordonnier~\etal exclusively trained on CIFAR-10 and
CIFAR-100~\cite{4531741}, 
Dosovitskiy~\etal performed pretraining with the proprietary JFT-300M
dataset~\cite{Sun_2017_ICCV}, ImageNet-21k, and
ImageNet-1k~\cite{deng2009imagenet}.
Their work showed that with large-data pretraining that one could outperform
ResNet~\cite{He_2016_CVPR} trained models, although later work showed that by
training ResNets with modern training procedures that classification accuracy
becomes similar~\cite{wightman2021resnetstrikesbackimproved}.
Dosovitskiy~\etal performed a wide variety of experiments,
including using a CNN to generate their patch embedding and fine-tuning at
higher resolutions than
pretraining~\cite{NEURIPS2019_d03a857a,kolesnikov2020bigtransferbitgeneral}.
Their results suggested that only through large pretraining and large models could
ResNets be beat.

Dosovitskiy~\etal's work made an important claim: 
\textit{Transformers lack some of the inductive biases inherent to CNNs, such as
translational equivariance and locality, and therefore \uline{do not generalize well
when trained on insufficient amounts of data.} However, the picture changes if
the models are trained on larger datasets (14M-300M images). We find that large
scale training trumps inductive biases.}

If this problem could not be resolved then this would greatly limit research
contributions by labs without large compute infrastructures~\footnote{Often
called ``GPU Poor''}.
The community was quick to challenge Dosovitskiy~\etal's claim.

Touvron~\etal's 
\emph{Training Data-Efficient Image Transformers \& Distillation Through
Attention}~\cite{pmlr-v139-touvron21a}, quickly followed in an attempt to
address the claim, introducing the DeIT model.
In particular, they criticized the large pretraining and sought to counter the
claim that transformers do not generalize when trained on insufficient amounts
of data.
Their work similarly uses 3 models for training, but are a tiny (5M parameters),
small (22M parameters), and base (86M parameters).
The ViT was modified to introduce a knowledge transfer\footnote{We use
the phrasing \emph{knowledge transfer} instead of \emph{distillation} for
increased clarity; as the ``teacher'' network having fewer parameters than the ``student'' network}
token, and the training scheme was modified to include distillation from a 
pretrained convolutional based network.
For their convolutional network they selected a 
RegNetY-16GF~\cite{Radosavovic_2020_CVPR} network (84M parameters) as the 
default teacher network.

\section{Data Efficient Vision Transformers}\label{ch:escaping-cct}
While we recognize the importance of these works we believe alternative 
conclusions are possible.
The ViT results could be explained by several alternative hypotheses, including
the size of the network and through training techniques.
DeIT's results showed that part of the claim must be false, as even smaller 
models could achieve better performance, but this relied upon inheriting the
local inductive biases transferred by a CNN rather than learning them
themselves, which Cordonnier~\etal had demonstrated is possible.
The critical question remained: \emph{Can transformer models, be trained to 
outperform ResNets when model size and data were held equal?}
Both works suggested that the answer was no.
On the other hand, Transformers are universal approximators and
Cordonnier~\etal's work suggests there's no reason one should believe this
data threshold requirement.
Additionally, we believed ViT and DeIT were rejecting valuable information by
only passing a slice of the transformer's outputs to the classification
sub-network.

In an effort to resolve this, we proposed three hypothesis:
\begin{itemize}
    \item Non-overlapping image patches bias the transformer networks due to
        information loss at the boundaries.
    \item A learned transformation to map the transformer's outputs to the
        classification sub-network will improve performance.
    \item Transformer networks rely more on data variance than data quantity.
\end{itemize}

\subsection{Convolutional Tokenizer}\label{ch:escaping-tokenizer}
The first hypothesis was believed due to the discussion in the
background section (\Cref{sec:bg-scale-data}), where these models were gaining more benefit from data
variance than data quantity.
While diversity is a common side-effect of scaling, it is a distinct phenomena.
The second was inspired by subword tokenization that is commonly used by many 
language models
\cite{philipNewAlgo1994,sennrich-etal-2016-neural,devlin2019bert,vaswani2017attention}
and experience with computational modeling.
The belief here is that by using non-overlapping patches we weaken the network's
ability to incorporate information along the boundaries of the images.
Such boundary conditions often plague computational models, requiring ghost
cells and other forms of boundary communication techniques to de-bias
calculations.

\begin{figure}[hbtp]
    \centering
    \includegraphics[width=\linewidth]{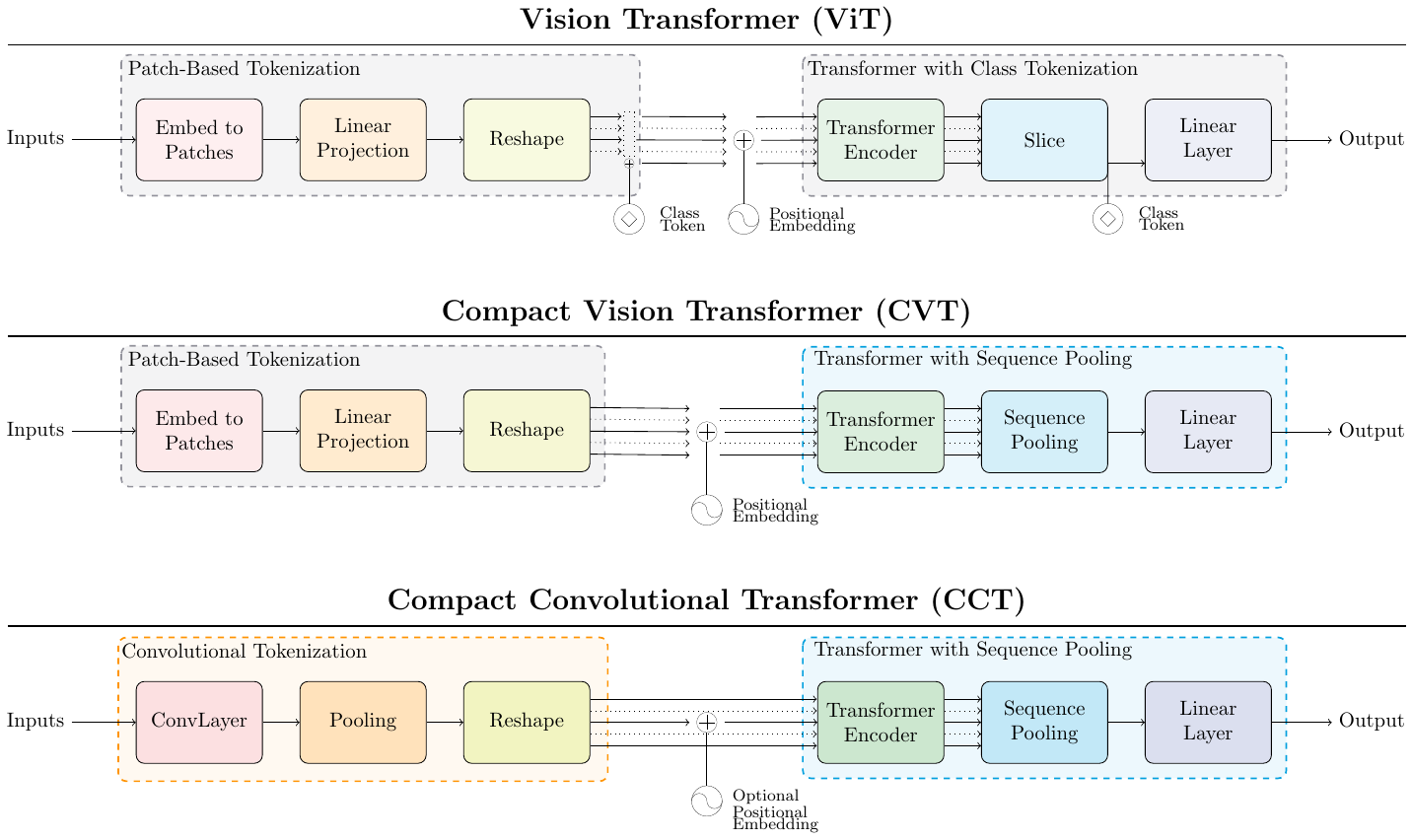}
    \caption[Variations of ViT Architectures]{A comparison of the Vision Transformer variants used throughout
    this study. On the left is the batching and embedding process
    (tokenization). On the right is the main neural architecture. The
    Transformer Encoder blocks and Linear Layers (classification sub-network) 
    are identical for all models. CVT follows ViT, removing the class token and
    introducing SeqPool. In CCT we modify the tokenization process, building
    from CVT.}
    \label{fig:vit-variants}
\end{figure}

ViT uses a simple patch and embedding procedure, where the image is evenly
divided into patches.
This in illustrated on the right half of \Cref{fig:cct-sym}, under Compact
Vision Transformer (CVT).
The process is to do a Group Normalization, ReLU, MaxPool, patch, and embed.
Notably, Dosovitskiy~\etal did the patching and embedding simultaniously with a 
convolution, matching strides to the kernel size~\footnote{This can be seen at
\href{https://github.com/google-research/vision_transformer/blob/main/vit_jax/models_vit.py\#L263-L270}{github.com/google-research/vit\_jax/models\_vit.py:264}}.
This same strategy is used for our ViT-Lite and Compact Vision Transformer (CVT)
models.
This procedure can be seen in \Cref{fig:vit-variants}.

We propose removing the restriction of making the convolutional kernels
and strides match, allowing these patches to overlap.
This would have an additional beneficial side-effect, allowing for better 
generalizability, by not requiring images to be integer multiples of the kernel
size.
This extends the embedding process to allow for arbitrary image sizes and aspect
ratios.
Additionally, we remove the Group Normalization layer from the ViT model,
finding it unnecessary.
Given an image or feature map $x\in\mathbb{R}^{H\times W\times C}$ we can
process our image as follows:
\begin{equation}
    \mathbf{x}_0 =
    \text{MaxPool}\left(
        \text{ReLU}\left(
            \text{Conv2d}(\textbf{x})
        \right)
    \right)
\end{equation}
Our convolution has a number of filters equal to the embedding dimension of the
transformer backbone, and both our convolution and pooling operations allow for
overlapping, which can introduce 
local inductive biases.

\subsection{SeqPool}\label{sec:escaping-seqpool}
In order to map the sequential output of a transformer to the linear
representation required by a feed-forward classification network ViT uses a
singular class index, or token, similar to language models like
BERT~\cite{devlin2019bert}.
This class token is learnable and then allows for the output of the transformer
to be sliced along the learned index.
Unfortunately, this underutilizes the relationships learned by the transformer
encoding layers.
This method makes the assumption that the transformer encoder can, and will,
decouple the relationships of the training data.
This disentanglement is the main task of the classification subnetwork, thus
forcing our Transformer to also perform this likely leads to underutilization
and overly constrains the encoding layers.

We propose SeqPool, an attention inspired pooling method.
The method is based on the assumption that the transformer encoder's output 
sequence contains information relevant to classification.
While this method is more computationally complex than slicing, it can 
reduce overall computation due to removal of an additional token that 
must be processed by the entirety of the network.
We use a network to generate a contraction 
$S : \mathbb{R}^{b\times n\times d}\mapsto \mathbb{R}^{b\times d}$, which then
is an appropriate shape to be processed by the classification sub-network.
\begin{equation}
    \text{Softmax}
    \left(
        g(\textbf{x}_L)^{T}
    \right)\textbf{x}_L
\end{equation}
Unlike dot-product attention we are not using keys, queries, and values, but 
instead learning a weighting of our sequence.
Our function $g$ is a single feed-forward layer mapping $g:\mathbb{R}^{b\times
n\times d} \mapsto \mathbb{R}^{b\times d}$.
We score this contraction and weight our original input producing the flattened
output.
This process can be seen as a learnable submersion, incorporating across
sequential data better, seemingly allowing us to take advantage of neuron 
polysemanticity~\cite{scherlispolysemanticity2022,hintonshaperepresentation1981} 
and superpositionality~\cite{elhage2022solu}.

\section{Experiments}\label{sec:escaping-experiments}
We perform a variety of experiments in order to test our research hypotheses.
We name our models similar to those of ViT, using the more explicit format: 
\begin{equation}
    [model]-[N\;layers]\;/\;[patch\;size]\times [N\;convolutions].
\end{equation}
The original ViT-B/16 model has 12 transformer encoder layers and a patch size
of 16, where we make the number of layers explicit: ViT-12/16.
We use this convention for all ViT and CVT models, dropping the number of
convolutions.
For CCT we specify the number of convolutions, even if only one.
This section is organized to first provide details of our experiments and
resources.
\Cref{sec:escaping-mainresults} contains our main results, demonstrating
high performance Vision Transformer models on small datasets.
\Cref{sec:escaping-ablations} includes details of our ablations, detailing the
effects of our changes to the architecture.
\Cref{sec:escaping-scaling} provides a scaling study, investigating the scaling
of both data and parameters.
Finally, \Cref{sec:escaping-nlp} includes our NLP experiments, to demonstrate
that these results generalize to language models.

\subsection{Datasets}\label{sec:escaping-datasets}
Our primary focus is on small datasets, where we train on CIFAR-10,
CIFAR-100~\cite{4531741}, MNIST~\cite{lecun2002gradient}, and
Fashion-MNIST~\cite{xiao2017}.
We also test our models on Oxford Flowers-102~\cite{4756141}~\footnote{We used
the dataset from Kaggle, which has a different data split than torchvision. 
Further discussion is provided later.} for
generalizability due to its large similarity between classes and high variance
for intra-class similarity.
We also use ImageNet~\cite{deng2009imagenet} to test the scailability of our
approach, allowing for more direct comparisons to ViT and DeiT.
We also test our approach in Natural Language Processing, using
AG-News~\cite{zhang2015character}, TREC~\cite{liroth2002learning},
SST~\cite{socher2013recursive}, IMDb~\cite{maas2011learning}, and
DBpedia~\cite{auer2007dbpedia}.

\subsection{Computational Resources}\label{sec:escaping-resources}
For most experiments we use a machine with an Intel(R) Core(TM) i9-9960X CPU @
3.10GHz and 4 NVIDIA RTX 2080Tis (11GB).
The exception was the CPU test which was performed with an AMD Ryzen 9 5900X,
where we found you could reach 90\% accuracy in under 30 minutes.
Our ImageNet experiments were performed on a single machine with either 2 AMD
EPYC) 7662s and 8 NVIDIA RTX A6000 (48GB) or 2 AMD EPYC 7713s and 8 NVIDIA A100s
(80GB).

\subsection{Hyperparameters}\label{sec:escaping-hyperparams}
We used the Pytorch Image Models library (timm)~\cite{rw2019timm} to train our
models for all image experiments.
Our augmentations include CutMix~\cite{yun2019cutmix},
Mixup~\cite{zhang2017mixup}, RandAugment~\cite{cubuk2020randaugment}, and Random
Erasing~\cite{zhong2020random}.
We performed hyperparameter sweeps for our differing methods and report the best
results we achieved.
All hyperparameter experiments were trained for 300 epochs, use a learning rate
of $5\times 10^{-4}$, a cosine learning rate scheduler, and weighted Adam 
optimizer ($\beta = \left[0.9, 0.999\right]$)\cite{kingma2017adam,Zhong_2020}.
For CNN models we found that some performed best with AdamW while others were
more performant with SGD with momentum $0.9$.
For reproducibility we release our checkpoints corresponding to the reported
numbers and YAML files corresponding to our experimental settings.
These can be found on our public GitHub 
repository~\footnote{\url{https://github.com/SHI-Labs/Compact-Transformers}}.

\begin{table}
    \begin{subfigure}{0.49\linewidth}
        \centering
        \resizebox{1.00\linewidth}{!}{%
            \begin{tabular}{l|cccc}
    \toprule
    \noalign{\smallskip}
    \textbf{Model} & \textbf{\# Layers} & \textbf{\# Heads} & \textbf{Ratio} & \textbf{Dim} \\
    \noalign{\smallskip}
    \hline
    \noalign{\smallskip}
    \textbf{ViT-Lite-6} & 6 & 4 & 2 & 256 \\
    \textbf{ViT-Lite-7} & 7 & 4 & 2 & 256 \\
    \noalign{\smallskip}
    \hline
    \noalign{\smallskip}
    \textbf{CVT-6} & 6 & 4 & 2 & 256 \\
    \textbf{CVT-7} & 7 & 4 & 2 & 256 \\
    \noalign{\smallskip}
    \hline
    \noalign{\smallskip}
    \textbf{CCT-2} & 2 & 2 & 1 & 128 \\
    \textbf{CCT-4} & 4 & 2 & 1 & 128 \\
    \textbf{CCT-6} & 6 & 4 & 2 & 256 \\
    \textbf{CCT-7} & 7 & 4 & 2 & 256 \\
    \textbf{CCT-14} & 14 & 6 & 3 & 384 \\
    \bottomrule
\end{tabular}

        }
        \caption{%
        Transformer Hyperparameters
        }\label{tab:escaping-variants1}
    \end{subfigure}
    \hfill
    \begin{subfigure}{0.49\linewidth}
        \centering
        \resizebox{1.00\linewidth}{!}{%
            \begin{tabular}{l|cccc}
    \toprule
    \noalign{\smallskip}
    \textbf{Model} & \textbf{\# Layers} & \textbf{\# Convs} & \textbf{Kernel} & \textbf{Stride} \\
    \noalign{\smallskip}
    \hline
    \noalign{\smallskip}
    \textbf{ViT-Lite-7/8} & 7 & 1 & 8\texttimes8 & 8\texttimes8 \\
    \textbf{ViT-Lite-7/4} & 7 & 1 & 4\texttimes4 & 4\texttimes4 \\
    \noalign{\smallskip}
    \hline
    \noalign{\smallskip}
    \textbf{CVT-7/8} & 7 & 1 & 8\texttimes8 & 8\texttimes8 \\
    \textbf{CVT-7/4} & 7 & 1 & 4\texttimes4 & 4\texttimes4 \\
    \noalign{\smallskip}
    \hline
    \noalign{\smallskip}
    \textbf{CCT-2/3x2} & 2 & 2 & 3\texttimes3 & 1\texttimes1 \\
    \textbf{CCT-7/3x1} & 7 & 1 & 3\texttimes3 & 1\texttimes1 \\
    \textbf{CCT-7/7x2} & 7 & 2 & 7\texttimes7 & 2\texttimes2 \\
    \bottomrule
\end{tabular}

        }
        \caption{%
        Tokenizer Hyperparameters
        }\label{tab:escaping-variants2}
    \end{subfigure}
    \caption[CCT Hyperparameters]{%
    Hyperparmeters used in different model configurations.
    \cref{tab:escaping-variants1} (left) shows transformer hyperparameters 
    while \cref{tab:escaping-variants2} (right) shows those for tokenizers.
    }\label{tab:escaping-hparams}
\end{table}

\begin{table}
    \centering
    \setlength{\tabcolsep}{4pt}
    \resizebox{\textwidth}{!}{
        \begin{tabular}{l|cccc|cr}
    \hline\noalign{\smallskip}
    \textbf{Model} & \textbf{CIFAR-10} & \textbf{CIFAR-100} &
    \textbf{FashionMNIST} & \textbf{MNIST} & \textbf{\# Params} & \textbf{FLOPs}\\
    \noalign{\smallskip}
    \hline
    \noalign{\smallskip}
    \multicolumn{7}{l}{\hspace{2em}\textit{Convolutional Networks (Designed for ImageNet)}}\\
    \noalign{\smallskip}
    \hline
    \noalign{\smallskip}
    \textbf{ResNet18} & $90.27 \% $ & $66.46 \% $ & $94.78 \% $ & $99.80 \% $ & $11.18$ M & $0.04$ G \\
    \textbf{ResNet34} & $90.51 \% $ & $66.84 \% $ & $94.78 \% $ & $99.77 \% $ & $21.29$ M & $0.08$ G \\
    \textbf{ResNet50} & $91.63 \% $ & $68.27 \% $ & $94.99 \% $ & $99.79 \% $ & $23.53$ M & $0.08$ G \\
    \noalign{\smallskip}
    \hline
    \noalign{\smallskip}
    \textbf{MobileNetV2/0.5} & $84.78 \% $ & $56.32 \% $ & $93.93 \% $ & $99.70 \% $ & $0.70$ M & $<\textbf{0.01}$ G \\
    \textbf{MobileNetV2/1.0} & $89.07 \% $ & $63.69 \% $ & $94.85 \% $ & $99.75 \% $ & $2.24$ M & $0.01$ G \\
    \textbf{MobileNetV2/1.25} & $90.60 \% $ & $65.24 \% $ & $95.05 \% $ & $99.77 \% $ & $3.47$ M & $0.01$ G \\
    \textbf{MobileNetV2/2.0} & $91.02 \% $ & $67.44 \% $ & $95.26 \% $ & $99.75\% $ & $8.72$ M & $0.02$ G \\
    \noalign{\smallskip}
    \hline
    \noalign{\smallskip}
    \multicolumn{7}{l}{\hspace{2em}\textit{Convolutional Networks (Designed for CIFAR)}}\\
    \noalign{\smallskip}
    \hline
    \noalign{\smallskip}
    \textbf{ResNet56\cite{he2016deep}} & $94.63 \% $ & $74.81 \% $ & $95.25 \% $ & $99.27 \% $ & $0.85$ M & $0.13$ G \\
    \textbf{ResNet110\cite{he2016deep}} & $95.08 \% $ & $76.63 \% $ & $95.32\% $ & $99.28 \% $ & $1.73$ M & $0.26$ G \\
    \textbf{ResNet164-v1\cite{he2016identity}} & $94.07 \% $ & $74.84 \% $ & $-$ & $-$ & $1.70$ M & $0.26$ G \\
    \textbf{ResNet164-v2\cite{he2016identity}} & $94.54 \% $ & $75.67 \% $ & $-$ & $-$ & $1.70$ M & $0.26$ G \\
    \textbf{ResNet1k-v1\cite{he2016identity}} & $92.39 \% $ & $72.18 \% $ & $-$ & $-$ & $10.33$ M & $1.55$ G \\
    \textbf{ResNet1k-v2\cite{he2016identity}} & $95.08 \% $ & $77.29 \% $ & $-$ & $-$ & $10.33$ M & $1.55$ G \\
    \textbf{ResNet1k-v2$^{\star}$\cite{he2016identity}} & $95.38 \% $ & $-$ & $-$ & $-$ & $10.33$ M & $1.55$ G \\
    \textbf{Proxyless-G\cite{cai2018proxylessnas}} & $97.92 \% $ & $-$ & $-$ & $-$ & $5.7$ M & $-$ \\
    \noalign{\smallskip}
    \hline
    \noalign{\smallskip}
    \multicolumn{7}{l}{\hspace{2em}\textit{Vision Transformers}}\\
    \noalign{\smallskip}
    \hline
    \noalign{\smallskip}
    \textbf{ViT-12/16} & $83.04 \% $ & $57.97 \% $ & $93.61 \% $ & $99.63 \% $ & $85.63$ M & $0.43$ G \\
    \noalign{\smallskip}
    \hline
    \noalign{\smallskip}
    \textbf{ViT-Lite-7/16} & $78.45 \% $ & $52.87 \% $ & $93.24 \% $ & $99.68 \% $ & $3.89$ M & $0.02$ G \\
    \textbf{ViT-Lite-6/16} & $78.12 \% $ & $52.68 \% $ & $93.09 \% $ & $99.66 \% $ & $3.36$ M & $0.02$ G \\
    \noalign{\smallskip}
    \hline
    \noalign{\smallskip}
    \textbf{ViT-Lite-7/8} & $89.10 \% $ & $67.27 \% $ & $94.49 \% $ & $99.69 \% $ & $3.74$ M & $0.06$ G \\
    \textbf{ViT-Lite-6/8} & $88.29 \% $ & $66.40 \% $ & $94.36 \% $ & $99.73 \% $ & $3.22$ M & $0.06$ G \\
    \noalign{\smallskip}
    \hline
    \noalign{\smallskip}
    \textbf{ViT-Lite-7/4} & $93.57 \% $ & $73.94 \% $ & $95.16 \% $ & $99.77 \% $ & $3.72$ M & $0.26$ G \\
    \textbf{ViT-Lite-6/4} & $93.08 \% $ & $73.33 \% $ & $95.14 \% $ & $99.74 \% $ & $3.19$ M & $0.22$ G \\
    \noalign{\smallskip}
    \hline
    \noalign{\smallskip}
    \multicolumn{7}{l}{\hspace{2em}\textit{Compact Vision Transformers}}\\
    \noalign{\smallskip}
    \hline
    \noalign{\smallskip}
    \textbf{CVT-7/8} & $89.79 \% $ & $70.11 \% $ & $94.50 \% $ & $99.70 \% $ & $3.74$ M & $0.06$ G \\
    \textbf{CVT-6/8} & $89.50 \% $ & $68.80 \% $ & $94.53 \% $ & $99.74 \% $ & $3.21$ M & $0.05$ G \\
    \noalign{\smallskip}
    \hline
    \noalign{\smallskip}
    \textbf{CVT-7/4} & $94.01 \% $ & $76.49 \% $ & $95.32 \% $ & $99.76 \% $ & $3.72$ M & $0.25$ G \\
    \textbf{CVT-6/4} & $93.60 \% $ & $74.23 \% $ & $95.00 \% $ & $99.75 \% $ & $3.19$ M & $0.22$ G \\
    \noalign{\smallskip}
    \hline
    \noalign{\smallskip}
    \multicolumn{7}{l}{\hspace{2em}\textit{Compact Convolutional Transformers}}\\
    \noalign{\smallskip}
    \hline
    \noalign{\smallskip}
    \textbf{CCT-2/3\texttimes2} & $89.75 \% $ & $66.93 \% $ & $94.08 \% $ & $99.70 \% $ & $\textbf{0.28}$ M & $0.04$ G \\
    \textbf{CCT-4/3\texttimes2} & $91.97 \% $ & $71.51 \% $ & $94.74 \% $ & $99.73 \% $ & $0.48$ M & $0.05$ G \\
    \textbf{CCT-6/3\texttimes2} & $94.43 \% $ & $77.14 \% $ & $95.34 \% $ & $99.75 \% $ & $3.33$ M & $0.25$ G \\
    \textbf{CCT-7/3\texttimes2} & $95.04 \% $ & $77.72 \% $ & $95.16 \% $ & $99.76 \% $ & $3.85$ M & $0.29$ G \\
    \noalign{\smallskip}
    \hline
    \noalign{\smallskip}
    \textbf{CCT-6/3\texttimes1} & $95.70 \% $ & $79.40 \% $ & $95.41 \% $ & $99.79 \% $ & $3.23$ M & $1.02$ G \\
    \textbf{CCT-7/3\texttimes1} & $96.53 \%$ & $80.92 \%$ & $\textbf{95.56} \%$ & $\textbf{99.82} \%$ & $3.76$ M & $1.19$ G \\
    \textbf{CCT-7/3\texttimes1}$^{\star}$ & $\textbf{98.00\%} $ & $\textbf{82.72\%} $ & $-$ & $-$ & $3.76$ M & $1.19$ G \\
    \hline
\end{tabular}

    }
    \caption[CCT Main Results]{Comparisons of various models when trained on
    small datasets. $\star$ was trained for longer, see
    \Cref{tab:escaping-longertrain} for additional details. Our 3.76M parameter
    CCT model is about to outperform both ResNets and ViTs across all datasets,
    with longer training only being necessary to outperform the 5.7M Proxyless-G
    model on CIFAR-10.
    }\label{tab:escaping-main}
\end{table}

\subsection{Transformers On Small Datasets}\label{sec:escaping-mainresults}
The main results of this work are the success of training Vision Transformers on
small datasets. 
We follow the aforementioned training procedure, except our best model we
further train as it did not appear to be saturated.
Our full results can be read in \Cref{tab:escaping-main}, where we show a
comparison of various ResNet based models, ViTs, CVT, and CCT, testing our small
vision datasets with comparisons of model size and required compute.
Notably, on CIFAR-10, we are able to achieve a 10\% improvement over similarly 
sized ViT-Lite models (ViT-Lite-7/8) and an 18\% improvement over the ViT-12/16
(ViT-B/16) model while our model has a 95.6\% reduction in the number of
parameters.
Our best model only contains a single convolutional layer within the embedding
process, meaning that the transformer architecture is performing the main
computation, achieving an accuracy of 98\% while using only 3.76M parameters.
This result is only slightly less than Vaswani~\etal's much larger models that
include JFT-300M or ImageNet-21k pretraining and outperforms VIT-12/32, ViT-24/16,
and ViT-24/32 when using ImageNet-1k pretraining and fine-tuning at 384
resolution (Table 5 of Vaswani~\etal).
We found that an increase in convolutions tended to harm model performance.

\begin{table}[htpb]
    \centering
    \begin{tabular}{l|c|cc}
    \toprule
    \noalign{\smallskip}
    \textbf{\# Epochs} & \textbf{Pos. Emb.} & \textbf{CIFAR-10} & \textbf{CIFAR-100}\\
    \noalign{\smallskip}
    \hline
    \noalign{\smallskip}
    300 & Learnable & $96.53 \% $ & $80.92 \% $ \\
    1500 & Sinusoidal & $97.48 \% $ & $82.72\% $ \\
    5000 & Sinusoidal & $\textbf{98.00\%} $ & $\textbf{82.87\%}$ \\
    \hline
\end{tabular}

    \caption[Extended CCT Training]{%
        Training of \textbf{CCT-7/3\texttimes1} with an increased number of
        epochs.
    }\label{tab:escaping-longertrain}
\end{table}

These results show that our CCT based model is able to outperform both standard
Vision Transformers as well as ResNet models.
We demonstrate that neither large scale pretraining nor knowledge distillation 
are needed to overcome the biases found in smaller scale data.
Furthermore, we strongly suspect that the underlying issue is due to the
tokenization process of overlapping patches.
We include a comparison of Salient
Maps~\cite{erhanVisualizing2009,simonyan2014deepinsideconvolutionalnetworks} in
\Cref{fig:escaping-salients}, comparing visualizations on ImageNet.
Saliency maps operate by looking visualizing the gradient accumulations across
the network.
We should take care as to fully interpret the semantic meaning of these maps,
but the visualizations do clearly indicate how the original patching may be
recovered in the standard ViT model while we have a much smoother representation
in CCT, evidencing the first research hypothesis.

\begin{figure}[htbp]
    \centering
    \begin{subfigure}{0.24\linewidth}
        \makebox[\linewidth][c]{\Large{\textbf{ImageNet}}}\\
        \includegraphics[width=\linewidth]{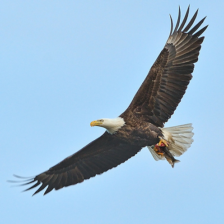}
    \end{subfigure}
    \begin{subfigure}{0.24\linewidth}
        \makebox[\linewidth][c]{\Large{\textbf{ViT}}}\\
        \includegraphics[width=\linewidth]{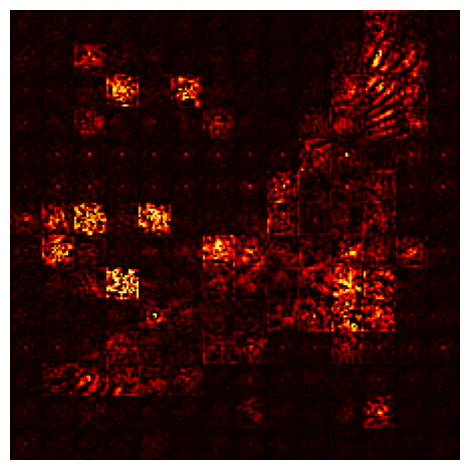}
    \end{subfigure}
    \begin{subfigure}{0.24\linewidth}
        \makebox[\linewidth][c]{\Large{\textbf{CCT}}}\\
        \includegraphics[width=\linewidth]{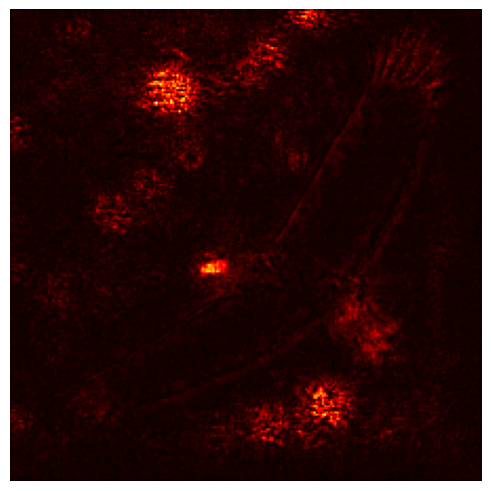}
    \end{subfigure}
    \begin{subfigure}{0.24\linewidth}
        \makebox[\linewidth][c]{\Large{\textbf{NAT}}}\\
        \includegraphics[width=\linewidth]{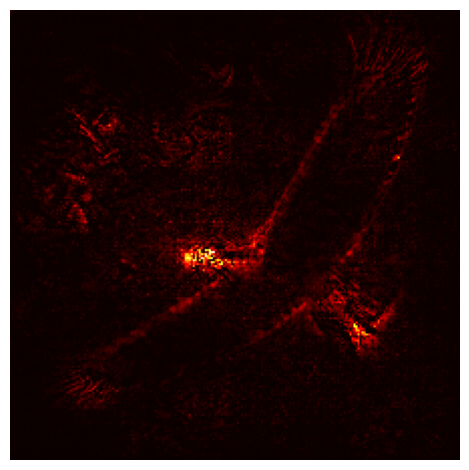}
    \end{subfigure}
    \\
    \begin{subfigure}{0.24\linewidth}
        \includegraphics[width=\linewidth]{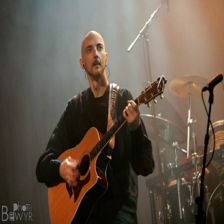}
    \end{subfigure}
    \begin{subfigure}{0.24\linewidth}
        \includegraphics[width=\linewidth]{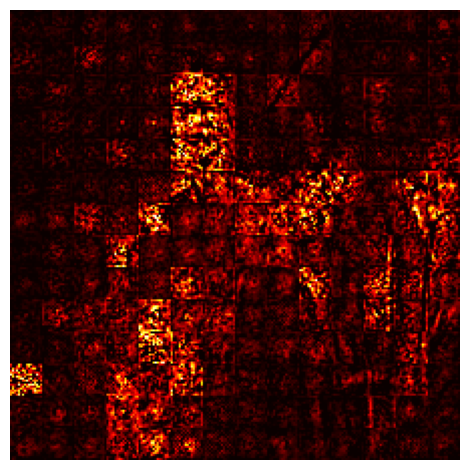}
    \end{subfigure}
    \begin{subfigure}{0.24\linewidth}
        \includegraphics[width=\linewidth]{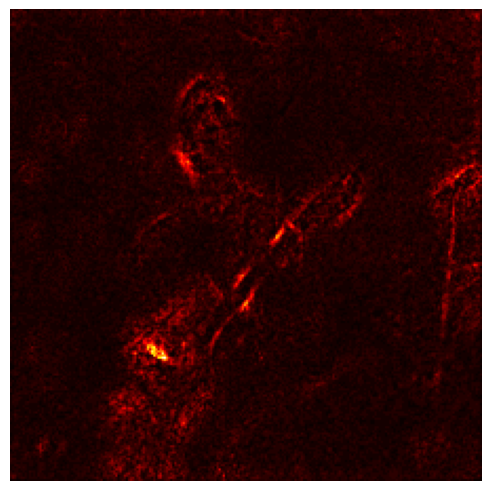}
    \end{subfigure}
    \begin{subfigure}{0.24\linewidth}
        \includegraphics[width=\linewidth]{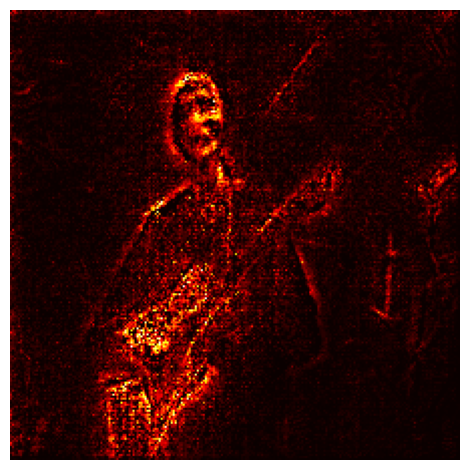}
    \end{subfigure}
    \\
    \begin{subfigure}{0.24\linewidth}
        \includegraphics[width=\linewidth]{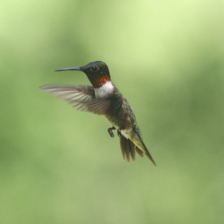}
    \end{subfigure}
    \begin{subfigure}{0.24\linewidth}
        \includegraphics[width=\linewidth]{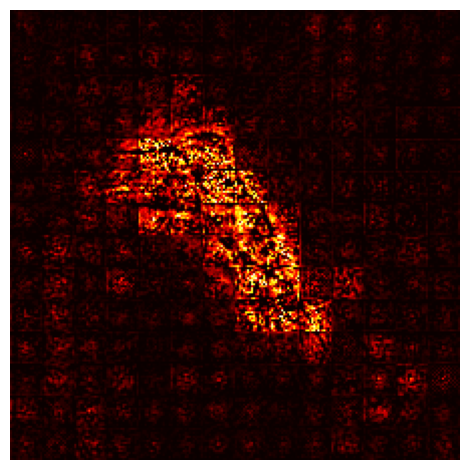}
    \end{subfigure}
    \begin{subfigure}{0.24\linewidth}
        \includegraphics[width=\linewidth]{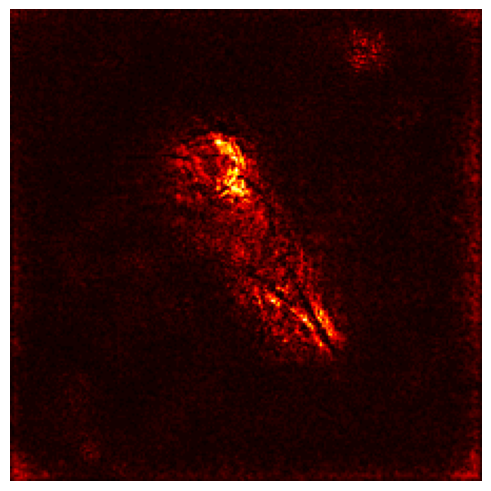}
    \end{subfigure}
    \begin{subfigure}{0.24\linewidth}
        \includegraphics[width=\linewidth]{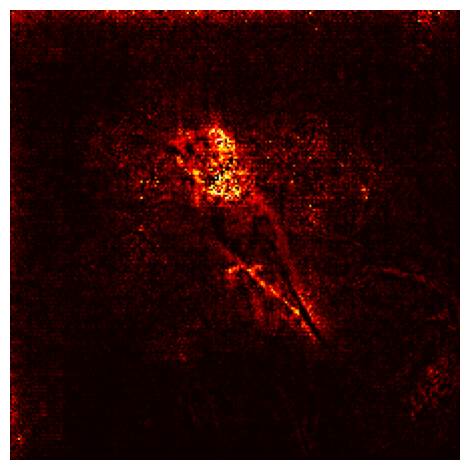}
    \end{subfigure}
    \\
    \begin{subfigure}{0.24\linewidth}
        \includegraphics[width=\linewidth]{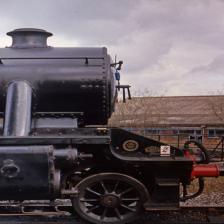}
    \end{subfigure}
    \begin{subfigure}{0.24\linewidth}
        \includegraphics[width=\linewidth]{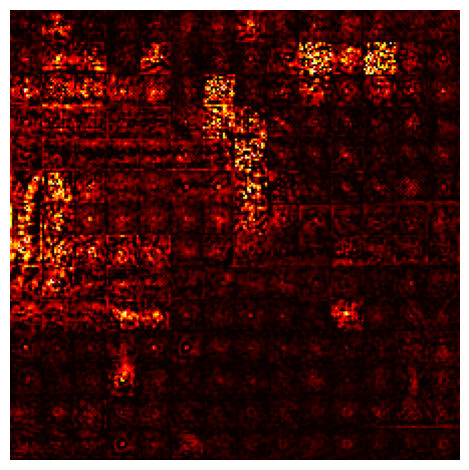}
    \end{subfigure}
    \begin{subfigure}{0.24\linewidth}
        \includegraphics[width=\linewidth]{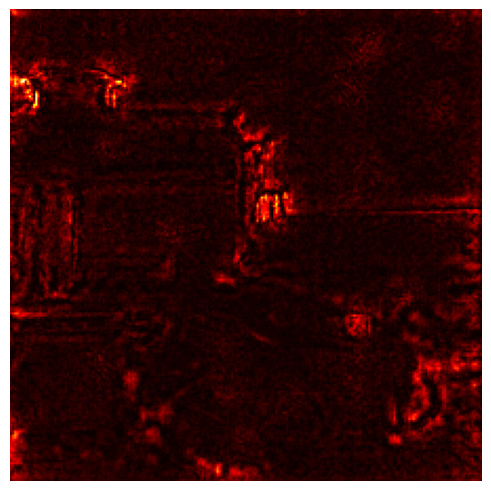}
    \end{subfigure}
    \begin{subfigure}{0.24\linewidth}
        \includegraphics[width=\linewidth]{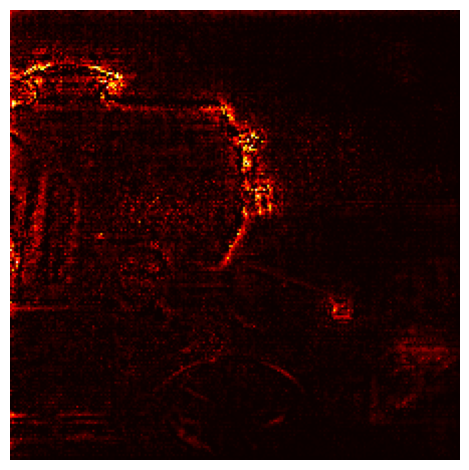}
    \end{subfigure}
    \caption[Vision Transformer Salient Maps]{%
        Salient maps of ViT, CCT, and NAT based on ImageNet-1k. 
        It can be seen that CCT removes the blocking artifacts from ViT. 
        CCT sometimes creates displacement, but this is resolved by NAT
        (presented in \Cref{ch:stylenat}).
    }\label{fig:escaping-salients}
\end{figure}

\begin{table}[hbtp]
    \setlength{\tabcolsep}{6pt}
    \centering
    \resizebox{\textwidth}{!}{
        \begin{tabular}{l|ccccc|cc|cc}
    \toprule
    \textbf{Model}
        & \textbf{CLS}
        & \textbf{\# Conv}
        & \textbf{Conv Size}
        & \textbf{Aug}
        & \textbf{Tuning}
        & \textbf{C-10}
        & \textbf{C-100}
        & \textbf{\# Params}
        & \textbf{FLOPS}\\
    \noalign{\smallskip}
    \toprule
    \noalign{\smallskip}
    \multicolumn{10}{l}{\hspace{2em}\textbf{\emph{``Large'' Models} ($\approx
            85$M Parameters)}}\\
    \noalign{\smallskip}
    \toprule
    \textbf{ViT-12/16} & CT & \xmark & \xmark & \xmark & \xmark & $69.82\%$ & $40.57\%$ & $85.63$ M & $0.43$ G \\
    \noalign{\smallskip}
    \hline
    \noalign{\smallskip}
    \textbf{ViT-12/16} & CT & \xmark & \xmark & \cmark & \cmark & $80.72\%$ & $56.73\%$ & $85.63$ M & $0.43$ G \\
    \textbf{CVT-12/16}  & SP & \xmark & \xmark & \cmark & \cmark & $80.84\%$ & $58.05\%$ & $85.63$ M & $0.34$ G \\
    
    \noalign{\smallskip}
    \hline
    \noalign{\smallskip}
    
    \textbf{ViT-12/8}  & CT & \xmark & \xmark & \cmark & \cmark & $90.24\%$ & $69.81\%$ & $85.20$ M & $1.45$ G \\
    \textbf{ViT-12/4}  & CT & \xmark & \xmark & \cmark & \cmark & $94.07\%$ & $76.08\%$ & $85.12$ M & $5.61$ G \\
    
    \noalign{\smallskip}
    \hline
    \noalign{\smallskip}
    \textbf{CCT-12/7\texttimes1}  & SP & 1 & $7 \times 7$ & \cmark & \cmark & $93.72\%$ & $76.21\%$ & $85.20$ M & $5.55$ G \\
    \textbf{CCT-12/3\texttimes2}  & SP & 2 & $3 \times 3$ & \cmark & \cmark & $\textbf{94.50\%}$ & $\textbf{77.05\%}$ & $85.53$ M & $5.63$ G \\
    
    \noalign{\smallskip}
    \toprule
    \noalign{\smallskip}
    \multicolumn{10}{l}{\hspace{2em}\textbf{\emph{Small Models} ($\approx 4$M
            Parameters)}}\\
    \noalign{\smallskip}
    \toprule
    \textbf{ViT-Lite-7/16} & CT & \xmark & \xmark & \xmark & \xmark & $71.78\%$ & $41.59\%$ & $3.89$ M & $0.02$ G \\
    \textbf{ViT-Lite-7/8} & CT & \xmark & \xmark & \xmark & \xmark & $83.38\%$ & $55.69\%$ & $3.74$ M & $0.06$ G \\
    \textbf{ViT-Lite-7/4} & CT & \xmark & \xmark & \xmark & \xmark & $83.59\%$ & $58.43\%$ & $3.72$ M & $0.26$ G \\
    \noalign{\smallskip}
    \hline
    \noalign{\smallskip}
    \textbf{CVT-7/16} & SP & \xmark & \xmark & \xmark & \xmark & $72.26\%$ & $42.37\%$ & $3.89$ M & $0.02$ G \\
    \textbf{CVT-7/8}  & SP & \xmark & \xmark & \xmark & \xmark & $84.24\%$ & $55.49\%$ & $3.74$ M & $0.06$ G \\
    \textbf{CVT-7/8}  & SP & \xmark & \xmark & \cmark & \xmark & $87.15\%$ & $63.14\%$ & $3.74$ M & $0.06$ G \\
    \textbf{CVT-7/4}  & SP & \xmark & \xmark & \xmark & \xmark & $88.06\%$ & $62.06\%$ & $3.72$ M & $0.25$ G \\
    \textbf{CVT-7/4}  & SP & \xmark & \xmark & \cmark & \xmark & $91.72\%$ & $69.59\%$ & $3.72$ M & $0.25$ G \\
    \textbf{CVT-7/4}  & SP & \xmark & \xmark & \cmark & \cmark & $92.43\%$ & $73.01\%$ & $3.72$ M & $0.25$ G \\
    \textbf{CVT-7/2}  & SP & \xmark & \xmark & \xmark & \xmark & $84.80\%$ & $57.98\%$ & $3.76$ M & $1.18$ G \\
    \noalign{\smallskip}
    \hline
    \noalign{\smallskip}
    \textbf{CCT-7/7\texttimes1} & SP & 1 & $7 \times 7$ & \xmark & \xmark & $87.81\%$ & $62.83\%$ & $3.74$ M & $0.26$ G \\
    \textbf{CCT-7/7\texttimes1} & SP & 1 & $7 \times 7$ & \cmark & \xmark & $91.85\%$ & $69.43\%$ & $3.74$ M & $0.26$ G \\
    \noalign{\smallskip}
    \hline
    \noalign{\smallskip}
    \textbf{CCT-7/7\texttimes1} & CT & 1 & $7 \times 7$ & \cmark & \cmark & $91.67\%$ & $72.07\%$ & $3.74$ M & $0.26$ G \\
    \textbf{CCT-7/7\texttimes1} & SP & 1 & $7 \times 7$ & \cmark & \cmark & $92.29\%$ & $72.46\%$ & $3.74$ M & $0.26$ G \\
    \noalign{\smallskip}
    \hline
    \noalign{\smallskip}
    \textbf{CCT-7/3\texttimes2} & CT & 2 & $3 \times 3$ & \cmark & \cmark & $93.36\%$ & $74.77\%$ & $3.85$ M & $0.29$ G \\
    \textbf{CCT-7/3\texttimes2} & SP & 2 & $3 \times 3$ & \cmark & \cmark & $93.65\%$ & $74.77\%$ & $3.85$ M & $0.29$ G \\
    \noalign{\smallskip}
    \hline
    \noalign{\smallskip}
    \textbf{CCT-7/3\texttimes1} & SP & 1 & $3 \times 3$ & \cmark & \cmark & $\textbf{94.47\%}$ & $\textbf{75.59\%}$ & $3.76$ M & $1.19$ G \\
    \bottomrule
\end{tabular}

    }
    \caption[CCT Ablation Study]{%
        Ablation study, transforming ViT into CCT. We measure CIFAR validation 
        accuracy across each modification as well as the number of model
        parameters and computation (MACs). All ViT models use a class token
        (CT), while CVT and CCT use SeqPool (SP). We report the number of
        convolutions used during embedding (\# Conv), its kernel size, if we
        utilized image augmentation (Aug), and tuning.
    }\label{tab:escaping-ablation}
\end{table}

\subsection{Ablations}\label{sec:escaping-ablations}
We include ablations of our parameters to better understand the impact of our
changes to the ViT model.
In \Cref{tab:escaping-ablation} we step through the process of converting our
ViT model into CCT.%
In our table we denote if we used a class token (CT) or SeqPool (SP), the number
of convolutions user (overlapping patches), the kernel size, if image
augmentations were used, and additional tuning.
Our tuning includes dropout, attention dropout, and stochastic depth.
We separate our models into two sections, with ``Large'' models, with
approximately 85M parameters and small models, with approximately 4M parameters.

By directly comparing similar ViT-Lite models to our CVT models we can see the
effect of our SeqPool method.
In all cases we see that there is a minor performance improvement due to this,
with a much lower effect with the large 85M parameter models.
When comparing on CIFAR-10, models with 7 transformer encoders, a patch size of 
16 we observe a $0.7\%$ increase, $1.0\%$ for a patch size of 8, a $5.3\%$ 
increase with a patch size of 4.
For the larger 12 transformer layer models with a patch size of 16 we only
notice a $0.1\%$ increase, but these models included tuning and augmentation,
likely reducing the impact.

In the smaller models we see that the larger contribution to performance
increase is due to decreased patch size.
For ViT models, decreasing from a patch size of 16 to 8 increased model
performance by $16.2\%$, but reducing to a patch size of 4 only accounted for an
additional $0.3\%$ increase.
For CVT the decrease to a patch size of 8 showed a similar $16.6\%$ improvement,
but further reduction to a patch of 4 gave another $4.5\%$ increase.
Larger impacts can be observed when looking at CIFAR-100, except in the case of
a patch size of 8 where SeqPool appears to have a slight negative ($<0.5\%$)
impact.
We see a $+1.8\%$, $-0.4\%$, and $+6.2\%$ difference for SeqPool, for our 3
patches.
On ViT the patch reduction accounts for a $33.9\%$ and $4.9\%$ improvements
while CVT shows $31.0\%$ and $11.8\%$ improvements.
With decreased patch sizing the transformer appears to be able to overcome the
primary issues presented by smaller training sets.
Our SeqPool method still demonstrates greater performance, especially as patch
size decreases, showing greater network utilization.

\begin{table}[hbtp]
    \centering
    \resizebox{0.90\textwidth}{!}{
        \begin{tabular}{lc|cc}
    \hline\noalign{\smallskip}
    Model & PE & CIFAR-10 & CIFAR-100\\
    \noalign{\smallskip}
    \hline
    \noalign{\smallskip}
    \multicolumn{4}{l}{\textit{Conventional Vision Transformers are more dependent on Positional Embedding }}\\
    \noalign{\smallskip}
    \hline
    \noalign{\smallskip}
    
    \multirow{3}{*}{ViT-12/16} & Learnable & $69.82\%$ {\color{red}\small \textit{($+3.11\%$)}} & $40.57\%$ {\color{red}\small \textit{($+1.01\%$)}} \\
    & Sinusoidal & $69.03\%$ {\color{red}\small \textit{($+2.32\%$)}} & $39.48\%$ {\color{blue}\small \textit{($-0.08\%$)}}\\
    & None & $66.71\%$ {\color{black}(\small \textit{baseline)}} & $39.56\%$ {\color{black}(\small \textit{baseline)}}\\
    
    \noalign{\smallskip}
    \hline
    \noalign{\smallskip}
    
    \multirow{3}{*}{ViT-Lite-7/8} & Learnable & $83.38\%$ {\color{red}\small \textit{($+7.25\%$)}} & $55.69\%$ {\color{red}\small \textit{($+7.15\%$)}} \\
    & Sinusoidal & $80.86\%$ {\color{red}\small \textit{($+4.73\%$)}} & $53.50\%$ {\color{red}\small \textit{($+4.96\%$)}} \\
    & None & $76.13\%$ {\color{black}(\small \textit{baseline)}} & $48.54\%$ {\color{black}(\small \textit{baseline)}} \\
    
    \noalign{\smallskip}
    \hline
    \noalign{\smallskip}
    
    \multirow{3}{*}{CVT-7/8} & Learnable & $84.24\%$ {\color{red}\small \textit{($+6.52\%$)}} & $55.49\%$ {\color{red}\small \textit{($+7.23\%$)}} \\
    & Sinusoidal & $80.84\%$ {\color{red}\small \textit{($+3.12\%$)}} & $50.82\%$ {\color{red}\small \textit{($+2.56\%$)}} \\
    & None & $77.72\%$ {\color{black}(\small \textit{baseline)}} & $48.26\%$ {\color{black}(\small \textit{baseline)}} \\
    
    \noalign{\smallskip}
    \hline
    \noalign{\smallskip}
    \multicolumn{4}{l}{\textit{Compact Convolutional Transformers are less dependent on Positional Embedding }}\\
    \noalign{\smallskip}
    \hline
    \noalign{\smallskip}
    
    \multirow{3}{*}{CCT-7/7} & Learnable & $82.03\%$ {\color{red}\small \textit{($+0.21\%$)}} & $63.01\%$ {\color{red}\small \textit{($+3.24\%$)}} \\
    & Sinusoidal & $81.15\%$ {\color{blue}\small \textit{($-0.67\%$)}} & $60.40\%$ {\color{red}\small \textit{($+0.63\%$)}} \\
    & None & $81.82\%$ {\color{black}(\small \textit{baseline)}} & $59.77\%$ {\color{black}(\small \textit{baseline)}} \\

    \noalign{\smallskip}
    \hline
    \noalign{\smallskip}
    \multirow{3}{*}{CCT-7/3\texttimes2} & Learnable & $90.69\%$ {\color{red}\small \textit{($+1.67\%$)}} & $65.88\%$ {\color{red}\small \textit{($+2.82\%$)}} \\
    & Sinusoidal & $89.93\%$ {\color{red}\small \textit{($+0.91\%$)}} & $64.12\%$ {\color{red}\small \textit{($+1.06\%$)}} \\
    & None & $89.02\%$ {\color{black}(\small \textit{baseline)}} & $63.06\%$ {\color{black}(\small \textit{baseline)}} \\
    \noalign{\smallskip}
    \hline
    \noalign{\smallskip}
    \multirow{3}{*}{CCT-7/3\texttimes2$^\dagger$} & Learnable & $95.04\%$ {\color{red}\small \textit{($+0.64\%$)}} & $77.72\%$ {\color{red}\small \textit{($+0.20\%$)}} \\
    & Sinusoidal & $94.80\%$ {\color{red}\small \textit{($+0.40\%$)}} & $77.82\%$ {\color{red}\small \textit{($+0.30\%$)}} \\
    & None & $94.40\%$ {\color{black}(\small \textit{baseline)}} & $77.52\%$ {\color{black}(\small \textit{baseline)}} \\
    \noalign{\smallskip}
    \hline
    \noalign{\smallskip}
    \multirow{3}{*}{CCT-7/3\texttimes1$^\dagger$} & Learnable & $96.53\%$ {\color{red}\small \textit{($+0.29\%$)}} & $80.92\%$ {\color{red}\small \textit{($+0.65\%$)}} \\
    & Sinusoidal & $96.27\%$ {\color{red}\small \textit{($+0.03\%$)}} & $80.12\%$ {\color{blue}\small \textit{($-0.15\%$)}} \\
    & None & $96.24\%$ {\color{black}(\small \textit{baseline)}} & $80.27\%$ {\color{black}(\small \textit{baseline)}} \\
    \noalign{\smallskip}
    \hline
    \noalign{\smallskip}
    
    \multirow{3}{*}{CCT-7/7\texttimes1-noSeqPool} & Learnable & $82.41\%$ {\color{red}\small \textit{($+0.12\%$)}} & $62.61\%$ {\color{red}\small \textit{($+3.31\%$)}} \\
    & Sinusoidal & $81.94\%$ {\color{blue}\small \textit{($-0.35\%$)}} & $61.04\%$ {\color{red}\small \textit{($+1.74\%$)}} \\
    & None & $82.29\%$ {\color{black}(\small \textit{baseline)}} & $59.30\%$ {\color{black}(\small \textit{baseline)}} \\
    
    \noalign{\smallskip}
    \hline
    \noalign{\smallskip}
    
    \multirow{3}{*}{CCT-7/3\texttimes2-noSeqPool} & Learnable & $90.41\%$ {\color{red}\small \textit{($+1.49\%$)}} & $66.57\%$ {\color{red}\small \textit{($+1.40\%$)}} \\
    & Sinusoidal & $89.84\%$ {\color{red}\small \textit{($+0.92\%$)}} & $64.71\%$ {\color{blue}\small \textit{($-0.46\%$)}} \\
    & None & $88.92\%$ {\color{black}(\small \textit{baseline)}} & $65.17\%$ {\color{black}(\small \textit{baseline)}} \\
    \hline
\end{tabular}

    }
    \caption[CCT Positional Embedding Comparison]{%
    Validation accuracy comparison comparing Positional
    Embedding method. Augmentations and training techniques such as Mixup and
    CutMix were turned off for these experiments to better highlight differences.
    The numbers reported are best out of 4 runs with random initializations. 
    $\dagger$ denotes model trained with extra augmentation and hyperparameter 
    tuning.
    }\label{tab:pe_comparison}
\end{table}

The largest gains come from moving to CCT, which can also better take advantage 
of data augmentations, showing better capacity for generalization.
For example, ViT-Lite-7/8, CVT-7/8, and CCT-7/3\texttimes1 all have 3.74M
parameters, but their CIFAR-10 scores are $83.38\%$, $84.24\%$, and $87.81\%$
respectively.
Where CVT shows a $1\%$ improvement, CCT shows $5.3\%$.
We can see that CVT-7/8 improves to $87.15\%$ ($2.91\%$), while CCT-7/3\texttimes1
improves to $91.85\%$ ($4.04\%$) when introducing augmentation.
We can also see in our CCT experiments that by removing SeqPool and
reintroducing the class tokens that we drop performance by $0.62\%$,
demonstrating that SeqPool does not account for these differences.
A similar pattern can be found with larger models, though our comparisons are 
not as thorough. 
These results show that the overlapping patches and better extraction of data
from the transformer architecture result in significant improvements, evidencing
our first two hypotheses.

We also include a short study on Positional Embedding, in
\Cref{tab:pe_comparison}.
Because our overlapping tokenization allowed us to debias some of the positional
relationships within the data we test to find the importance of positional
embedding.
While ViT and CVT benefit strongly from positional embedding, CCT only gets
minor benefits.
This further demonstrates the bias introduced by patching in ViT.
Some additional positional embedding comparisons can be found in
\Cref{fig:escaping-hwcut}.

\subsection{Scaling Study}\label{sec:escaping-scaling}
While the previous results demonstrate that pretraining is unnecessary for
Vision Transformers to be effective on small datasets, we need to understand the
relationship of model size, data quantity, and data quality.

In order to address the \emph{Scale is All You Need} arguments, we begin with the
study of model size.
Our main study of model size can be seen in our ablations 
(\Cref{tab:escaping-ablation}), where we observe that our larger 85.53M
parameter model outperforms out 3.76M parameter model on both CIFAR-10 and
CIFAR-100, showing very minor improvement on CIFAR-10 and a 2\% increase on
CIFAR-100.

This result runs counter to ViT, where the larger model has a performance
decrease of up to $16.5\%$ and $30.5\%$, respectively.
When given additional augmentation, the larger ViT model is only able to 
outperform our largest ViT-Lite-7/16 model, which did not use tuning or
augmented training.
The two slightly smaller ViT-Lite models are still able to outperform this large
model without the inclusion of additional augmentation or training,
demonstrating that the smaller patch sizes play a more significant role, as
discussed in \Cref{sec:escaping-ablations}.
We believe that the smaller window sizes allow the transformer architecture to
better integrate data across patches, learning convolutions similar to what
Cordonnier~\etal had shown, but further study is required to confirm or deny.
The increase relationship between patch size and performance applies to both
large and small ViTs, with the large ViT approaching the performance of CCT
(surpassing ViT-Lite models) once the patch size is reduced to 4, yet still do
not surpass the performance of small CCT models on CIFAR-10.

Under most configurations, CVT also shows a decrease in performance, again with
improved performance primarily being attributed to the path size.
Performance decreases at a patch size of 2, similar to Cordonnier~\etal's
configuration, showing that the patches can be too small.
In a way, this demonstrates that scale plays an important role, but these trends
run counter to the conventional wisdom.
These results demonstrate the importance of the embedding process and that
na\"{i}vely scaling architectures may instead hinder performance.
Careful design of the neural architecture trumps scaling.

\begin{table}[hbtp]
    \centering
    \begin{tabular}{l|c|cc|c}
    \toprule
    \noalign{\smallskip}
    \textbf{Model} & \textbf{Top-1} & \textbf{\# Params} & \textbf{FLOPS} & \textbf{Epochs}\\
    \noalign{\smallskip}
    \hline
    \noalign{\smallskip}
    \textbf{ResNet50} & $77.15\%$ & $25.55$ M & $4.15$ G & $120$ \\
    \textbf{ResNet50 (2021)} & $79.80\%$ & $25.55$ M & $4.15$ G & $300$ \\
    \textbf{ViT-S}& $79.85\%$ & $\textbf{22.05}$ M & $\textbf{4.61}$ G & $300$ \\
    \textbf{CCT-14/7\texttimes2} & $\textbf{80.67\%}$ & $22.36$ M & $5.53$ G & $300$ \\
    \noalign{\smallskip}
    \midrule
    \noalign{\smallskip}
    \textbf{DeiT-S~\alambic} & $81.16\%$ & $22.44$M & $\textbf{4.63}$ G & $300$ \\
    \textbf{CCT-14/7\texttimes2~\alambic} & $\textbf{81.34\%}$ & $\textbf{22.36}$ M & $5.53$ G & $300$ \\
    \hline
\end{tabular}

    \caption[CCT ImageNet Accuracy]{ImageNet Top-1 validation accuracy comparison
        (no extra data or pretraining). Models with \alambic denotes 
        distillation and follow the knowledge distillation process as described
        in Touvron \emph{et al}~\cite{pmlr-v139-touvron21a}. 
        ResNet50 (2021) is reported from
        \cite{wightman2021resnetstrikesbackimproved} which has the same 
        training recipe as ours.%
        }\label{tab:escaping-imagenet}
\end{table}

To study relation of data to model performance we perform multiple scaling
studies.
In order to complete our parameter scaling study, we test our model's 
performance on larger amounts of data, with ImageNet, but leave further large
model and large data scaling studies to labs with resources similar to 
Vaswani~\etal.
In \Cref{tab:escaping-imagenet} we train a 14 layer (22M param) model, and
compare it to ViT-S and DeiT-S models from 
Touvron~\etal\cite{pmlr-v139-touvron21a}.
It is difficult to get these models to be exactly the same parameter size, but
our model is able to still outperform ViT on ImageNet-1k without any
pretraining.
We also compare to DeiT-S~\alambic, where our model is slightly smaller,
following the same knowledge distillation process.
Our model again shows improvements, demonstrating that our procedure does not
produce negative effects with increased data scale.

\begin{table}[hbtp]
    \centering
    \begin{tabular}{l|cc|c|cc}
    \hline\noalign{\smallskip}
    \textbf{Model} & \textbf{Resolution} & \textbf{Pretraining} & \textbf{Top-1} & \textbf{\# Params} & \textbf{FLOPs}\\
    \noalign{\smallskip}
    \hline
    \noalign{\smallskip}
    \textbf{CCT-14/7\texttimes2} & 224 & - & $97.19\%$ & $\phantom{0}22.17$ M & $\phantom{0}18.63$ G\\
    \noalign{\smallskip}
    \hline
    \noalign{\smallskip}
    \textbf{DeiT-B} & 384 & ImageNet-1k & $98.80\%$ & $\phantom{0}86.25$ M & $\phantom{0}55.68$ G\\
    \textbf{ViT-L/16} & 384 & JFT-300M & $99.74\%$ & $304.71$ M & $191.30$ G\\
    \textbf{ViT-H/14} & 384 & JFT-300M & $99.68\%$ & $661.00$ M & $504.00$ G\\
    \textbf{CCT-14/7\texttimes2} & 384 & ImageNet-1k & $\textbf{99.76\%}$ & $\phantom{0}\textbf{22.17}$ M & $\phantom{0}\textbf{18.63}$ G\\
    \hline
\end{tabular}

    \caption[CCT Flowers-102 Accuracy]{Flowers-102 Top-1 validation accuracy comparison. CCT outperforms 
    other competitive models, having significantly fewer parameters and GFLOPs. 
    This demonstrates the compactness on small datasets even with large images.
    }\label{tab:escaping-flowers}
\end{table}

We also test our 22M parameter model on the Flowers-102 dataset, which is
designed for high data variance and to test model generalizability.
For this we are able to achieve an accuracy of over $97\%$ without the use of
any pretraining data or higher resolution tuning.
These results can be found in \Cref{tab:escaping-flowers}.
When using ImageNet-1k pretraining and including higher resolution tuning,
following the procedure of DeiT, we are able to achieve state of the art
results, outperforming models that included more than a magnitude more
parameters and a more than a magnitude amount of pretraining data.
It should be noted that we used the Flowers-102 dataset provided from Kaggle and
that this uses a different data split than that which is included in the
torchvision version\footnote{The torchvision dataset collection did not include
Flowers-102 when initially trained.}.
This was brought to our attention through a GitHub
issue,\footnote{A wandb report showing training results can be found alongside
the issue here: \url{https://github.com/SHI-Labs/Compact-Transformers/issues/65}} 
where a user was unable to replicate our results.
We retrained our CCT-7/7\texttimes2 (4M params) and CCT-14/7\texttimes2 models 
at 224 resolution and obtained $68.26\%$ and $68.85\%$ accuracy, respectively.
When applying the same procedure to ViT-S/16 we obtained a result of $48.63\%$,
only showing our model having better performance applied to this
dataset.

\begin{figure}[htbp]
    \centering
    \includegraphics[width=\textwidth]{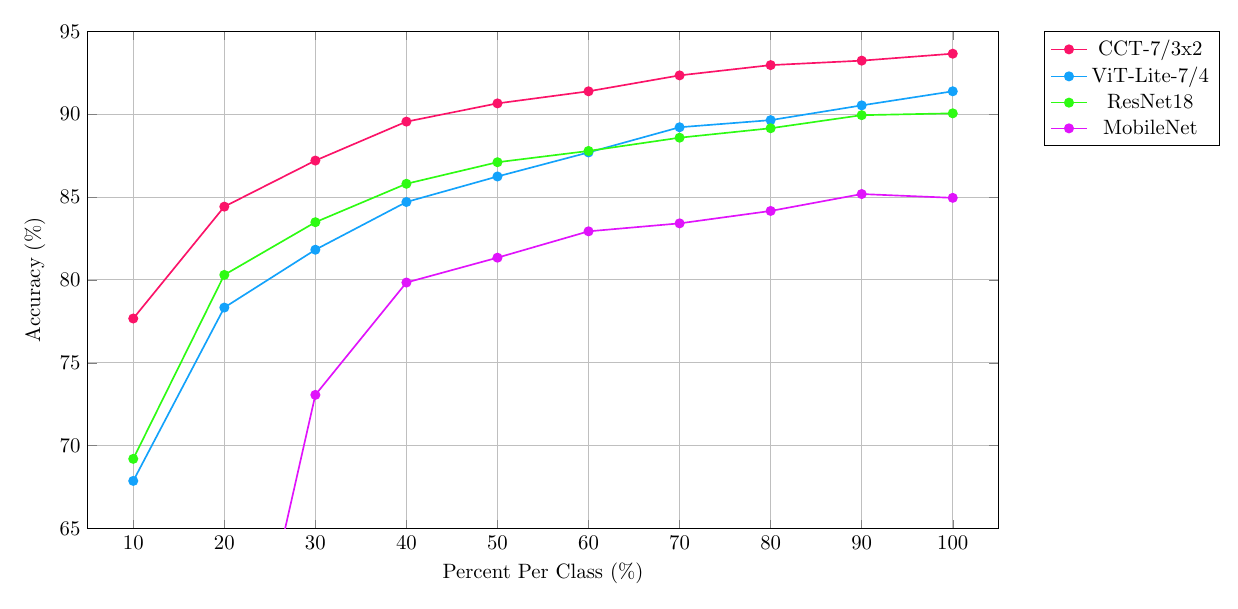}
    \caption[Accuracy of ViTs on Restricted Samples per Class]{%
        Comparison of models with restricted number of samples per class. At
        10\% models are trained on only 5000 images. Transformer based models
        demonstrate better scalability than ResNet based models.
    }\label{fig:escaping-samples}
\end{figure}

Moving on to further test the scalability of our model with respect to data, we 
study the performance with respect the number of samples as well as the size of
our images.
In \Cref{fig:escaping-samples} we restrict the number of samples in each class
within the CIFAR-10 dataset.
We compare the performance of CCT, ViT, ResNet18, and MobileNet when using only
$10\%$ of CIFAR-10 up to the full dataset.
With only $10\%$ of CIFAR-10, CCT is still able to achieve $77.7\%$ accuracy,
compared to ViT's $67.9\%$.
CCT is able to outperform the other models regardless of the data reduction.
ViT shows worse performance with data scaling, only beating ResNet18 when
including $70\%$ or more of the data.

\begin{figure}[htbp]
    \centering
    \begin{subfigure}{0.32\linewidth}
        \includegraphics[width=\textwidth]{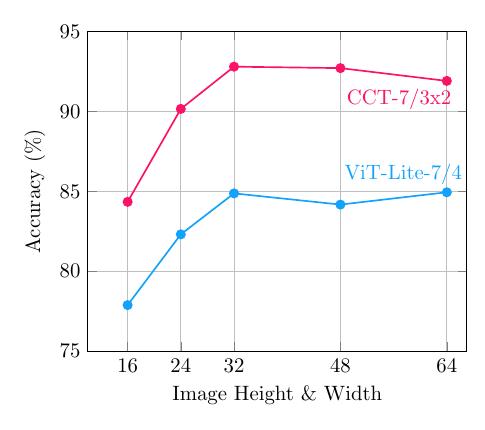}
    \end{subfigure}
    \begin{subfigure}{0.31\linewidth}
        \includegraphics[width=\textwidth]{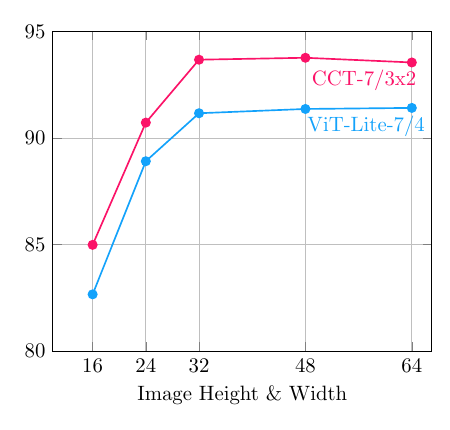}
    \end{subfigure}
    \begin{subfigure}{0.31\linewidth}
        \includegraphics[width=\textwidth]{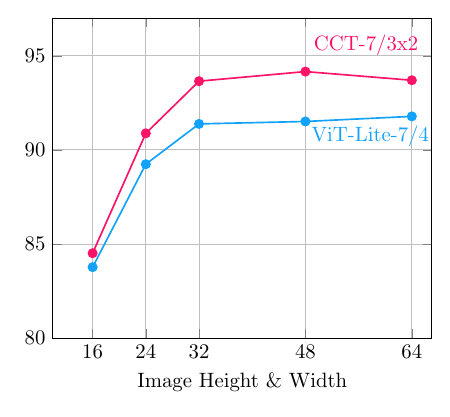}
    \end{subfigure}
    \begin{subfigure}{0.32\linewidth}
        \includegraphics[width=\textwidth]{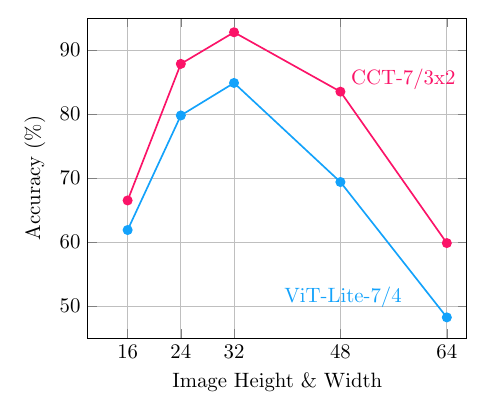}
        \caption{No P.E.%
        }\label{fig:escaping-nope}
    \end{subfigure}
    \begin{subfigure}{0.31\linewidth}
        \includegraphics[width=\textwidth]{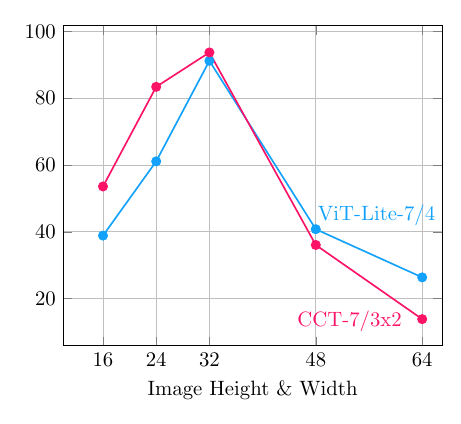}
        \caption{Sinusoidal P.E.%
        }\label{fig:escaping-lpe}
    \end{subfigure}
    \begin{subfigure}{0.31\linewidth}
        \includegraphics[width=\textwidth]{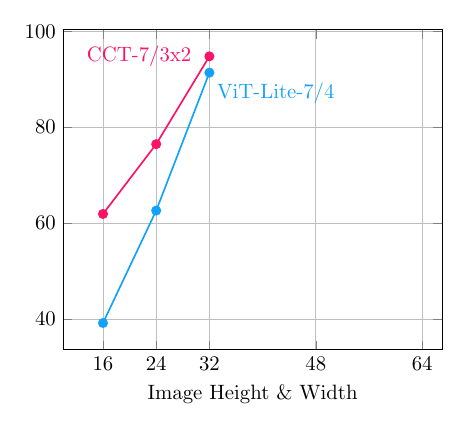}
        \caption{Learnable P.E.%
        }\label{fig:escaping-spe}
    \end{subfigure}
    \caption[Vision Transformer Resolution Based Performance]{%
        Comparison of ViT-Lite and CCT accuracy on CIFAR-10 with
        differing image resolutions. In first row, models are trained from
        scratch. In second row, models are inference and trained on $32\times32$
        images. \cref{fig:escaping-nope} is without
        positional embedding,
        \cref{fig:escaping-spe} with sinusoidal positional embedding,
        and \cref{fig:escaping-lpe} with a learnable positional embedding.
        Inference with learnable positional embedding cannot be extended to
        larger images without modifying model parameters.
    }\label{fig:escaping-hwcut}
\end{figure}

Additionally, we include a short study where we modify the image sizes of
CIFAR-10 to understand the dependence on resolution, found in
\Cref{fig:escaping-hwcut}.
With smaller resolution images models will likely be less able to rely upon
local structures within the data, as they will be merged.
When upscaling, we use a standard bicubic interpolation.
In the first row of the graphs we train our models from scratch, allowing them
to discover these associations.
In the second row, we only run inference, testing our models' capacity to
generalize to novel resolutions.
We also show comparisons without positional embedding, with Sinusoidal
Positional Embedding, and with Learnable Positional Embedding.
In our inference results Learnable Positional Embedding models are unable to 
process larger resolution images than they were trained on, creating a
significant limitation to this method.
In all cases, except inference with Sinusoidal Positional Embedding, CCT is 
able to out perform ViT, further demonstrating data generalizability.

\subsection{Natural Language Processing}\label{sec:escaping-nlp}
Finally, we test our method on small natural language processing datasets.
This network needs slight modification, incorporating
GloVe~\cite{pennington-etal-2014-glove} to provide word embeddings for our
model.
We do not train these embedding parameters and we do not include GloVe in our
model parameter sizes, which is about 20M.
To process the data we treat the text as single channel data, use an embedding
dimension of 300, and a convolution kernel of size 1.
We also perform masking in the typical manner.

By using CCT on these datasets we are able to achieve up to a $3\%$ improvement
when comparing to vanilla transformers.
Additionally, our CCT model is able to do this while using fewer parameters.
Our CCT models that are able to perform best have less than 1M parameters,
making GloVe a significantly larger part of the network.
We report a comparison of vanilla transformers, ViT, CVT, and CCT in
\Cref{tab:escaping-nlp}

\begin{table}[hbtp]
    \centering
    \resizebox{1.00\linewidth}{!}{%
    \begin{tabular}{l|ccccc|r}
    \toprule
    \noalign{\smallskip}
    \textbf{Model} & \textbf{AGNews} & \textbf{TREC} & \textbf{SST} & \textbf{IMDb} & \textbf{DBpedia} & \textbf{\# Params} \\
    \noalign{\smallskip}
    \hline
    \noalign{\smallskip}
    \multicolumn{7}{l}{\textit{Vanilla Transformer Encoders}}\\
    \noalign{\smallskip}
    \toprule
    \noalign{\smallskip}
    \textbf{Transformer-2} & $93.28\%$  & $90.40\%$  & $67.15\%$  & $86.01\%$  & $98.63\%$  & $1.086$ M \\
    \textbf{Transformer-4} & $93.25\%$  & $92.54\%$  & $65.20\%$  & $85.98\%$  & $96.91\%$  & $2.171$ M \\
    \textbf{Transformer-6} & $93.55\%$  & $92.78\%$  & $65.03\%$  & $85.87\%$  & $98.24\%$  & $4.337$ M \\
    \noalign{\smallskip}
    \toprule
    \noalign{\smallskip}
    \multicolumn{7}{l}{\textit{Vision Transformers (ViT)}}\\
    \noalign{\smallskip}
    \toprule
    \noalign{\smallskip}
    \textbf{ViT-Lite-2/1} & $93.02\%$  & $90.32\%$  & $67.66\%$  & $87.69\%$  & $98.99\%$  & $0.238$ M \\
    \textbf{ViT-Lite-2/2} & $92.20\%$  & $90.12\%$  & $64.44\%$  & $87.39\%$  & $98.88\%$  & $0.276$ M \\
    \textbf{ViT-Lite-2/4} & $90.53\%$  & $90.00\%$  & $62.37\%$  & $86.17\%$  & $98.72\%$  & $0.353$ M \\
    \textbf{ViT-Lite-4/1} & $93.48\%$  & $91.50\%$  & $66.81\%$  & $87.38\%$  & $99.04\%$  & $0.436$ M \\
    \textbf{ViT-Lite-4/2} & $92.06\%$  & $90.42\%$  & $63.75\%$  & $87.00\%$  & $98.92\%$  & $0.474$ M \\
    \textbf{ViT-Lite-4/4} & $90.93\%$  & $89.30\%$  & $60.83\%$  & $86.71\%$  & $98.81\%$  & $0.551$ M \\
    \textbf{ViT-Lite-6/1} & $93.07\%$  & $91.92\%$  & $64.95\%$  & $87.58\%$  & $99.02\%$  & $3.237$ M \\
    \textbf{ViT-Lite-6/2} & $92.56\%$  & $89.38\%$  & $62.78\%$  & $86.96\%$  & $98.89\%$  & $3.313$ M \\
    \textbf{ViT-Lite-6/4} & $91.12\%$  & $90.36\%$  & $60.97\%$  & $86.42\%$  & $98.72\%$  & $3.467$ M \\
    \noalign{\smallskip}
    \toprule
    \noalign{\smallskip}
    \multicolumn{7}{l}{\textit{Compact Vision Transformers (CVT)}}\\
    \noalign{\smallskip}
    \toprule
    \noalign{\smallskip}
    \textbf{CVT-2/1} & $93.24\%$  & $90.44\%$  & $67.88\%$  & $87.68\%$  & $98.98\%$  & $0.238$ M \\
    \textbf{CVT-2/2} & $92.29\%$  & $89.96\%$  & $64.26\%$  & $86.99\%$  & $98.93\%$  & $0.276$ M \\
    \textbf{CVT-2/4} & $91.10\%$  & $89.84\%$  & $62.22\%$  & $86.39\%$  & $98.75\%$  & $0.353$ M \\
    \textbf{CVT-4/1} & $93.53\%$  & $92.58\%$  & $66.64\%$  & $87.27\%$  & $99.04\%$  & $0.436$ M \\
    \textbf{CVT-4/2} & $92.35\%$  & $90.36\%$  & $63.90\%$  & $86.96\%$  & $98.93\%$  & $0.474$ M \\
    \textbf{CVT-4/4} & $90.71\%$  & $90.14\%$  & $61.98\%$  & $86.77\%$  & $98.80\%$  & $0.551$ M \\
    \textbf{CVT-6/1} & $93.38\%$  & $92.06\%$  & $65.94\%$  & $86.78\%$  & $99.02\%$  & $3.237$ M \\
    \textbf{CVT-6/2} & $92.57\%$  & $91.14\%$  & $64.57\%$  & $86.61\%$  & $98.86\%$  & $3.313$ M \\
    \textbf{CVT-6/4} & $91.35\%$  & $91.66\%$  & $61.63\%$  & $86.13\%$  & $98.76\%$  & $3.467$ M \\
    \noalign{\smallskip}
    \toprule
    \noalign{\smallskip}
    \multicolumn{7}{l}{\textit{Compact Convolutional Transformers (CCT)}}\\
    \noalign{\smallskip}
    \toprule
    \noalign{\smallskip}
    \textbf{CCT-2/1x1} & $93.40\%$  & $90.86\%$  & $\textbf{68.76\%}$ & $88.95\%$  & $99.01\%$  & $0.238$ M \\
    \textbf{CCT-2/2x1} & $93.38\%$  & $91.86\%$  & $67.19\%$  & $\textbf{89.13\%}$ & $99.04\%$  & $0.276$ M \\
    \textbf{CCT-2/4x1} & $\textbf{93.80\%}$ & $91.42\%$  & $64.47\%$  & $88.92\%$  & $99.04\%$  & $0.353$ M \\
    \textbf{CCT-4/1x1} & $93.49\%$  & $91.84\%$  & $68.21\%$  & $88.71$\%  & $99.03$\%  & $0.436$ M \\
    \textbf{CCT-4/2x1} & $93.30\%$  & $\textbf{93.54\%}$ & $66.42\%$  & $88.94\%$ & $\textbf{99.05\%}$ & $0.474$ M \\
    \textbf{CCT-4/4x1} & $93.09\%$  & $93.20\%$  & $66.57\%$  & $88.86\%$  & $99.02\%$  & $0.551$ M \\
    \textbf{CCT-6/1x1} & $93.73\%$  & $91.22\%$  & $66.59\%$  & $88.81\%$  & $98.99\%$  & $3.237$ M \\
    \textbf{CCT-6/2x1} & $93.29\%$  & $92.10\%$  & $65.02\%$  & $88.74\%$  & $99.02\%$  & $3.313$ M \\
    \textbf{CCT-6/4x1} & $92.86\%$  & $92.96\%$  & $65.84\%$  & $88.68\%$  & $99.02\%$  & $3.467$ M \\
    \hline
\end{tabular}

    }
    \caption[CCT Text Classification]{Top-1 validation accuracy on text classification datasets. The 
    number of parameters does not include the word embedding layer, because we 
    use pretrained word-embeddings and freeze those layers while training.
    }\label{tab:escaping-nlp}
\end{table}

\section{Conclusion}\label{sec:escaping-conclusion}
In this work we saw the importance of properly embedding information into our
machine learning models.
We need to ensure that this is done properly or we may severely limit our
model's capabilities.
Even small seemingly trivial differences can have tremendous effects on these
models, making it important to care when designing our neural architectures.
If great care is not taken we will make the wrong conclusions and hinder our own
progress.

While pretraining can help with model performance, when working with very
large datasets it becomes difficult to deduplicate data, and works have shown
that despite attempts to deduplicate these datasets may still be reduced upwards
of $50\%$~\cite{abbas2023semdedup}.
These duplications reduce model performance and generalizability, as they push
the models to over attend to certain semantics.
While reducing the requisite dataset size doesn't solve this problem, it
certainly makes it a much more tractable problem.
Given such results it makes it difficult to distinguish if large pretrained models
are generalizing or simply memorizing data.

An important result of this work was the ability to achieve comparable
performance while using orders of magnitude fewer parameters.
While there are still a large number of parameters, having fewer decreases a
model's ability to overfit.
Smaller models also enable them to be used by more people, with fewer
computational resources, and in more domains.
Despite the rapid advancement of computational power, such small models are
still critical tools for many areas of science, which may not have access to
multiple GPUs or the ability to obtain large datasets.
While datasets like CIFAR-10 are considered to be small by machine learning
standards, they are often orders of magnitude larger than datasets available
within other research domains.
This work makes transformer models available to these researchers.

\chapter{Variadic Neighborhood Attention}\label{ch:stylenat}
\chapterquote{
    Random numbers should not be generated with a method chosen at random.
}{Donald Knuth}
\textbf{Nota Bene:}
This chapter is based on the previous published co-authored work \emph{Efficient
Image Generation with Variadic Attention Heads}~\cite{WaltonStyleNAT2025CVPR},
formerly released as \emph{StyleNAT: Giving Each Head a New Perspective}.
Additionally, this chapter involves content from \emph{Neighborhood Attention
Transformer}~\cite{Hassani_2023_CVPR} (NAT) in order to facilitate the
discussion of StyleNAT, but is not the focus of this chapter.
\begin{itemize}
    \item Steven Walton programmed the majority of the source code for StyleNAT 
        and ran the majority of experiments. This includes creating all the 
        research questions and designing all the necessary experiments to 
        evidence them. His contributions also include all the visual analysis 
        as well as the development of the attention maps to visualize 
        restricted attention mechanisms. He was also the main writer of the 
        paper. Steven also made significant contributions to the work of NAT, 
        helping develop the theory (primarily around generalization), made 
        contributions to the source code, provided advice, and help write the 
        paper.
    \item Ali Hassani developed the \natten CUDA kernel that was used in both 
        StyleNAT and NAT. He provided important insights, especially with the 
        rapidly changing \natten code, made contributions to the source code, 
        helped perform experiments, and provided key insights for the development 
        of the restricted attention visualization. Ali Hassani was also the 
        primary author of the NAT paper, writing the majority of code, 
        performing the majority of experiments, and was the largest contributor 
        to the paper's text.
    \item Xingqian Xu contributed advice and insights around the underlying 
        StyleGAN architecture.
    \item Zhangyang Wang provided guidance during the research and feedback for 
        the project.
    \item Jaichen Li provided feedback for the NAT design and contributed to 
        the writing of the paper.
    \item Shen Li provided general design feedback for the \natten CUDA kernel 
        and support for running the large scale experiments.
    \item Humphrey Shi was the advisor for both StyleNAT and NAT, 
        contributing overall guidance on the research as well as funding for 
        both works. Humphrey also contributed to the writing of the paper and 
        ensuring research stayed on track.
\end{itemize}

While \Cref{ch:escaping}'s success with CCT demonstrated that ViTs could be
significantly improved in terms of data and computational efficiency, it left
the core neural
architecture untouched.
These impacts come from preparing the data for processing, but further
improvements can be made by also improving the processing.
Our ViT models still struggle with their $O(n^2)$ complexities, in both time and
space, so making improvements to these layers can have significant impacts.
Still, the work showed that transformers did not need big data nor big models to
be successful.
This motivates further work into improving these architectures themselves.

Transformers were born with language in mind, but had been adapted for vision.
The computational challenges are particularly challenging in Computer Vision 
due to the multi-dimensional data that must be processed, $c\times w\times h$ 
which frequently leads to out-of-memory (OOM)
issues~\cite{NEURIPS2021_98dce83d,lee2022vitgan,pmlr-v97-zhang19d}.
The \emph{de-facto} solution to this problem had been to use Convolutional Neural
Networks (CNNs)\cite{lecun2002gradient,NIPS1989_53c3bce6,NIPS2014_f033ed80}.
This is because CNNs provide memory efficiency by operating only on a localized 
context window as well as naturally incorporating multi-dimensional spatial 
relationships.

On the other hand, transformer networks attend over the entire data, allowing
for arbitrary connections to be made.
As previously discussed (\cref{ch:escaping-vit}), transformers are capable of
learning convolution filters, so it should be possible for them to be just as
powerful.
These benefits come at a cost of $\mathcal{O}(n^2)$ both in computational
complexity as well as memory complexity, but our previous work demonstrated that
smaller ViTs could outperform CNNs.
This then begs the question if ViTs can be better adapted to vision tasks.
Are we able to achieve $\mathcal{O}(n)$ performance while also being able to
incorporate both local and global structures within our data?

\begin{figure}[htpb]
    \centering
    \includegraphics[width=\linewidth]{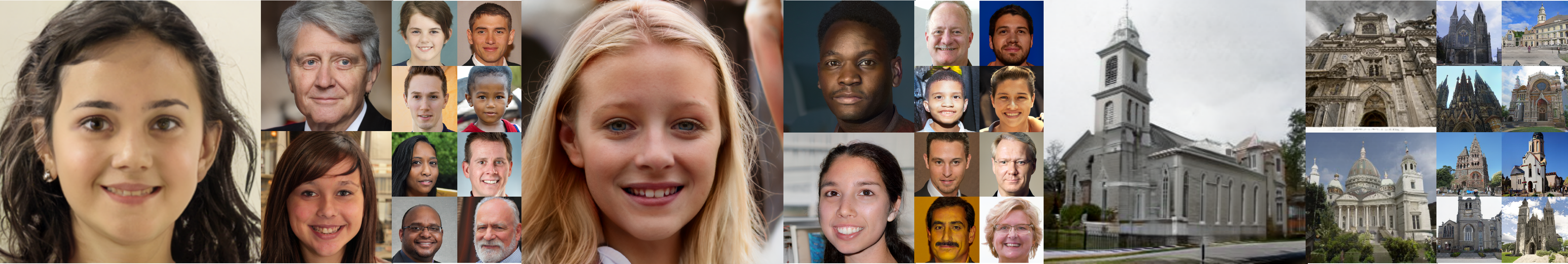}
    \caption[StyleNAT Samples (FFHQ-256, FFHQ-1024, LSUN Church)]{%
    Samples form FFHQ-256 (left) with FID: 2.05, FFHQ-1024 (center) with FID:
    4.17, and Church (right) with FID: 3.40 generated by our StyleNAT network,
    using Hydra Neighborhood Attention.
    }\label{fig:stylenat-header}
\end{figure}

This chapter studies the core architecture of the network, by introducing
\emph{Efficient Image Generation with Variadic Attention
Heads}~\cite{WaltonStyleNAT2025CVPR}, which allows the vision transformer to do
more with less.
The primary modification for this work is simple, yet powerful: allow attention
heads to attend to independent receptive fields.
Our results demonstrate that some simple modifications to our attention heads
can allow our Vision Transformers to better integrate local and global
relationships during image generation.
The result of this is the ability to train a StyleGAN~\cite{Karras_2019_CVPR}
based model, using a modified version of Neighborhood
Attention~\cite{Hassani_2023_CVPR,hassani2023dilatedneighborhoodattentiontransformer}, 
which pushes the Pareto Frontier for image generation on FFHQ-256.
Our model makes significant improvements in terms of visual fidelity while being
smaller and has a higher throughput than other comparative models.

\section{Localized Attention}\label{ch:stylenat-local-attn}
In an effort to address the computational challenges of transformers, 
researchers looked to a number of different solutions.
One such solution is to only perform attention on some localized region instead
of the whole input.
This formulation is natural as analysis of attention maps shows that there is
strong correlation between neighboring 
tokens~\cite{kovaleva-etal-2019-revealing,bahdanau2016neuralmachinetranslationjointly,Viti_2025_WACV},
or having \emph{Attention Sinks}~\cite{xiao2024efficientstreaminglanguagemodels}.
Works like Image Transformer~\cite{parmar2018image} and Stand Alone
Self-Attention (SASA)~\cite{NEURIPS2019_3416a75f} use localized context
windows for their transformer algorithms, similar to the ideas
proposed in Longformer~\cite{beltagy2020longformer}.
These methods reduced the computational burden of attention mechanisms,
\emph{approximating} $\mathcal{O}(n)$ complexity, but had issues generalizing 
as the window size increased.
Other works like HaloNet~\cite{vaswani2021scalinglocalselfattentionparameter} 
and the Window Self-Attention (WSA) from Swin
Transformer~\cite{liu2021swin, Liu_2022_CVPR} partitioned the query and context 
sets, independently performing self-attention.
These blocks become highly parallelizeable but does not account for cross-block
interactions.
Swin tried to address this issue by introducing shifted windows (SWSA), where
subsequent attentions would shift their windows.
With a hierarchical structure the network can is able to
attend to every pixel in an image to attend to one another, but incorporates 
biases around boundaries, similar to the issues faced in the non-overlapping 
blocks of ViT (\Cref{ch:escaping}).

\section{Neighborhood Attention}
\begin{figure}[htbp]
    \centering
    \includegraphics[width=\textwidth]{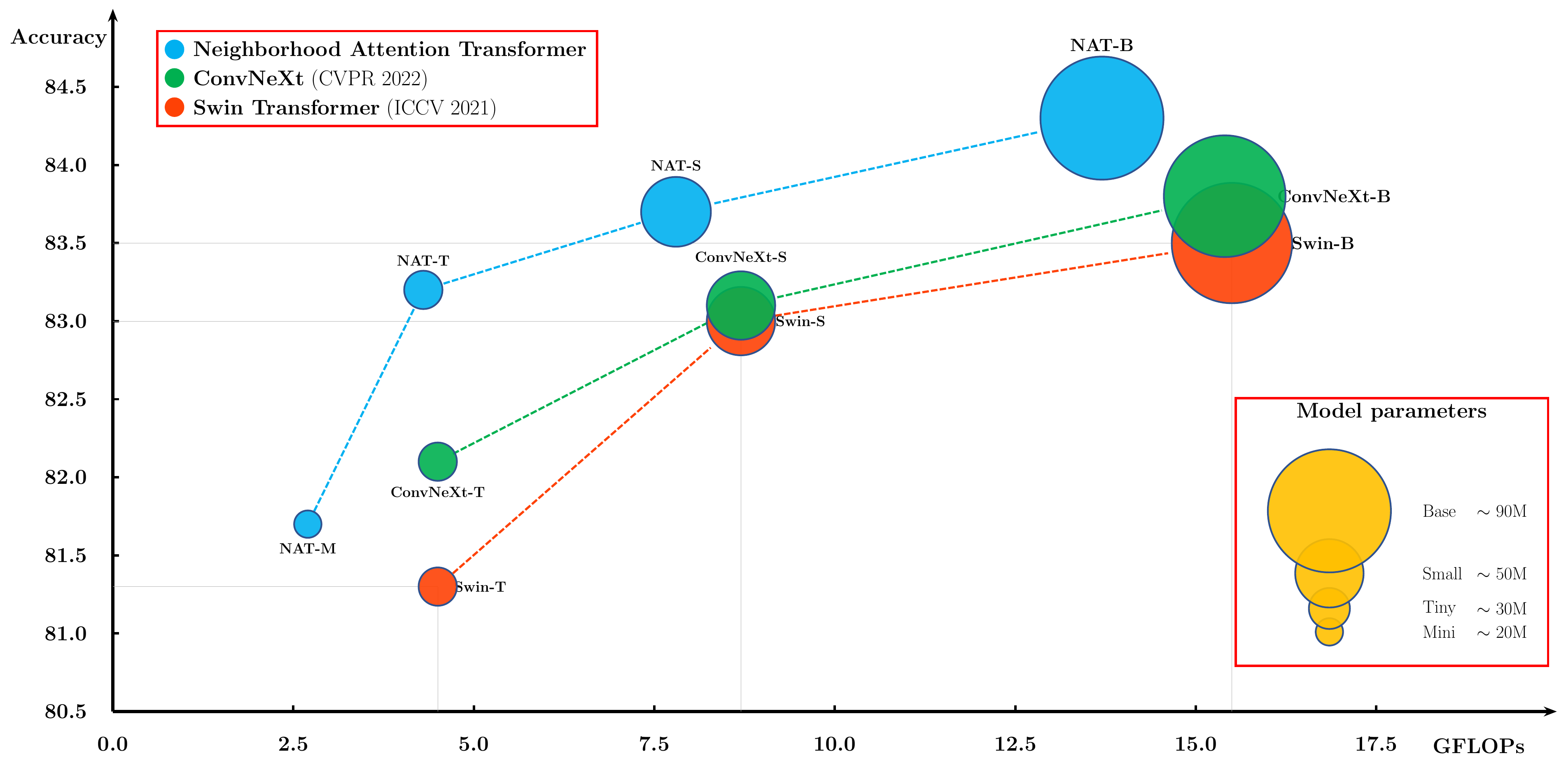}
    \caption[NAT vs Swin Vs ConvNext ImageNet Performance]{Comparison of Neighborhood Attention, Swin, and ConvNeXt on
    ImageNet classification.%
    }\label{fig:na-classification}
\end{figure}

To resolve these issues, Hassani~\etal developed the Neighborhood Attention
Transformer (NAT)~\cite{Hassani_2023_CVPR}.
The architecture is similar to SASA but resolved the generalization issue,
ensuring that when the window size was equal to the image size that Neighborhood
Attention (NA) would be identical to the traditional dot-product self-attention 
mechanism.
Like a convolution, NA considers a context window around each individual input
\emph{queries}, $Q$.
The \emph{keys}, $K$, then evaluate over the surrounding neighborhood (a square).
If a (relative) positional bias~\cite{Hu_2019_ICCV,JMLR:v21:20-074}, $B$, is used then this must also be
modified to account for the key location.
Similarly, the \emph{value}, $V$, must be updated to correspond with the local
neighborhood.
We can describe this attention variant as follows:

\begin{equation}
    \textbf{A}_i^k 
    = \begin{bmatrix}
        Q_i K^T_{\rho_1(i)} + B_{i,\rho_1(i)}\\
            \vdots\\
        Q_i K^T_{\rho_k(i)} + B_{i,\rho_k(i)}\\
    \end{bmatrix}
    \qquad\qquad\qquad
    \textbf{V}_i^k 
        = \begin{bmatrix}
            V_{\rho_1(i)}^T\\
                \vdots\\
            V_{\rho_k(i)}^T
    \end{bmatrix}
    \qquad\quad
    \label{eqn:NA-AV}
\end{equation}
For an input token, $i$, we consider a window of size $k$ and its neighborhood,
$\rho$.
Specifically, $\rho_j(i)$ denotes the $i$'s $j^\text{th}$ nearest neighbor.
We can then consider the full attention about a token as:

\begin{equation}
    \text{NA}_k(i) =
    \text{Softmax}
        \left(%
           \frac{\textbf{A}_i^k}{\sqrt{d}}%
        \right)%
    \textbf{V}_i^k
    \label{eqn:NA}
\end{equation}

Testing Neighborhood Attention in discriminative settings showed that it was
able to outperform other attention variants, such as Swin, as well as modernized
Convolutional variants such as ConvNext~\cite{liu2022convnet}, show in
\Cref{fig:na-classification}.
Importantly, for a fixed number of parameters, NA outperformed others on
accuracy, memory, and flops, but did not on throughput.
Further development of NA led to a Dilated variant
(DiNA)~\cite{hassani2023dilatedneighborhoodattentiontransformer} and improving
the GPU
kernel~\cite{hassani2023nadrsa,hassani2024fasterneighborhoodattentionreducing}
and generalizing the
architecture~\cite{hassani2025generalizedneighborhoodattentionmultidimensional},
allowing for arbitrary attention configurations, similar to tools like Flash
Attention~\cite{dao2022flashattention,dao2023flashattention,shah2024flashattention}
and xFormers~\cite{xFormers2022}.
These improvements led to significant improvements in speed, with over a
$100\times$ improvement in the forward pass and $80\times$ for both forward and
backwards passes, at FP16 and on an NVIDIA A100 GPU.
These improvements led to further adoption of the model throughout other
works~\cite{Jain_2023_ICCV,Jain_2023_CVPR,dupont_augmented_2019}.

While localized attention provides significant advantages in reducing the 
computational load, they have limitations due to the restricted context window 
over which they attend to. 
This creates a similar to CNNs, gaining advantages of the localized structure of
the data at the cost of global structures.
Similar to CNNs, this can often be resolved by using a hierarchical model, where
downsampling and pooling allow for full token mixing.
For example, NAT takes in an input image sized $\mathbb{R}^{H\times W}$, then
embeds this into $\mathbb{R}^{\frac{H}{4}\times\frac{W}{4}}$ through an 
overlapping tokenizer, which uses the same embedding process as 
CCT~\cite{hassani2022escapingbigdataparadigm}.
There are 3 transformer blocks before another overlapping downsampling is
performed and another 4 transformer layers process the image at
$\mathbb{R}^{\frac{H}{8}\times\frac{W}{8}}$.
This process repeats with 18 layers at
$\mathbb{R}^{\frac{H}{16}\times\frac{W}{16}}$ and 5 layers at
$\mathbb{R}^{\frac{H}{32}\times\frac{W}{32}}$.
\Cref{fig:stylenat-nat} depicts the architecture.
This formulation works especially well for discriminative tasks, due to the
network map's endomorphic formulation.
In the case of classification the network is learning the map
$f:\mathbb{R}^{C\times H\times W} \mapsto \mathbb{N}^0$.
Through this hierarchical mapping and overlapping downsampling all tokens become
sufficiently mixed and there is an assurance that all inter-relations can be
accounted for.

\begin{figure}[htbp]
    \centering
    \includegraphics[width=\textwidth]{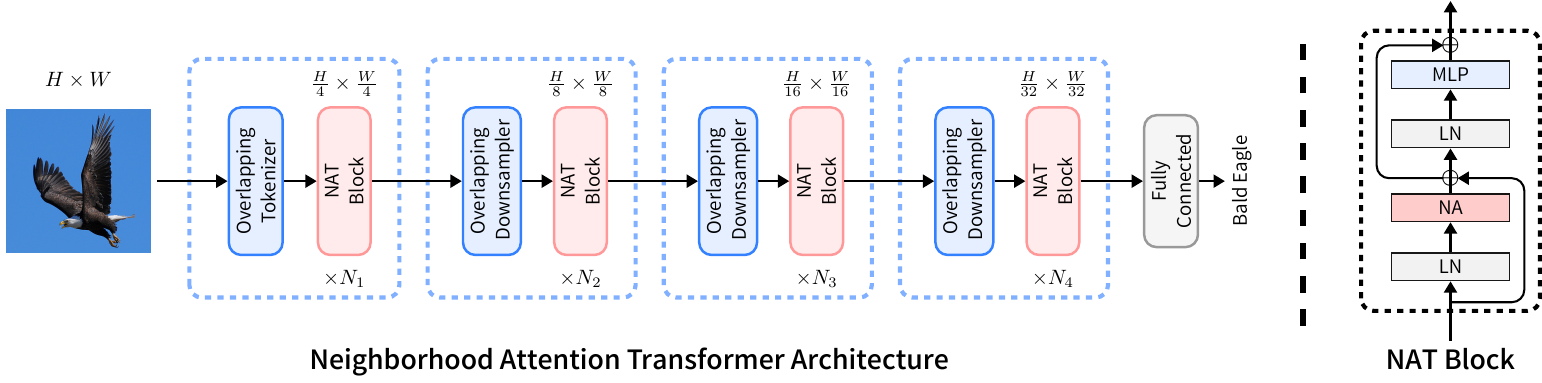}
    \caption[Neighborhood Attention Transformer (NAT)]{%
        Diagram depicting the Neighborhood Attention
        Transformer~\cite{Hassani_2023_CVPR} as applied to vision
        classification.
        }\label{fig:stylenat-nat}
\end{figure}

\section{Variadic Attention Heads}\label{ch:stylenat-vah}
A subtle feature of multi-headed attention~\cite{vaswani2017attention} is that
attention heads are independent of one another.
This property allows each head to attend to different features within the data,
analogous to feature maps in CNNs.
This feature plays a key role in the performance of attention
models~\cite{NEURIPS2019_2c601ad9,tang-etal-2018-analysis}, 
with a few specialized attention heads being
the primary drivers~\cite{voita-etal-2019-analyzing}.
We thus propose the following hypothesis:
\emph{Decoupling attention heads will improve the performance of Neighborhood 
Attention.}

An important feature of Neighborhood Attention is that it generalizes to
standard attention and allows for dilated receptive
fields~\cite{hassani2023dilatedneighborhoodattentiontransformer}.
With this in mind, we are able to reduce the locality bias imparted by local
receptive fields.
By allowing attention heads to attend to different receptive fields we allow for
the intermixing of global and local information, similar to the standard
attention.
Thus, we modify the standard Neighborhood Attention mechanism to allow
attention heads to have independent kernel sizes and dilations.
This modification can be explicitly represented as follows:

\begin{equation}
    \textbf{A}_{i,h}^k
    = \begin{bmatrix}
        Q_{i,h(k,d)} K^T_{\rho_1(i),h(k,d)} + B_{i,\rho_1(i),h(k,d)}\\
            \vdots\\
        Q_{i,h(k,d)} K^T_{\rho_k(i),h(k,d)} + B_{i,\rho_k(i),h(k,d)}\\
    \end{bmatrix}
    \qquad\qquad\qquad
    \textbf{V}_{i,h}^k
        = \begin{bmatrix}
            V_{\rho_1(i),h(k,d)}^T\\
                \vdots\\
            V_{\rho_k(i),h(k,d)}^T
    \end{bmatrix}
    \qquad\quad
    \label{eqn:HydraNA-AV}
\end{equation}
With this variation we specify that each query, $Q$, and key, $K$, are 
independent calculations, split across each attention head, $h$. 
Where the head is a function of the window size, $k$ and dilation, $d$.
The positional bias may also need be offset in a head-wise fashion.

\begin{equation}
    \text{NA}_{k(i)} = \text{Softmax}%
        \left(%
            \frac{%
                \textbf{A}^k_{i,h}%
            }{%
                \sqrt{d}%
            }%
        \right)%
        \textbf{V}^k_{i,h}
        \label{eqn:HydraNA}
\end{equation}

This formulation maintains the computational and memory advantages of
Neighborhood Attention, but mitigates the losses to architectural flexibility.
Within this formulation \emph{arbitrary combinations of window sizes and 
dilations may be used}, allowing for higher flexibility and integration of
information.
Consequently, these hyper-parameters, $k$ and $d$, need not be fixed.

\section{Generating The Right Experiment}
\begin{figure}[htbp]
    \centering
    \includegraphics[width=\textwidth]{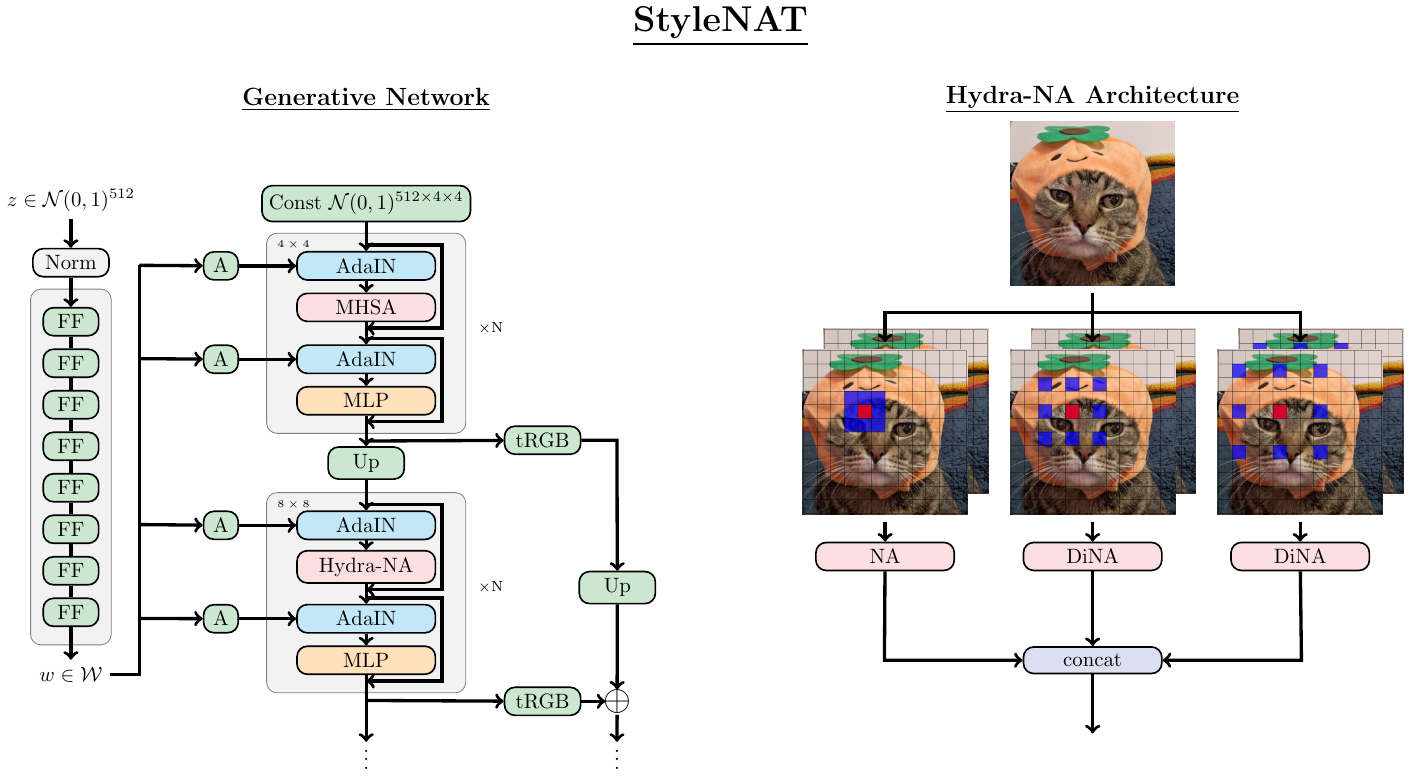}
    \caption{StyleNAT Architecture}
    \label{fig:stylenat-arch}
\end{figure}
To evidence our hypothesis presents a challenge, due to the nature of
most neural architectures accounting for these limitations and mixing data as 
dept increases.
With limited compute infrastructure there is significant pressure to design the
right experiment to properly test the research questions and ensure we limit any
observed effects to our procedure.
Most architecture changes are demonstrated through discriminative tasks; such as
classification, 
detection~\cite{lin2014microsoft,he2017mask,cai2018cascade}, and
segmentation~\cite{zhou2017scene}, as performed in our Neighborhood Attention 
papers.
Many of the small (e.g. CIFAR-10~\cite{4531741,krizhevsky2009learning}) and 
medium-sized (e.g. ImageNet-1k~\cite{deng2009imagenet}) datasets,
which would be within our computational budget, are nearly saturated; with many 
results similar to the labeling error rates in the data.
Consequently, improvements tend to me minor, often with only a percent or lower
difference.
This makes it difficult to evaluate the effect of the modifications, even with
exhaustive search, as the performance gains become difficult to distinguish from
many other factors.
Their hierarchical nature also makes it difficult to isolate our contribution as
this already performs local-global token mixing.
To combat these issues while remaining mindful of computational budgets, we 
must intentionally design an experiment to limit the variables of interest and 
ensure proper variable isolation. 

While discriminating tasks often learn a mapping from some
$\mathbb{R}^m\mapsto\mathbb{R}^n$ where $n<m$ it is common for generative tasks
to learn maps where $n\geq m$, or even a map onto itself
(\Cref{ch:bg-data-maps}).
By placing focus on these formulations we can better isolate our variables of
interest.
Specifically, the 
StyleGAN~\cite{Karras_2019_CVPR,Karras_2020_CVPR,NEURIPS2020_8d30aa96,NEURIPS2021_076ccd93,lee2022vitgan}
architecture uses a progressive~\cite{karras2018progressive} structure,
generating an image starting from a small noise sample, relying on the 
Latent Manifold Hypothesis, making the assumption that the necessary
information for the generation of samples is smaller than the dimensionality of
the images themselves.
This progressive nature is designed for the local receptive fields of CNNs,
with low resolution images capturing more global structures and as the
resolution grows the neural net can learn more appropriate upsampling methods
that account for finer detail synthesis.
This is effectively in reverse to the structure of hierarchical classification
models, benefiting locality.

This structure also has the benefit that, in general, it synthesizes an image at
progressive image resolutions, allowing for greater computational efficiencies.
Unfortunately, due to these biases the architectures commonly struggle with long
range fine detail synthesis, commonly resulting in features like heterochromia
(eyes differing in color) when doing human face synthesis.
Such fine grain detail may not exist at low resolutions and only appear later
when the image resolution is significantly larger than the convolution window.
Thus, locality is a double edged sword for these architectures: providing high
utility at low resolutions but becoming detrimental as the image grows.
The prolific nature of these architectures also yields a large number of 
comparitors.
The well studied nature of GANs allows for better isolation and allows us to
assume that these architectures are reasonably optimized.
This setting makes it a great platform for variable isolation and allowing us to
test our hypothesis.

\begin{figure}[htbp]
    \centering
    \includegraphics[width=\textwidth]{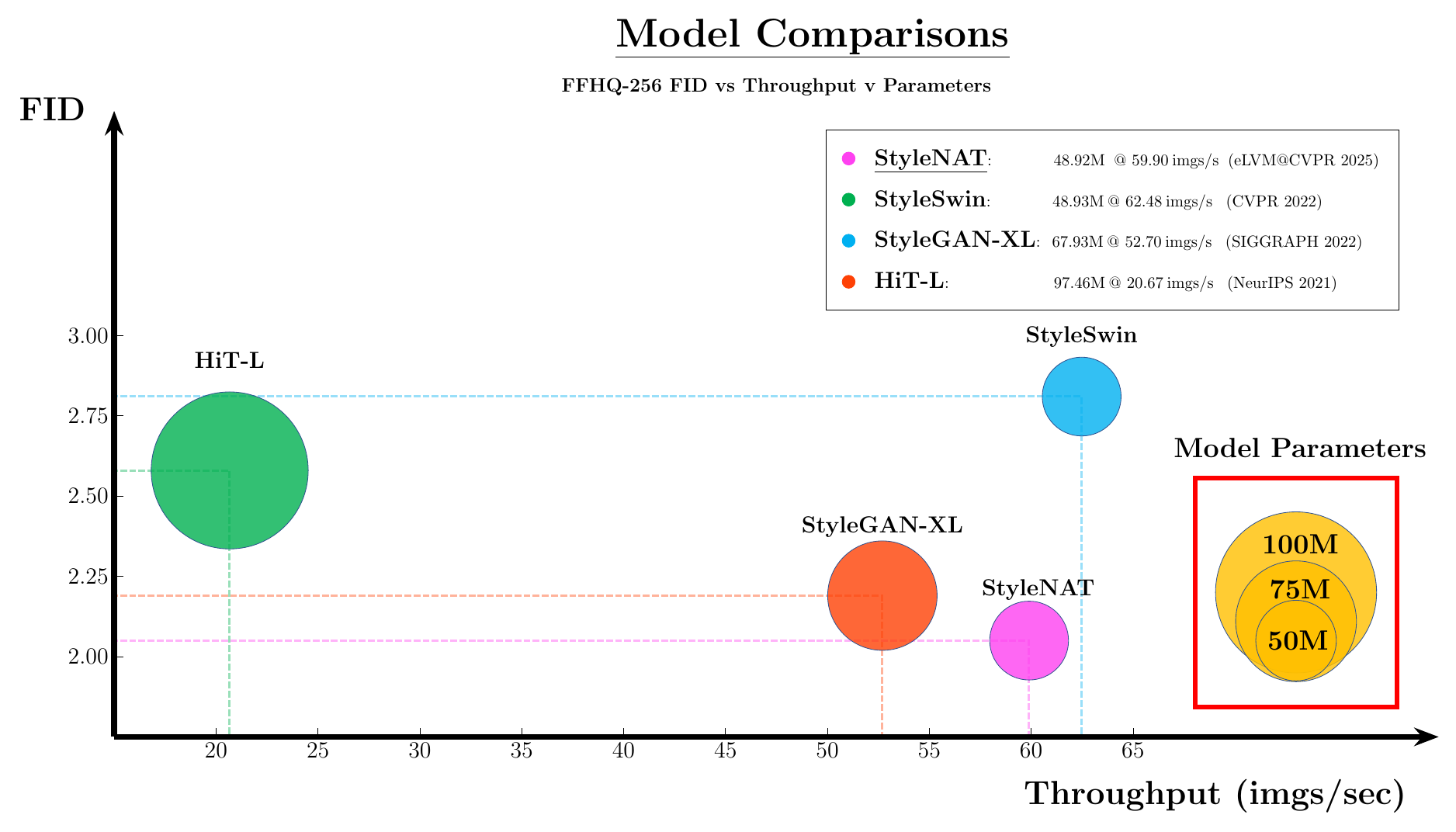}
    \caption[StyleNAT: FID vs. Throughput vs. Parameters]{%
    StyleNAT represents the Pareto Frontier for FID (y-axis), 
    Parameters (bubble size), and throughput (x-axis) on FFHQ-256.
    StyleNAT has the lowest FID of 2.05, with fewer parameters (48.92M) than
    similarly performing models and is capable of generating images in 
    real-time (59.90 imgs/s). This makes StyleNAT smaller, faster, and better
    than its competitors. (Note: StyleNAT's NATTEN kernel is not optimized and
    may still be improved for higher throughputs.)
    }\label{fig:stylenat-pareto}
\end{figure}

Additionally, there were many attempts to integrate attention and 
transformers~\cite{takida2024san,pmlr-v97-zhang19d,hudson2021ganformer,hudson2021ganformer2}
to resolve the issues in long range detail generation.
These works also needed to ensure that methods were computationally efficient,
as performing full attention across the entire image is computationally
intractable at high resolutions.
Specifically, the work Hit-GAN~\cite{NEURIPS2021_98dce83d} and 
StyleSwin~\cite{Zhang_2022_CVPR}had created similar formulations, allowing
for clearer experimental control.
Both these networks base their architectures off of the StyleGAN architecture.
Additionally, HiT-GAN and ViT-GAN~\cite{lee2022vitgan} found that transformers
struggled with the generation of images above $64\times64$ resolution, allowing
our work to demonstrate its capabilities at the much more common, and
computationally reasonable, scale of $256\times256$.

HiT-GAN divided their generative architecture into two sub-networks, one for
low-resolution ($\leq 32\times32$) and the other for high-resolution.
To generate spatial mixing in the low-resolution network extended the method of 
Nested Hierarchical Transformer~\cite{zhang2022nested}, dividing the input into 
non-overlapping patches, applying positional encoding, and bifurcating the
attention heads such that each operates on a different spatial axis (height and
width).
They note that they make an assumption that proper spatial mixing in
low-resolution stages will allow for high resolution stages to focus on image
synthesis, replacing their attention modules with MLPs.
Additionally, they replaced the common AdaIn~\cite{Huang_2017_ICCV} and
modulated layers~\cite{Karras_2020_CVPR} with cross-attention due to excessive 
memory issues and being unable to produce images above $64\times64$ resolution.

StyleSwin takes a different approach, much more closely following the original
StyleGAN architecture.
The typical Swin Transformer~\cite{liu2021swin,Liu_2022_CVPR} architecture
performs spatial mixing by using non-overlapping patches, performing a cyclic 
permutation on subsequent layers.
Through multiple layers this ensures there is spatial mixing, allowing for some
long range attention across edges.
StyleSwin instead splits their attention heads such that half utilize the window
patches and the other half utilizes the shifted windows, removing the necessity
of communication across layers for spatial mixing.
Otherwise the architecture closely follows that of
StyleGAN-2~\cite{Karras_2020_CVPR,NEURIPS2020_8d30aa96}, using 2 synthesis
layers per resolution level, incorporating
wavelets~\cite{gal2021swagan}, TTUR~\cite{NIPS2017_8a1d6947}, and
bCR~\cite{zhao2021improved}.
For their attention they use an attention dimension equivalent to the number of
channels in StyleGAN convolutions.
They divide this by 32 to get the number of attention heads, setting a minimum
number of heads to 4.
This results in 16 heads for resolutions 4 - 64, 8 heads at 128 and 4 at
256.
Additionally, a learnable parameter size $512\times4\times4$ is used to seed 
the synthesis network.
StyleSwin faced ``blocking artifacts'' and similar to HiT-GAN found that
self-attention could be removed at higher resolutions but that it failed to
model high-frequency details.

With the progressively growing structure we have a good platform that can
clarify the impact of our architectural changes.
The high rates of visual artifacts also meets the goals that this architecture
is designed to mitigate.
Furthermore, HiT-GAN and StyleSwin provide strong control baselines that can be
used to isolate our variables of interest.
Given this setting, we utilize StyleSwin as our experimental platform.

Due to computational budget limitations, we must minimize experimentation and
datasets to the most impactful.
This model has a large number of potential configurations, as our kernels can
range from a size of $3$ to the nearest odd integer smaller than the resolution,
$\mathcal{R}$.
Fixing the hyper-parameters determining window size and dilation reduces the
computational search space, helping stability, which is a common problem with
GANs.
Fixed kernel sizes and dilations can also help increase interpretability,
allowing for more direct visual analysis (\Cref{ch:stylenat-amaps}).
This will also help us better attribute effects of the context mixing, making it
a more direct comparison to StyleSwin.
For simplicity we assume an image is square, thus, for a given resolution, 
$\mathcal{R}$, we will have $\frac{\mathcal{R}}{2}-1$ potential kernels.
Each dilation must also be positive and while the dilation can make the
effective kernel size larger than the image, this results in attending to no
information.
Using this restriction results in
$\frac{\mathcal{R}}{4}$ kernels that can have dilations, and the max dilation
for a given kernel is
$\left\lfloor\frac{\mathcal{R}}{k}\right\rfloor$.
The total number of configuration of attention heads, per resolution, is:
\begin{align}\label{eq:num_configs}
    N_c &= \sum_{i=1}^{\mathcal{R}/2-1} 
                \left\lfloor
                    \frac{\mathcal{R}}{2i + 1}
                 \right\rfloor
                 \\
        \label{eq:simp_num_configs}
        &= \frac{\mathcal{R}}{4} + 
                \sum_{i=1}^{\mathcal{R}/4-1}
                \left\lfloor
                     \frac{\mathcal{R}}{2i+1}
                 \right\rfloor
\end{align}
Following the StyleSwin architecture, for an image size 
$3\times256\times256$ ($\mathcal{R}=256$) this results in
$2\times\left((16\times(4+14+37+97))+(8\times237)+(4\times565) \right) = 13176$
possible configurations and exceeds 47k configurations for high resolution
generation ($\mathcal{R}=1024$).\footnote{Neither kernels nor dilations need be
square, further increasing the potential configurations}
Given our interpretation of the Bitter Lesson, this flexibility is promising to 
the architecture, but is not feasible to exhaustively test within a modest 
computational budget.
We leave such work for bigger labs and instead focus on evidencing the research
hypothesis.

\begin{table}[hbpt]
    \begin{subfigure}[b]{0.49\linewidth}
        \centering
        \begin{tabular}{c|c|c|c}
    \toprule
    Level & Kernel & Dilation & Dilated Size \\
    \hline
    $\phantom{111}4$ & - & \phantom{--}- & \phantom{--}- \\
    $\phantom{111}8$ & 7 & \phantom{11}1 & \phantom{11}7 \\
    $\phantom{11}16$ & 7 & \phantom{11}2 & \phantom{1}14\\
    $\phantom{11}32$ & 7 & \phantom{11}4 & \phantom{1}28 \\
    $\phantom{11}64$ & 7 & \phantom{11}8 & \phantom{1}56\\
    $\phantom{1}128$ & 7 & \phantom{1}16 & 112 \\
    $\phantom{1}256$ & 7 & \phantom{1}32 & 224 \\
    $\phantom{1}512$ & 7 & \phantom{1}64 & 448\\
    $1024$           & 7 & 128           & 896\\
    \bottomrule
\end{tabular}

        \caption{%
            Split Head configuration
        }\label{tab:stylenat-splithead} 
    \end{subfigure}
    \begin{subfigure}[b]{0.49\linewidth}
        \centering
        \begin{tabular}{c|c|l}
    \toprule
    Level & Kernel & Dilations \\
    \hline
    $\phantom{111}4$    & -  & -  \\
    $\phantom{111}8$    & 7 & 1 \\
    $\phantom{11}16$   & 7 & 1,2\\ 
    $\phantom{11}32$   & 7 & 1,2,4\\ 
    $\phantom{11}64$   & 7 & 1,2,4,8\\ 
    $\phantom{1}128$  & 7 & 1,2,4,8,16\\ 
    $\phantom{1}256$  & 7 & 1,2,4,8,16,32\\ 
    $\phantom{1}512$  & 7 & 1,2,4,8,16,32,64\\ 
    $1024$ & 7 & 1,2,4,8,16,32,64,128\\
    \bottomrule
\end{tabular}

        \caption{%
        Progressive configuration
        }\label{tab:stylenat-pyramidhead}
    \end{subfigure}
    \caption[StyleNAT Configurations]{%
    Configurations for different attention head configurations.
    \cref{tab:stylenat-splithead} shows our 2 headed configuration, used in
    FFHQ, where half the heads are dense kernels and half are sparse dilated. We
    show the dilations and their effective size.
    \cref{tab:stylenat-pyramidhead} shows the progressive head configuration
    that was used in some LSUN Church experiments.
    }\label{tab:stylenat-headconfig}
\end{table}

\subsection{Datasets}\label{ch:stylenat-datasets}
For our main dataset we use the Flickr-Face-HQ Dataset (FFHQ), introduced in 
the original StyleGAN work~\cite{Karras_2019_CVPR}.
FFHQ is widely used across generative modeling research, has a high resolution
($1024\times1024$) variant, and a wide diversity of faces and accessories (e.g.
glasses, hats, jewelry).
Importantly, by generating human faces we are better able to analyze the images
as human are naturally attuned for facial perception, being able to detect
subtle distinctions~\cite{humanFaceRecognition}.
For our second dataset, we use the Church Outdoor class from the Large Scale 
image database (LSUN)~\cite{yu15lsun}.
This dataset is relatively common and presents a significant challenging,
containing many features with straight lines, often not found in biological
objects.

\subsection{Hyperparameters}\label{ch:stylenat-hyperparams}
To ensure that we \emph{only} measure the effects of the architecture we follow
the same training procedure as StyleSwin: using TTUR, a discriminator learning
rate of $2\times10^{-4}$, bCR with $\lambda_{real}=\lambda_{fake}=10$, and $r_1$
regularization~\cite{pmlr-v80-mescheder18a} with $\gamma=10$, as well as model
hyper-parameters.
The only deviations we make is the iteration we begin our LR-decay, which is
experimentally driven, and we doubled the batch size in later experiments, to
64, finding we were able to maintain stability at this level.
This difference may be due to our usage of 80GB A100 GPUs as opposed to the 32GB
V100s that StyleSwin used, as our batch size required 36GB of VRAM.
We include all hyper-parameters in our GitHub
repository\footnote{https://github.com/SHI-Labs/StyleNAT} as well as save these
values to our model checkpoints, including our seed values, for reproducibility.
We do not perform hyper-parameter optimization, seeking to evaluate our model on
architectural effect rather than ultimate performance.
A minor architectural change is made such that at the lowest resolution,
$4\times4$ we utilize a standard Multi-Headed Attention (MHA) layer as opposed
to a Neighborhood Attention layer.
At this layer StyleSwin uses a $4\times4$ window size and NA is restricted to
odd window sizes.
Experimentally we found no difference when using a $3\times3$ window size, but
the MHA layer slightly increases the computational efficiency due to the
unoptimized performance of NA at the time.\footnote{\natten has undergone
significant optimizations since the time of these experiments.}
Elsewhere StyleSwin uses a window size of $8\times8$ while StyleNAT uses
$7\times7$, giving a slight context range advantage to StyleSwin.
\Cref{table:exp-maintable} shows a full comparison of our results, comparing
different generative models by generative performance, the number of model
parameters, and the rate at which they can generate images.
We determine parameter size from officially released model checkpoints, removing
any non-generative parameters such as discriminators, hyper-parameters, or
exponential moving
averages~\cite{polyak1992,izmailov2018averaging,He_2022_CVPR}.
Similarly, we gather all throughput measures, ensuring consistent GPU
architecture and versions of Python and 
PyTorch~\cite{paszke2019pytorchimperativestylehighperformance}.
To ensure throughput is properly calculated, we first warm-up the models to
ensure they are properly cached, generating 50 samples, and then generate an
additional 100 samples, which we find the average of.
For all GANs we use a batch size of 1 and for diffusion models we maximize the
batch size for available memory~\footnote{Doing otherwise results in
significantly decreased throughputs while GANs show little to no deviation.}.
We include our procedure in our public repository for additional transparency.
For evaluation we will primarially rely on the Fr\'{e}chet Inception Distance
(FID)~\cite{NIPS2017_8a1d6947}, but include more discussion and evaluation in
\Cref{ch:stylenat-amaps}.

\begin{table}[htpb]
    \centering
    \resizebox{0.98\linewidth}{!}{%
        \begin{tabular}{l|l|cc|c|cc}
    \toprule
    {\multirow{2}{*}{\textbf{Arch}}} &
    {\multirow{2}{*}{\textbf{Model}}} 
        & \multicolumn{2}{|c|} {\textbf{FFHQ FID} $\downarrow$} 
        & \textbf{Church}
        & \multicolumn{2}{|c}
        {\textbf{Usage Metrics (256)}}\\
            & & \textbf{256} & \textbf{1024} & \textbf{256} 
            & \textbf{img/s} & \textbf{Params (M)} \\ \hline
    {\multirow{5}{*}{Convolution}}
        &StyleGAN2~\cite{Karras_2020_CVPR} & 3.83 & 2.84 & 3.86 
            & 84.85 & 30.03 \\
        &StyleGAN3-T~\cite{NEURIPS2021_076ccd93} & - & 2.70 & - 
            & 108.84$^\star$ & 23.32$^\star$ \\
        &Projected GAN~\cite{NEURIPS2021_9219adc5} & 3.39 & - & 1.59 
            & 143.64 & 105.84\\
        &INSGen~\cite{NEURIPS2021_4e0d67e5} & 3.31 & - & - 
            & 89.00 & 24.94 \\
        &StyleGAN-XL~\cite{sauer2022stylegan} & 2.19 & 2.02 & -
            & 14.29 & 67.93\\
    \hline
    {\multirow{7}{*}{Attention}}
        &GANFormer~\cite{hudson2021ganformer} & 7.42 & - & - 
            & - & 32.48\\
        &GANFormer2~\cite{hudson2021ganformer2} & 7.77 & - & -
            & - & -\\
        &HiT-S~\cite{NEURIPS2021_98dce83d} & 3.06 & - &  - 
            & 86.64$^\dagger$ & 38.01$^\dagger$ \\
        &HiT-B~\cite{NEURIPS2021_98dce83d} & 2.95 & - & - 
            & 52.09$^\dagger$ & 46.22$^\dagger$ \\
        &HiT-L~\cite{NEURIPS2021_98dce83d} & 2.58 & 6.37 & -
            & 20.67$^\dagger$ & 97.46$^\dagger$ \\
        &StyleSwin~\cite{Zhang_2022_CVPR} & 2.81 & 5.07 & 2.95
            & 62.48 & 48.93 \\
        &StyleNAT (\textbf{ours}) & 2.05 & 4.07 & 3.40 
            & 59.90 & 48.92\\
    \hline
    {\multirow{7}{*}{Diffusion}}
        &DDPM~\cite{NEURIPS2020_4c5bcfec} & - & - & 7.89
                & - & 256.00\\
        &D.StyleGAN2~\cite{wang2023diffusiongan} & - & 2.83 & 3.17
            & - & \phantom{0}23.94\\\
        &D.Proj.Gan~\cite{wang2023diffusiongan} & - & - & 1.85
            & - & 105.85\\
        &LDM~\cite{Rombach_2022_CVPR} & 4.98 & - & 4.02
            & 1.28 & 329.32\\
        &LFM~\cite{dao2023flow} & 4.55 & - & 5.54
            & 4.18 & 457.06\\
        &UDM~\cite{pmlr-v162-kim22i} & 5.54 & - & - 
            & - & \phantom{0}65.58\\
        &Unleashing~\cite{bond2022unleashing} & 6.11 & - & 4.07 
            & 6.65 & 159.96\\
    \bottomrule
\end{tabular}

    }
    \caption[Comparison of Generative Models]{%
    FID50k results. Usage Metrics are evaluated at $256\times256$
    resolution for fair comparison and were collected ourselves. StyleNAT does
    not utilize any FID enhancing processing, such as StyleGAN's truncation
    trick. $^\dagger$HiT-L was optimized for TPU and there is no existing
    PyTorch version to compare. There is no public checkpoints 
    for Hit-GAN~\cite{NEURIPS2021_98dce83d} and we use their reported V100
    values. While most architectures are built off of the official StyleGAN 
    models, they may not all be able to utilize the custom CUDA kernels, which 
    can significantly increase throughput~\cite{Karras_2020_CVPR}.
    We use no truncation or tempering for StyleNAT.
    }\label{table:exp-maintable}
\end{table}

\begin{figure}[htpb]
    \centering
    \begin{subfigure}[b]{0.49\linewidth}
        \includegraphics[width=\linewidth]{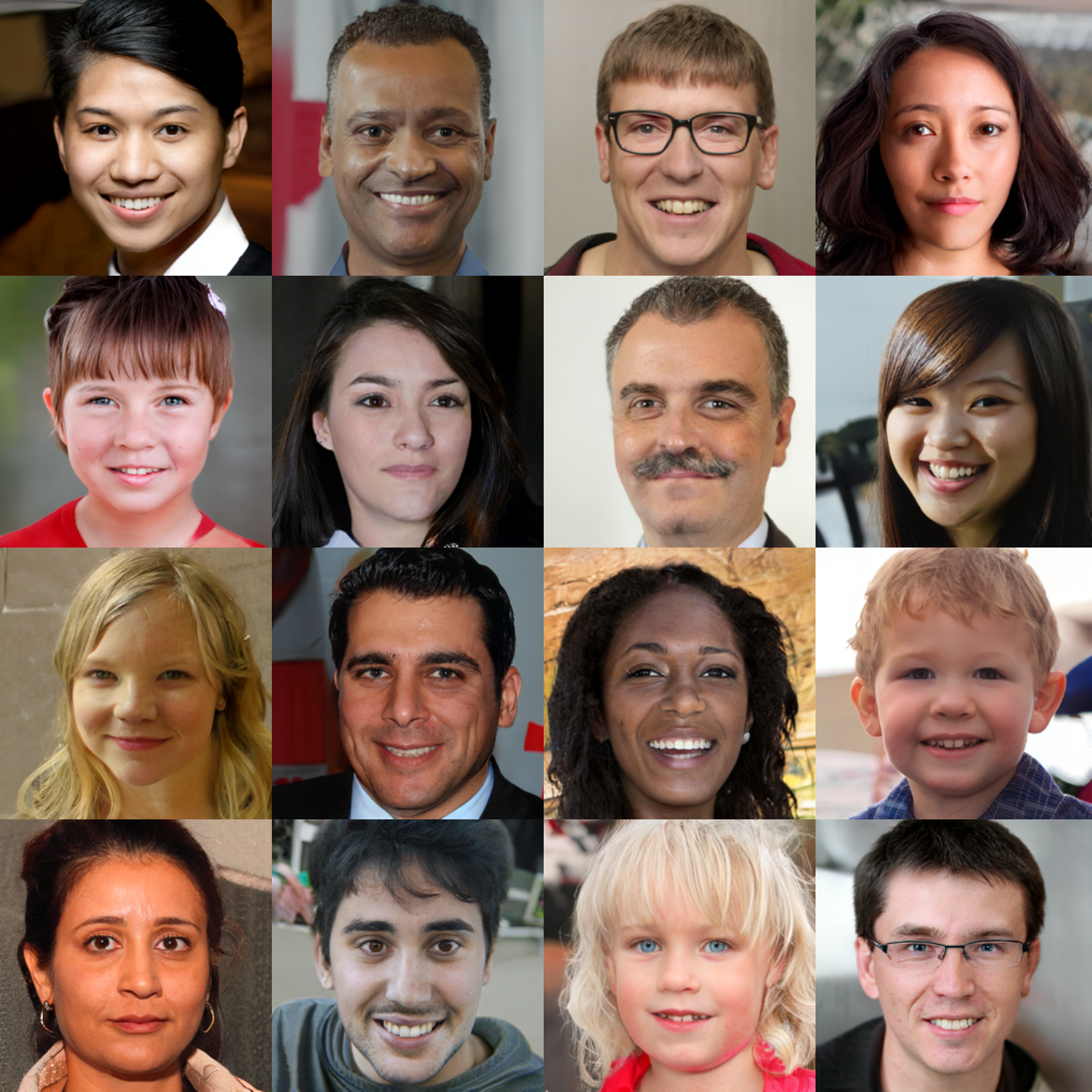}
        \caption{FFHQ-256 Samples}\label{stylenat-ffhqsamps}
    \end{subfigure}
    \begin{subfigure}[b]{0.49\linewidth}
        \includegraphics[width=\linewidth]{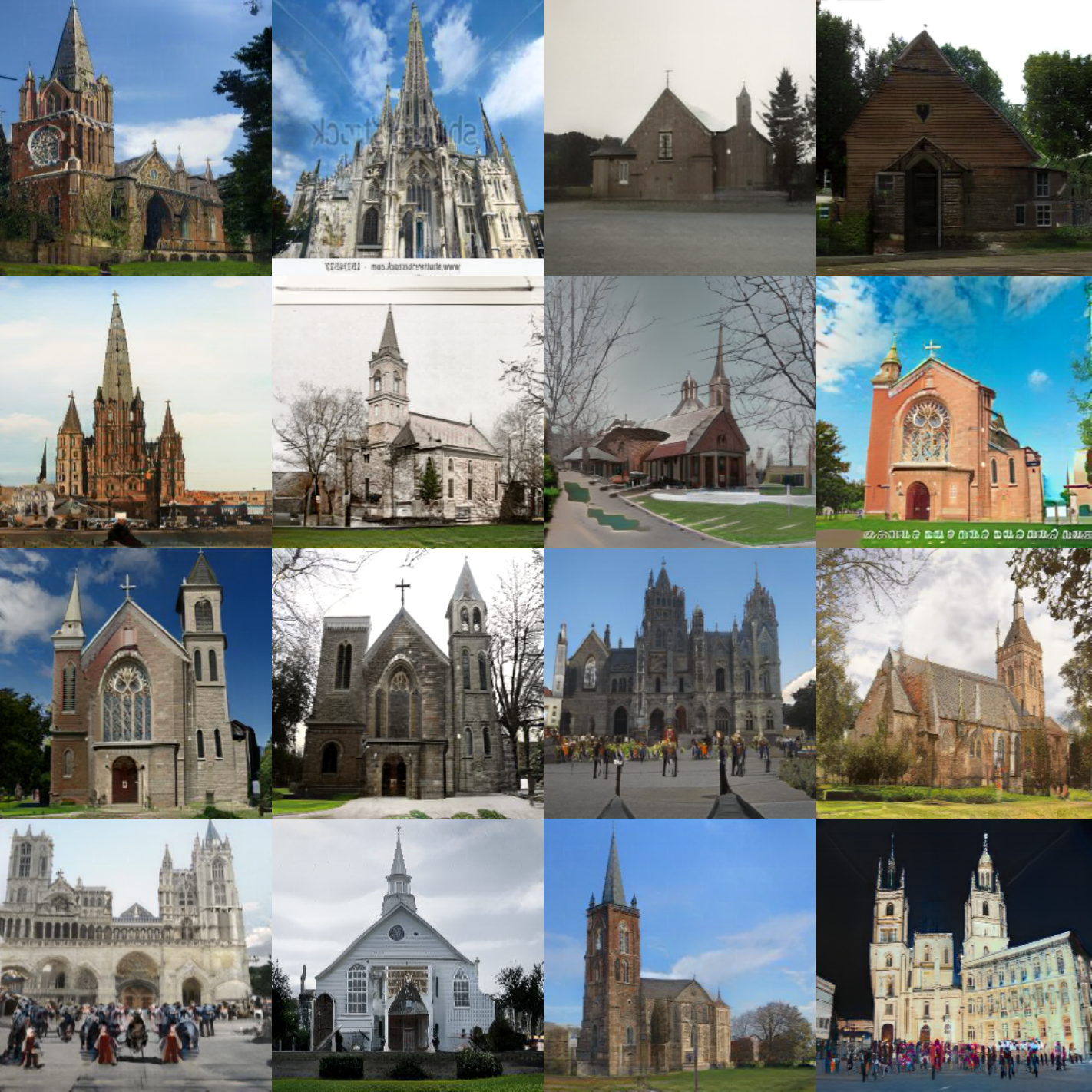}
        \caption{LSUN Church Samples}\label{stylenat-churchsamps}
    \end{subfigure}
    \caption[StyleNAT Samples: FFHQ \& LSUN Church]{%
    Samples generated by StyleNAT. We do not use truncation, softmax
    tempering, nor any other such enhancement techniques.
    }\label{fig:snat-samples}
\end{figure}

\section{When Faced With Sparse Attention}\label{ch:stylenat-ffhq}
To gather a baseline value we first replace the Swin layers in StyleSwin with an
unmodified Neighborhood Attention Transformer, focusing on the FFHQ dataset. 
This modification results in a minor improvement of $0.07$ FID.
Following this, we incorporate Hydra-NA, using a kernel size of $7\times7$ for
all attention heads, but set half the heads to have a dilation increasing by a
power of 2 ($d=2^N$), maximally for the resolution at a given level, 
where $N = \left\lfloor \log_2\left(\frac{\mathcal{R}}{k}\right)\right\rfloor$.\footnote{At a $256\times256$ image resolution
and a kernel of size 7 this gives us a dilation of size 32, making a highly
sparse receptive field across 224 pixels.}
This method allows for dense local receptive fields as well as highly sparse
global receptive fields to intermix through the attention mechanism.

This improved the performance by an additional $0.5$, strongly suggesting that
this method is better able to learn the data generating function.
Notably, this result is only outperformed by
StyleGAN-XL~\cite{sauer2022stylegan}, which is ${\approx}40\%$ larger and $24\%$ 
the throughput, and a variant StyleSAN-XL~\cite{takida2024san} that introduces
a novel training objective.
At the time of our work, StyleGAN-XL was the state of the art network on
FFHQ-256 and this result caused StyleNAT to push the Pareto Frontier in both 
FID vs model size as well as FID vs throughput.
We noticed that StyleSwin had utilized random horizontal flips when training on
LSUN Church, but was not used on FFHQ and
decided to perform this training as NA had shown to be quasi-equivariant to
translations and rotations~\cite{Hassani_2023_CVPR}, and given transformers'
preference for augmentation, that this would improve the score while
demonstrating better generalization capabilities.
This model trained for $10^6$ iterations, beginning the LR-Decay at 740k
iterations, and achieved out best result at 940k iterations (60.2M images). 
Notably the model was continuing along a decreasing trajectory, as show in
\Cref{fig:ffhq-fiditer}.
Since our result had surpassed the state of the art at the time, StyleGAN-XL, we
chose to move on, considering computational restraints. 
Our goal is not to achieve state of the art performance, but rather to
demonstrate the integration of local and global structures within data.
\begin{figure}
    \begin{subfigure}{0.49\textwidth}
        \includegraphics[width=\textwidth]{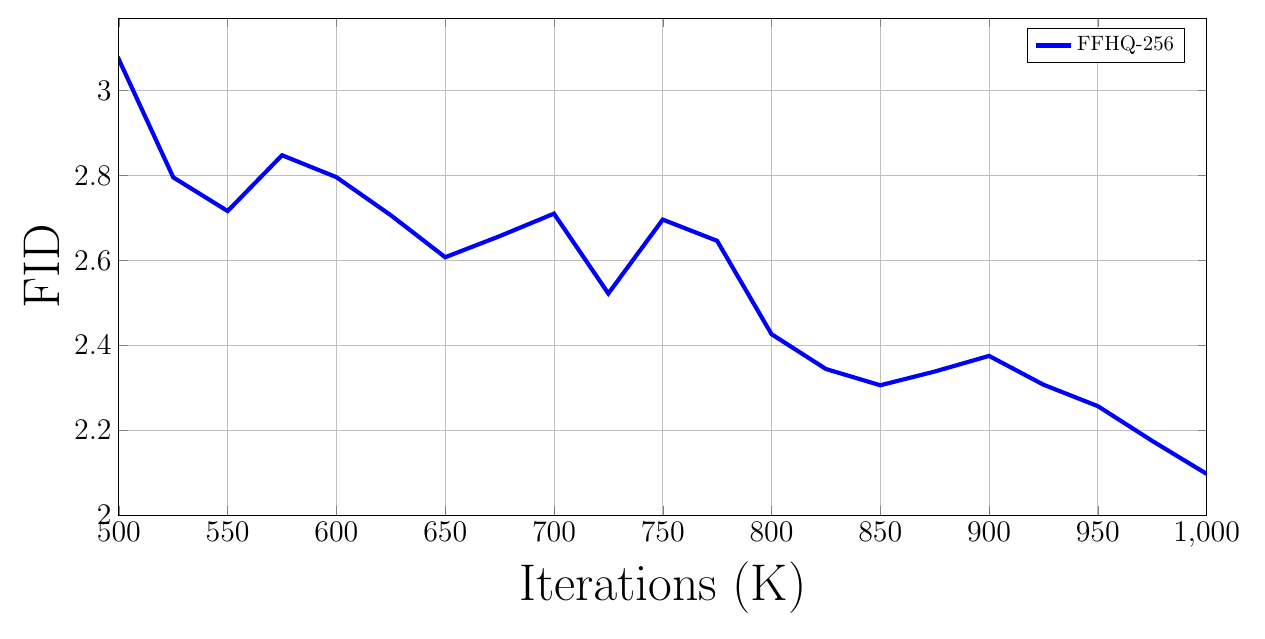}
        \caption{%
            FFHQ-256
        }\label{fig:ffhq256iters}
    \end{subfigure}
    \begin{subfigure}{0.49\textwidth}
        \includegraphics[width=\textwidth]{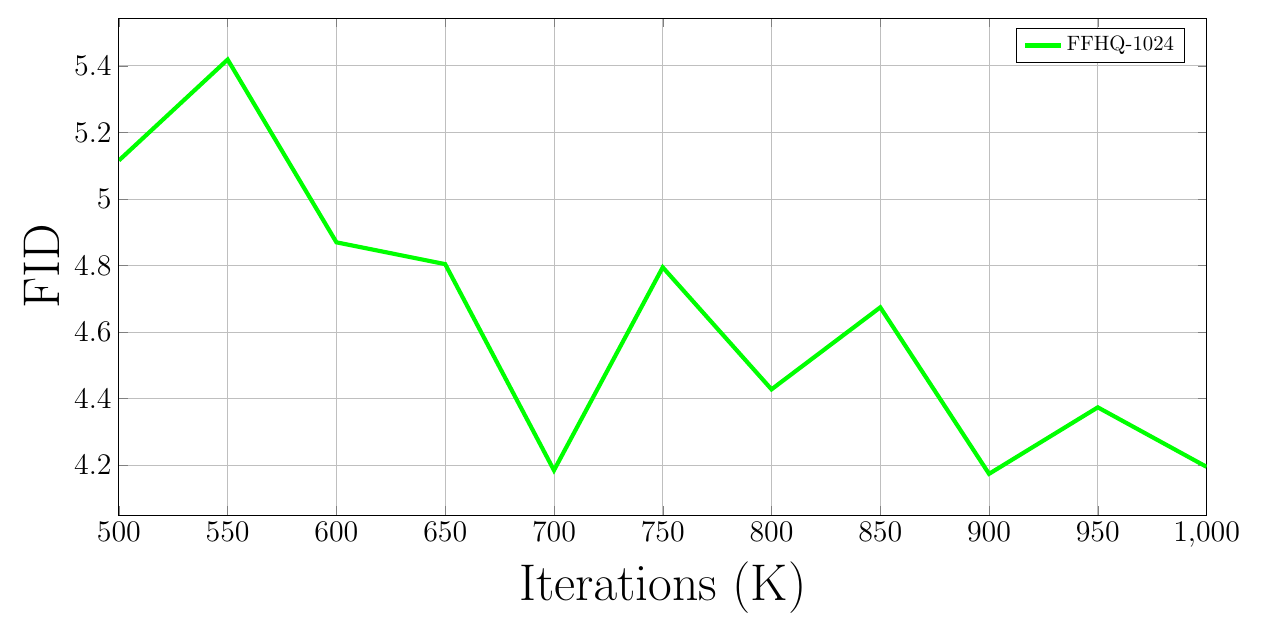}
        \caption{%
            FFHQ-1024
        }\label{fig:ffhq1024iters}
    \end{subfigure}
    \caption[StyleNAT FID vs Iteration]{%
        FFHQ training: FID vs Iteration (in thousands). We see that the FID
        performance has not converged. This suggests the models are not
        optimally trained.
    }\label{fig:ffhq-fiditer}
\end{figure}

\begin{table}[hbpt]
    \begin{subtable}{0.47\linewidth}
    \centering
    \begin{tabular}{l|c|r}
    \toprule
    \multicolumn{3}{c}{\textbf{FFHQ Ablation}}\\
    \hline
        \textbf{Ablation} & \textbf{FID} $\downarrow$ & $\mathbf{\Delta}~\downarrow$\\
    \hline
    StyleSwin & 2.81 & --\phantom{00} \\
    + NA & 2.74 &  \gain{-0.07}\\
    + \textbf{Hydra-NA} & 2.24 & \textbf{\gain{-0.50}}\\
    + Flips & \textbf{2.05} & \gain{-0.19}\\
    + Prog Di (4) & 2.55 & \loss{+0.50}\\
    \bottomrule
    \end{tabular}
    \caption{Ablation study comparing models on FFHQ-256 dataset. Starting with
    StyleSwin~\cite{Zhang_2022_CVPR} we first add unmodified Neighborhood
    Attention (NA)~\cite{Hassani_2023_CVPR}, then Hydra-NA, horrizontal flipping
    data augmentation, and progressive dilations.%
    }\label{tab:ffhq256-ablation}
\end{subtable}
    \hfill%
    \begin{subtable}{0.51\linewidth}
    \centering
    \begin{tabular}{c|c|c|r}
    \toprule
    \multicolumn{4}{c}{\textbf{Church Ablation}}\\
    \hline
    \textbf{Splits} & \textbf{Heads} & \textbf{FID} $\downarrow$ &
        $\mathbf{\Delta}~\downarrow$\\
    \hline
    2 & 4 & 23.33 & --\phantom{00} \\
    4 & 4 & 6.08 &  \gain{-17.25}\\
    6 & 8 & 5.50 & \gain{-0.58}\\
    8 & 8 & 3.40 & \gain{-2.10} \\
    \bottomrule
    \end{tabular}

    \caption{Comparison for number of head partitions (splits) when learning 
    LSUN Church. Min heads represents the minimum number of heads in our
    transformer. Early layers begin with 16 heads and halve until minimum
    beginning at $128\times128$ resolution.
    }\label{tab:exp-church-heads} 
\end{subtable}%

    \caption[StyleNAT Ablations]{Ablation studies of StyleNAT architecture, studying different
    configurations. Results for FFHQ-256 and LSUN Church, respectively.%
    }\label{tab:ablations}
\end{table}

Subsequently, we tried a few other dilation patterns but did not see significant
changes.
Additionally, we attempted further partitioning of the attention heads,
including two intermediate dilations, but observed a decline in performance and
frequent model collapse.
We believe this is due to only having 4 attention heads at resolutions
$\geq256\times256$ and 8 attention heads at $128\times128$, requiring undue
burden for each head.
We believe that increasing the number of heads and head embedding dimension may
lead to increased performance.
The results of these ablation studies can be found in
\Cref{tab:ffhq256-ablation}

To test the scalability of this work we also perform \emph{a single} training
for high resolution, at $1024\times1024$.
Identical training procedures were utilized, but this time we started our
LR-Decay at 500k iterations and stopped training at 900k iterations, achieving
an FID of 4.17.
We did not perform any parameter search at this scale due to the costly
computational budget but believe this demonstration demonstrates scalability as
the result significantly outperforms all other transformer based architectures.
While we did not surpass the FID of StyleGAN3 our model is able to produce
images of higher visual fidelity and does not contain many of the visual
artifacts that StyleGAN3 creates.
Further discussion is provided in \Cref{ch:stylenat-amaps}.

\section{A Bump While Headed To Church}
We also train our model using the LSUN Church dataset, which includes images of
cathedrals, churches, temples, and towers.
This dataset presents significantly different challenges, images containing both
biological and non-biological features and with much more complex scenes.
While FFHQ has images center cropped around human faces and minimize
backgrounds, this dataset has diverse foregrounds and backgrounds (usually the
sky).
This creates strong dependence on localized features and lower dependence on
global ones, as many long range features may be determined entirely through
local ones (e.g. the sky).
The highly asymmetric nature of the images also reduces these global
dependencies, with features such as windows frequently appearing in different
sizes and shapes.
This frequently results in generators having significant performance gaps
between FFHQ and LSUN Church.
While StyleGAN-XL demonstrated that Style-based generators could scale with data
diversity, this requires significant architectural changes and additional
parameters.
Despite these challenges, this dataset can help to better understand the biases 
of our architectural changes and how well it can adapt to more complex 
environments.

We initially follow identical training procedures and architecture, splitting
heads between dense local windows and sparse global windows.
We observe that this model quickly diverges, resulting in mode collapse.
We then increase the number of partitions, following the architecture that
diverged in FFHQ.
This variant dramatically improves FID and stability, showing the dataset's
stronger dependence on localization.
We further increase the partitions to 6 and change our minimum head count, which
only affects the final layer, to 8.
This necessitates a decrease in the head dimension and results in a more modest 
increase in FID.
A final configuration is attempted increasing the number of partitions to 8,
assigning 2 heads to each partition in layers operating on resolutions below 128
and 1 attention head for those larger.
This results in a larger FID gain, and while the result is not as impressive as
those in FFHQ the result is highly competitive.
The results of this ablation can be found in~\Cref{tab:ablations}.

\section{Metrics Are Not Enough}\label{ch:stylenat-amaps}
While the FID results in \Cref{table:exp-maintable} show substantial effects, it
is important to recognize the biases and limitations of the evaluation metrics
(\Cref{sec:bg-measures}).
\emph{The main issue is that most of these metrics were developed when the
quality of generation was substantially lower.}
The authors of the metrics were not deceived by the correlations they found, but
the rapid success of generative research forces us to face their limitations.
They still provide utility but we must be careful to not become overly reliant
upon them as they are not perfectly aligned with the things we wish to measure.

FID uses a Fr\'{e}chet Divergence, which measures the difference between two
Gaussians, $G_0,G_1$

\begin{equation}
    d(G_0(\mu_0,\Sigma_0), G_1(\mu_1,\Sigma_1)) = \sqrt{||\Delta\mu||_2^2 +
    Tr\left(\Delta\Sigma - 2\sqrt{(\Sigma_0\Sigma_1)}\right)}
\end{equation}

Where $\mu,\Sigma$ is the mean and covariance, respectively, and $Tr$ is the
trace of the matrix.
The \emph{Inception} part of this is metric refers to the fact that the
Gaussians are drawn from the final pooling layer of a Inception-V3
Network~\cite{Szegedy_2016_CVPR} that has been trained on
ImageNet~\cite{deng2009imagenet}.
While the performance was sufficient at the time, the accuracy is sub-par by
today's standards.
Other work has demonstrated that FID can create
distortions~\cite{kynkaanniemi2023the} or there can be flaws in evaluation in
subtle effects like through the image downsampling method 
used~\cite{Parmar_2022_CVPR}.
These subtle effects can make evaluation difficult, vary dramatically between
libraries and even library versions.
Simply updating the Inception Network to a different model can provide
improvements, as shown by Kynk\"{a}\"{a}nniemi~\etal\cite{kynkaanniemi2023the},
this does not resolve the underlying problem.

After the pre-print of this paper was released
Stein~\etal\cite{stein2023exposing} performed a large study to determine which
metrics strongly correlated with human preference.
The work involved the largest human preference study to date and used
StyleNAT in their analysis due to its state of the art performance on FFHQ-256
at the time.\footnote{We have no affiliation with Stein~\etal{} nor have had any
communication.}
Their work sought to better understand the biases of many different image
evaluation metrics.
To determine this, they crafted a rather straight-forward experiment, measuring
participants ability to determine if a given image was genuine or a deep fake.
This metric serves as a proxy to determining if images are 
\emph{photo-realistic} or not.
Participants were paid for their, being given bonuses based on their accuracy,
and at a minimum had a Bachelor's level education.
Their results found that there was not a strong relationship between metrics
such as FID and participants ability to distinguish deep fakes from real
imagery.

The result of this forces us to carefully analyze our images and investigate our
network to better determine if our architectural changes actually caused the
improvements we sought. 
To do this we perform two forms of visual analysis to better understand our
network is doing.
The first, \Cref{ch:stylenat-amaps-visual}, compares visual fidelity of images
from StyleGAN3, StyleSwin, and StyleNAT on FFHQ-1024 images.
The second, \Cref{ch:stylenat-amaps-attn}, dives deeper into the attention maps
and what our models are actually attending to.
These visual analyses are limited, but can give us much deeper clues as to what
is happening within these networks.

\subsection{The Face Says It All}\label{ch:stylenat-amaps-visual}
Given our metric limitations we visually inspect samples from images of our
networks.
On FFHQ-1024, we compare StyleGAN3, StyleSwin, and StyleNAT which have FIDs of
2.79, 5.07, and 4.17, respectively.
We use our high resolution 1024 images because this allows us to better inspect
subtle features.
Local features can be more easily seen without the need to zoom and global
features have more difficulties being generated.
We use FFHQ because of our biological aptitudes at recognizing faces.
The human brain was designed to recognize faces, making us apt at identifying
subtleties.

\subsection{Quick Training on Deep Fake Detection}
Readers who are untrained or inexperienced in detecting deep fakes may wish to 
pay special attention to some key areas.
Often visual artifacts can be quickly identified by looking at ears, eyes, neck,
and hair, often in that order.
FFHQ has biases where faces will look much more in the style of profile photos:
that is, they are facing towards the camera.
In general, this results in a full face being visible, ears and all.

Ears make for quick detection due to their large distance within the image and
natural tendencies for faces to be highly symmetric (across the vertical
axis)\footnote{Faces are not perfectly symmetric, but in general, they are far 
more symmetric than asymmetric}.
The localization bias is thus used to our advantage.
Ears are not typically focused on by a typical viewer, so may be easily missed.
Both ears also may not appear in samples, as may not always be relied upon.

Eyes are said to be the window to the soul, and are surprisingly complex.
Issues can often be easy to detect but cultural biases may cause in how natural
this detection is.\footnote{e.g. some ethnic groups have high variance in iris 
colors while others don't. This plays a role in cultural attention to eye color.}
These are great features due to this complexity, their long range, and high
symmetry.
In particular, pupils (the black center of the eye) may not appear round in
generated images.
Difficulties in capturing long range symmetry result in high rates of medical
conditions such as anisocoria (unequal pupil size), dyscoria (misshapen pupils),
ectropion uveae (displacement). or other such effects.
Irises, the colored part of an eye, can also exhibit features like heterochromia
(differing eye colors or differing color in the same eye), aniridia (absence of
iris), or others.
Additionally, eyes often contain reflections, which can quickly give away the
synthetic nature.
Detailed reflections make this easy to spot, but the bright spot of a light 
source will often be non-physical.

Necks can provide more subtle clues.
There is higher variance in necks within these images, where some photos taken
with body facing the camera and the neck will be straight while others will have
their body slightly turned with their head facing the camera.
This can cause issues if we pay attention to the depth in an image and
especially around the chin.
In addition to this, mouth and teeth can exhibit these phasing artifacts, as
shown in the StyleGAN2 work~\cite{Karras_2020_CVPR}.
High variances result in our detectors being worse at these features so they
often slip through.
Hair also provides substantially high variance, but may require more detailed
attention.
These may be simple issues like rapidly changing texture and color or one may 
need to carefully follow some strands of hair.

Finally, accessories like jewelry, hats, glasses, and so on make for quick
identification.
These have lower sampling rates within the data and high variance, so are far
less likely to be coherent.
Ablations and subtle artifacts can appear, which are more difficult for the
detector to catch.

With this in mind we encourage the reader to carefully inspect our images.
We embed our images at high resolutions to make it possible to zoom in for
careful inspection.

It is also important to highlight the biases in the FFHQ dataset.
Many mistakes that these models make can be much better understood by
understanding the data they are trained on.
Some works have found that there are disproportionate representation of certain 
demographics~\cite{Leyva_2024_CVPR}.
In particular, there are higher rates of women than men, in particular of 
Caucasian and Latin descent.
The images are also center cropped, primarily containing a single individual,
and usually facing the camera.
Often people are smiling in these photos, eyes are open, looking at the camera,
in a portrait style.
There are still high variances within the dataset and subjects commonly may be 
wearing glasses, hats, have artistic face painting or cultural face painting 
(e.g. Bindi or Ashes), hands on their face, microphones, wearing costumes, and a
wide variety of situations exist.
In order to perform a serious evaluation generative researchers are strongly
encouraged to manually inspect the dataset so that they can better understand
what they are modeling.
Without manual inspection researchers will most certainly make false assumptions
about this data.%
\footnote{Samples may be available online, such as:
\url{https://huggingface.co/datasets/pravsels/FFHQ_1024}}

\subsection{Fingerprints}

These results are highly subjective but still can provide substantial value.
To ensure that we are not completely unfair in our comparisons we try to present
the best samples from these generators.
We wish to error in the direction of a steelman rather than a strawman.

Our goal is not to determine which image generator is better, but find patterns
in the unique flaws.
These flaws provide clues into how our networks interpret the data and can
provide hints at how to improve our generators.
Understanding these systematic flaws is critical to understanding what future
architectural changes need to be made.

\subsubsection{StyleGAN}

For StyleGAN3 we carefully searched through the public set of curated
images which is linked on their 
\href{https://github.com/NVlabs/stylegan3}{GitHub Repository} under the
directory 
\href{https://nvlabs-fi-cdn.nvidia.com/stylegan3/images/stylegan3-r-ffhq-1024x1024/}{StyleGAN3-r-ffhq-1024x1024}.
Being curated this is already biased towards higher quality samples.
We then manually search through this for what we believed was the best sample.
For StyleSwin and StyleNAT we instead generate 50 samples, discard any with
obvious artifacts (colloquially referred to as ``GAN Monsters'') and select the
best example.
This potentially creates a bias towards StyleGAN3, given the additional level of
curation.

\begin{wrapfigure}[20]{l}{0.55\textwidth}
    \vspace{-1.5em}
    \centering
    \begin{subfigure}{\linewidth}
        \input{Includes/Figures/StyleNAT/StyleGAN/artifacts.tex}
    \end{subfigure}
    \caption[StyleGAN3 Visual Artifacts]{%
    Visual artifacts from StyleGAN3 FFHQ-1024 samples (using image 0068).
    Sample highlights banding effects, hexagonal patterns, and other artifacts
    common to this generator.
    }\label{fig:stylenat-visual-sgan}
\end{wrapfigure}

\setlength{\parindent}{0em}
Within the StyleGAN images we notice a string of beads like artifacts. 
\setlength{\parindent}{\oldparindent}
These structures may be difficult to notice at first glance but become difficult
to ignore after noticed.
These appear most prominently between the two ``wrinkles'' in the middle of the
forehead of the sample image.
We noticed that such patterns appear throughout the face and were quite visible
in all images we looked at.
We do not know the cause of these patterns but found that they were noticeable 
in other datasets, including AFHQ, which contains images of animals.
This suggests this is a fingerprint of the architectural design rather than of
the dataset, potentially being a more advanced droplet artifacts discussed in
StyleGAN2~\cite{Karras_2020_CVPR}.
Those patterns were often masked by an animal's fur and more easily detected
when looking at noses or tongues.
\footnote{\href{https://nvlabs-fi-cdn.nvidia.com/stylegan3/images/stylegan3-t-afhqv2-512x512/0175.png}{stylegan3-t-afhqv2-512x512/0175.png}}
\footnote{\href{https://nvlabs-fi-cdn.nvidia.com/stylegan3/images/stylegan3-t-afhqv2-512x512/0138.png}{stylegan3-t-afhqv2-512x512/0138.png}}

In addition to this we noticed extremely high rates of geometric artifacts in
glasses.
Most visible around the edges of the glasses, but careful inspection will show
that these appear throughout.
These may be due to difficulties in capturing reflections.
The glasses also are non-physical, with the temples simply vanishing.
Another strong band can be found where the temples should be, and indicate that
these are statistical artifacts (like the droplets), fooling the detector into
thinking the temples exist.
Additionally, there is some non-physicality to the nose pads.
More inspection can reveal many other artifacts, including around the mouth,
melding teeth, hair, fused neck, and tear duct.
Specifically, the person in the photo appears to be missing a jaw, which
appears surprisingly frequently among the curated samples.

Despite StyleGAN3's high FID score these artifacts are trivially detectable if
one knows what to look for, but may easily be missed if only given a passing
glance.
In particular, StyleGAN3's errors typically highlight larger failures when it
comes to long range coherence.
While still highly symmetric, there are more symmetry errors than one would
expect of an average human.

\subsubsection{StyleSwin}

StyleSwin holds the lowest FID, and unfortunately produced a large number of low
quality samples at this scale.
The authors of the paper noted some ``block'' like artifacts, which can be
clearly seen in Figures 3 and 5 of their work~\cite{Zhang_2022_CVPR}.
We notice similar artifacts in all the samples we generated.
Detection can be a bit difficult depending on a reader's screen, but by zooming
into the forehead concentric rectangles become quite visible.
We believe that these artifacts are due to the Swin Transformer, and provide
further discussion alongside our attention maps.

\begin{wrapfigure}[19]{l}{0.55\textwidth}
    \vspace{-1.6em}
    \centering
    \begin{subfigure}{\linewidth}
        \input{Includes/Figures/StyleNAT/StyleSwin/artifacts.tex}
    \end{subfigure}
    \caption[StyleSwin Visual Artifacts]{%
    Visual artifacts from StyleSwin FFHQ-1024 samples (we generated these).
    Sample highlights rectangular geometric patterns on face, and
    poor texture on ears.
    }\label{fig:stylenat-visual-sswin}
\end{wrapfigure}

\setlength{\parindent}{0em}
We found that StyleSwin's integration of sliding windows (SWA) and shifted
windows (SWSA) does not properly integrate long range features.
\setlength{\parindent}{\oldparindent}
This is most apparent by looking at the eyes in our sample.
All parts of the eye differ in size: both iris and pupil.
The effect is as if the right eye is closer to the camera than the left eye, yet
this depth does not correctly correspond to the direction that the nose, eyes,
and mouth point in.
The eyes exhibit heterochromia, being different shades of blue (right is almost
green), anisocoria, dyscoria, and ectropion uveae.
The reflections within the eyes are also substantially different, as if looking
at completely different scenes.

Additionally, we found common issues with facial textures, easily noticeable in
the ear.
High rates of speckling can be seen by zooming in on the cheek, where some
non-physical banding may also be found.
While this image does not have the fused neck like StyleGAN, it has minor issues
with generating realistic depth and some artificial lines can be seen along the
neck.
Similar depth issues may be visible by looking at the nose, which blends into
the cheek.

\begin{wrapfigure}[19]{l}{0.55\textwidth}
    \vspace{-1.5em}
    \centering
    \begin{subfigure}{\linewidth}
        \input{Includes/Figures/StyleNAT/NA/artifacts.tex}
    \end{subfigure}
    \caption[StyleNAT Visual Artifacts]{%
    Visual artifacts from StyleNAT FFHQ-1024 samples (We generated these).
    Sample highlights minor skin texture issues, some chromatic aberrations, and
    unnatural blue speckling around eyes.
    }\label{fig:stylenat-visual-snat}
\end{wrapfigure}

\setlength{\parindent}{0em}
While this sample has many artifacts and may be more easily identified than the
StyleGAN3 sample, there are some aspects that perform better.
The concentric rectangles are often less noticeable compared to the beading in
StyleGAN, as well as the neck and jawline appear more realistic.

\subsubsection{StyleNAT}
StyleNAT has a FID $\frac{2}{3}$ the distance between StyleGAN3 and StyleSwin,
being much closer to StyleSwin on the metric.
Yet, we noticed that images were consistently much better than StyleSwin, and 
the fidelity was much closer to StyleGAN3.
We believe StyleGAN3 still produces better images at a higher frequency and our
work could likely have greatly benefited from tuning and continued training.
Despite producing many high quality images, our images are still not without
error.
\setlength{\parindent}{\oldparindent}

The eyes are a bit of interest and may demonstrate some aniridia (absence of
iris).
Careful inspection makes this unclear, as there is some brown and even a bit of
blue-gray.
Interestingly the eyes look quite similar, with both eyes following the same
pattern.
We are unable to differentiate if this is sectoral heterochromia (partitioned)
or central heterochromia (radial).
The colors being a brown and dark blue, especially on a male, make this much 
more natural but it should not be assumed the model learned that causal 
relationship.%
\footnote{Irises do not have blue or green pigments. Instead these 
colors are created through eye structure. Both are brown eyes with lower 
amounts of melanin.}%
\footnote{Green and Blue eyes are sexually dimorphic. Blue is more common among
men, green is more common among women.}

Like the other images, our model still struggles with the neck.
While the image much better captures depth there are some non-physical features.
Below the jaw the neck bulges and might be mistaken for an Adam's Apple or a
weird camera angle.
There is also some slight banding around the lower part of the neck, half way
between the jawline and shirt collar.
Additionally, the jaw bulges, as if merging a forward facing face and a slightly
turned face.

Interestingly, we do not observe systematic depth errors like StyleGAN and
StyleSwin.
Yet we do notice our images also create unique skin textures, different than
StyleGAN3 or StyleSwin.
These are most noticeable around the lips and the subject's nose.
Larger lines may be found around the forehead and like StyleSwin, we can trace
these to the attention maps (\Cref{ch:stylenat-amaps-attn}).
Additionally, we find a systematic blue speckling, most easily noticed around the 
eye or the beard of the subject's chin.
This may pattern can be difficult to detect depending on the reader's monitor.

\subsection{Attention To Details}\label{ch:stylenat-amaps-attn}
To better understand the cause of these systematic issues we visualize the
attention maps across our generator.
We modify the standard roll-out attention map to account for the localize
windows and have to undo the shifting for both methods.
This method is open sourced with our project and appears to be the first method
for visualizing attention maps for either Swin or \natten.
For quick reference \Cref{fig:progressiveAttnMaps} illustrates these
intermediate layers for both StyleNAT (\cref{fig:progressiveAttnMaps_na}) and 
StyleSwin (\cref{fig:progressiveAttnMaps_swin}).

When looking at the final layer of StyleNAT (\Cref{fig:attn-snat}) we observe
many of the same patterns that we found during visual inspection.
The matching patterns help validate our interpretation of these attention maps
and our method of extraction.
In the first transformer we are able to observe the same banding lines around
the face.
Using these as reference can aid in their detection if this was previously
unclear.
We also notice that these appear in the dense heads in the second transformer.
Similarly, we are able to observe the speckling, especially around the chin.
In most of these maps we are able to observe a circular shape in the forehead,
which corresponds to a hair curl.
Careful inspection of this curl in \Cref{fig:stylenat-visual-snat} shows that
this is too perfectly circular and may actually me more similar to a statistical
droplet that is better masked.
Given this, similar explanations might apply to the blue speckling.

\begin{figure}[H]
    \centering
    \input{Includes/Figures/StyleNAT/attention_maps_progressive.tex}
    \caption[StyleNAT and StyleSwin Attention Maps]{%
        Visualization of the first and last attention head progressing
        through StyleNAT. We start at a resolution of $16\times16$ and grow to 
        $1024\times1024$. 
        We generate 50 samples from each network and
        choose the best image from the sample to make comparisons as fair as 
        possible. 
        The top row shows the first attention head, with 2
        transformers per resolution level. The bottom row shows the last
        attention head. 
        \cref{fig:progressiveAttnMaps_na} visualizes for
        StyleNAT (\textbf{ours}) and \cref{fig:progressiveAttnMaps_swin}
        follows StyleSwin~\cite{Zhang_2022_CVPR}.
    }\label{fig:progressiveAttnMaps}
\end{figure}

Critical to the verification of our hypothesis, we observe that the attention
maps form two distinct groups, directly corresponding to our partitioning.
This grouping occurs in different transformers and at different resolutions,
directly matching the head partitioning regardless of the total number of heads.
Our dense kernels have much smoother attention maps, suggesting they are
attending locally.
Our sparse global kernels have more patterns and the highlighted regions (such
as ears and background) correspond to the long range patterns we expect.
While the dense maps also highlight some of the backgrounds their boundaries
closely correspond to similar coloring, which would be a local feature rather
than global.
These maps strongly suggest we have achieved our goals, even if our method is
not fully optimized.

\begin{figure}[H]
    \begin{subfigure}[b]{0.49\linewidth}
        \includegraphics[width=0.95\linewidth]{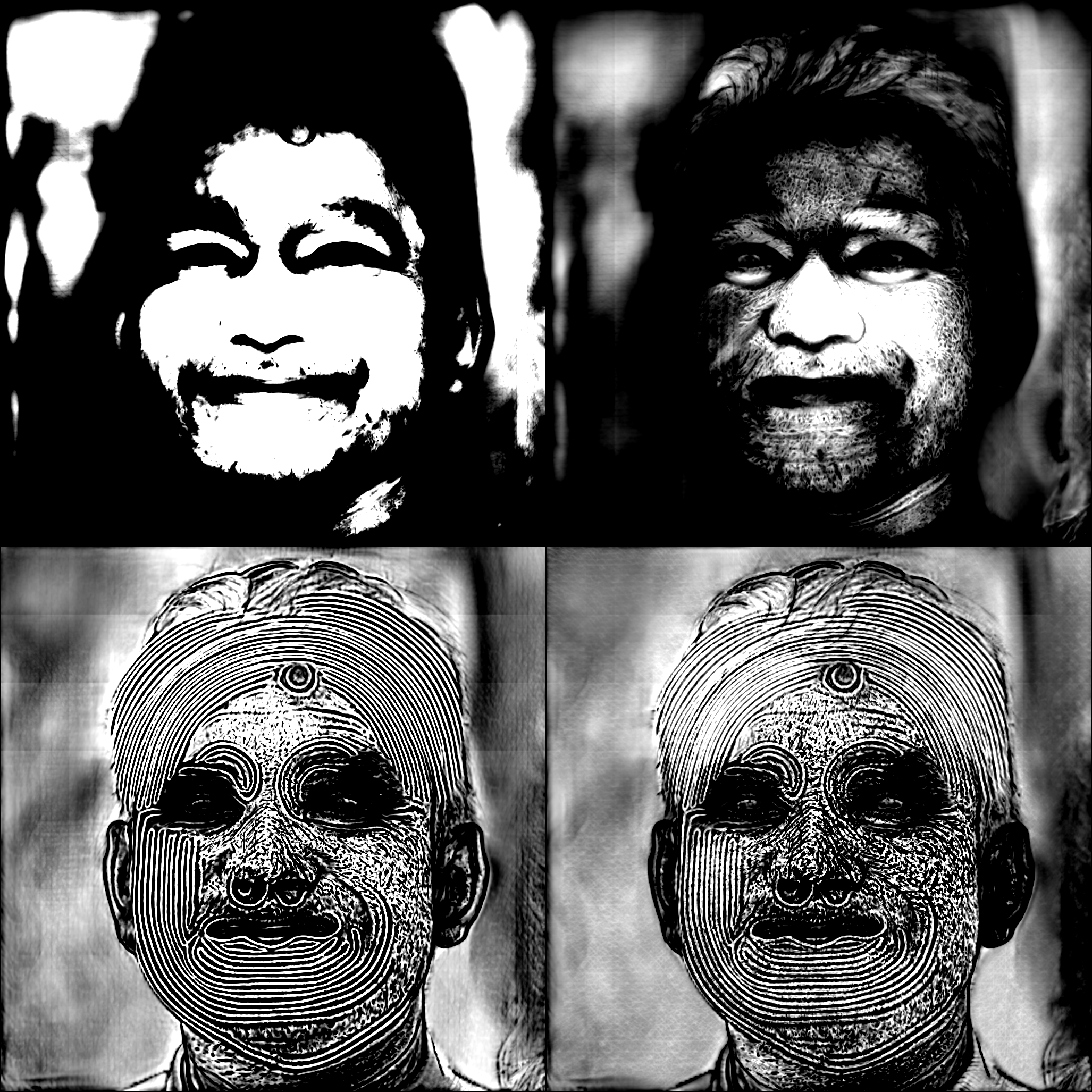}
        \caption{%
        Transformer 0
        }\label{fig:attn-snat-t0}
    \end{subfigure}
    \hfill
    \begin{subfigure}[b]{0.49\linewidth}
        \includegraphics[width=0.95\linewidth]{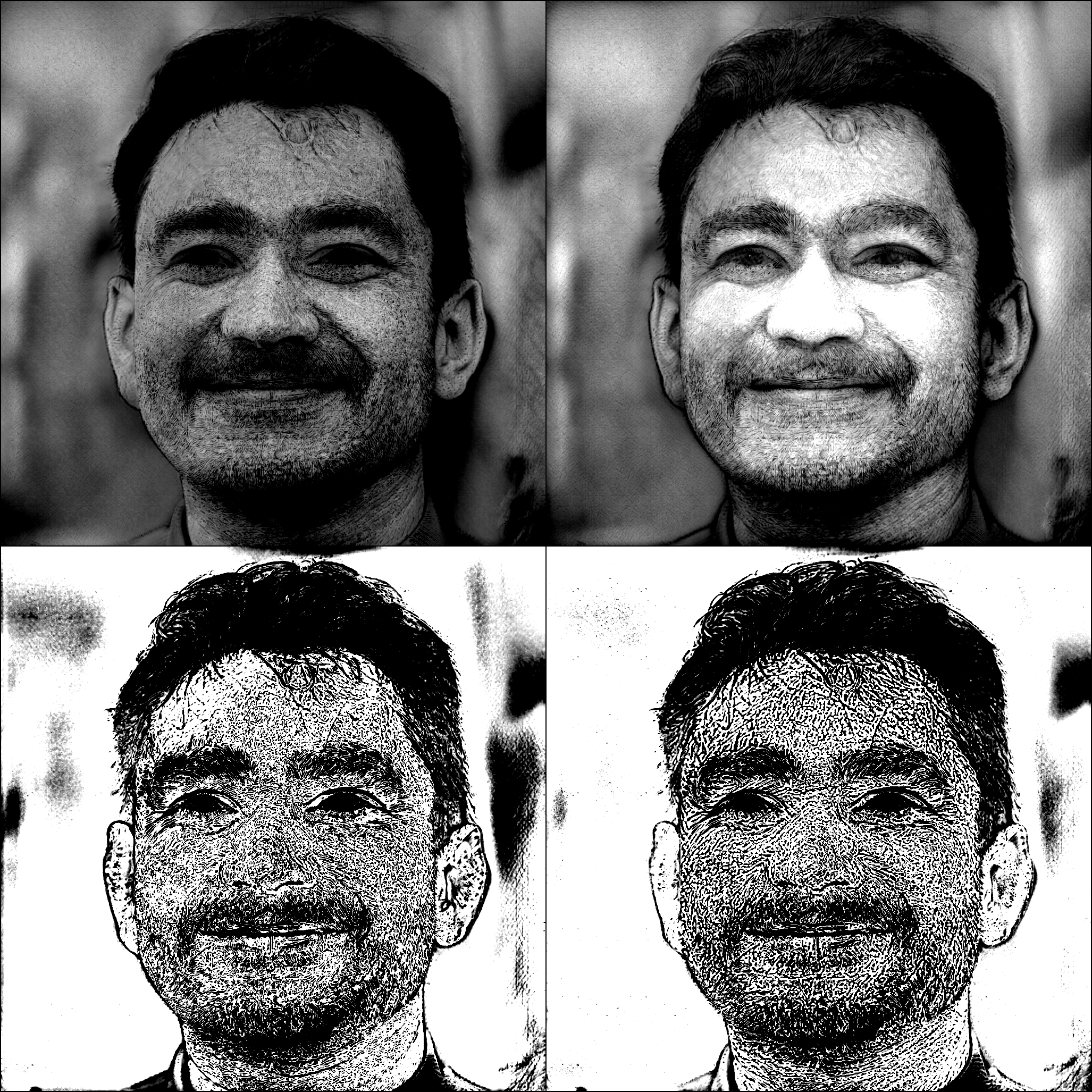}
        \caption{%
        Transformer 1
        }\label{fig:attn-snat-t1}
    \end{subfigure}
    \\
    \begin{subfigure}[b]{0.49\linewidth}
        \includegraphics[width=0.95\linewidth]{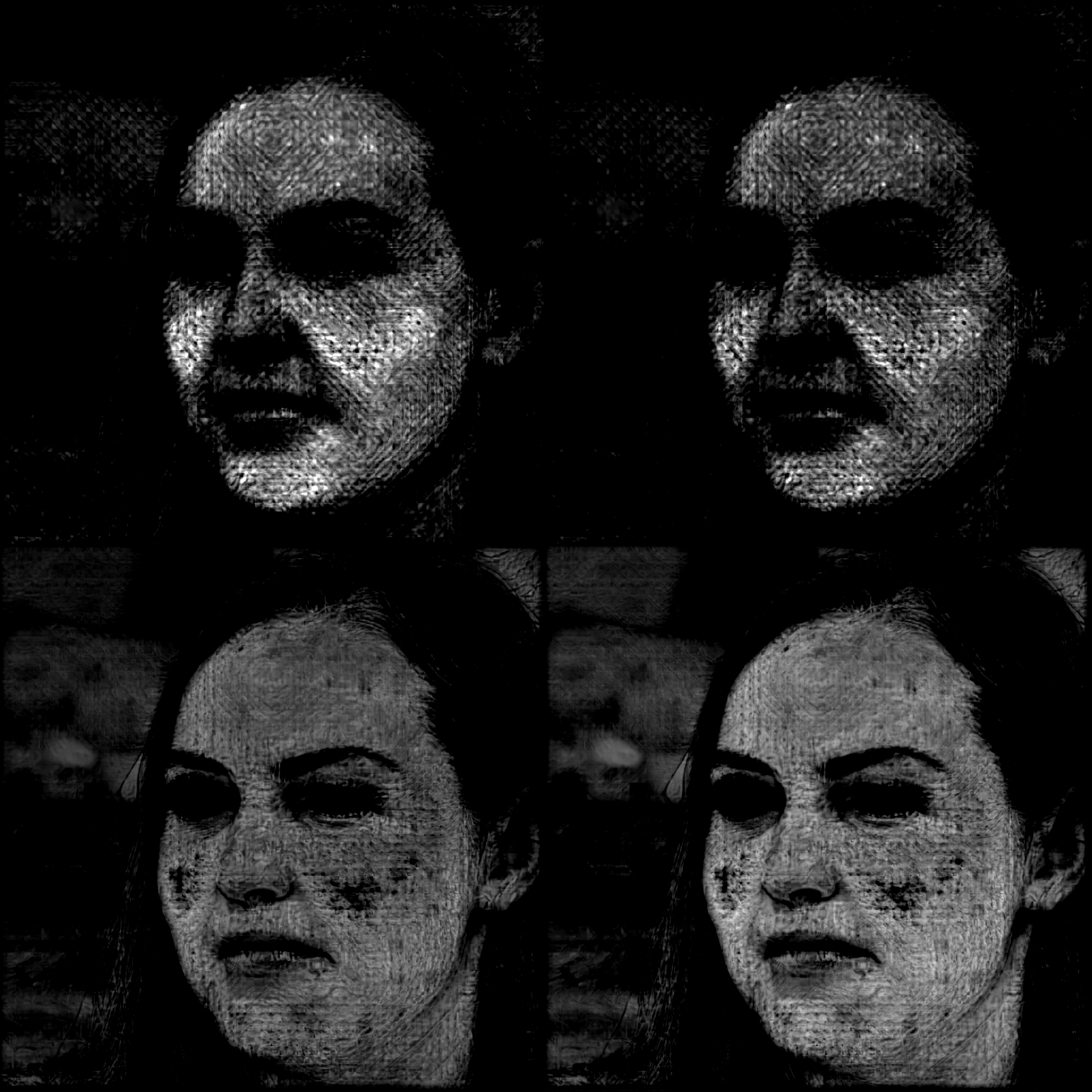}
        \caption{%
        Transformer 0
        }\label{fig:attn-swin-t0}
    \end{subfigure}
    \hfill
    \begin{subfigure}[b]{0.49\linewidth}
        \includegraphics[width=0.95\linewidth]{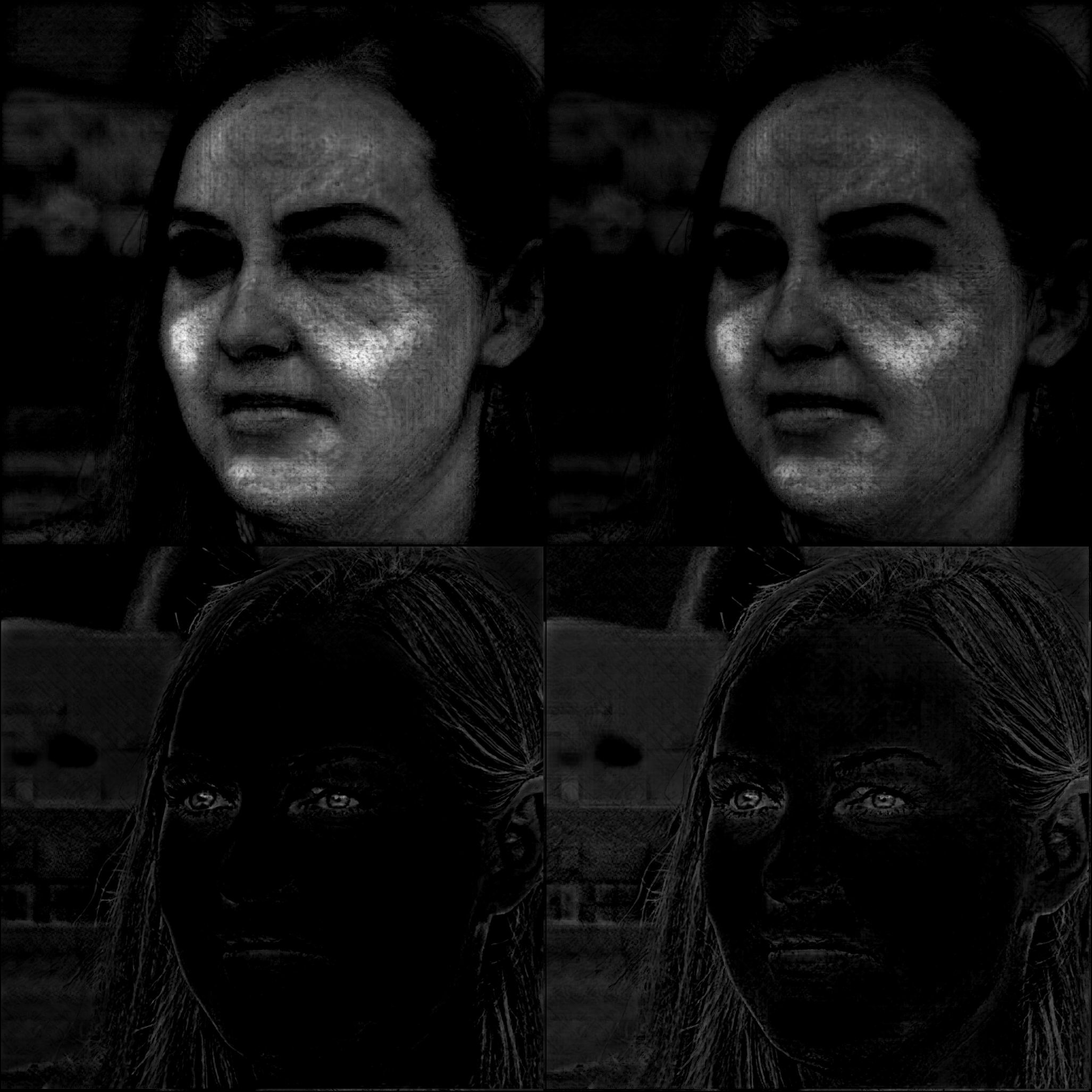}
        \caption{%
        Transformer 1
        }\label{fig:attn-swin-t1}
    \end{subfigure}
    \caption[StyleNAT and StyleSwin Attention Maps]{%
        Visualization of Attention maps 
        (\cref{fig:attn-snat-t0,fig:attn-snat-t1} StyleNAT,
        \cref{fig:attn-swin-t0,fig:attn-swin-t1} StyleSwin)
    for transformers at the 1024 resolution.
    Top row corresponds to localized dense kernels ($k=7,d=1$), second
    row corresponds to the sparse dilated kernels ($k=7,d=128$).
    Banding and blue speckling on images directly corresponds to those in the 
    attention maps.
    We observe divergent attention maps across heads, matching our exception.
    }\label{fig:attn-snat}
\end{figure}

\chapter{Distillation of Invertible Networks}\label{ch:flows}
\chapterquote{
I know numbers are beautiful. If they aren't beautiful, nothing is.
}{Paul Erd\H{o}s}
\textbf{Nota Bene:}
This work is based on the previously published co-authored work 
\emph{Distilling Normalizing Flows}~\cite{WaltonDNF2025CVPR}.
\begin{itemize}
    \item Steven Walton was the primary author of the source code and performed
        the majority of experiments. Steven was also the primary author of the
        paper.
    \item Valeriy Klyukin made significant contributions to the source code and
        to the experiments. Valeriy also provided feedback and contributed to
        the writing of the paper.
    \item Maksim Artemev provided software engineering expertise and feedback
        influential to the design and experiments. Maksim also contributed to
        the writing of the paper.
    \item Denis Derkach provided general support and feedback for the project.
    \item Nikita Orlov provided general support and feedback for the project. He
        also helped provide access to the hardware used for our experiments.
    \item Humphrey Shi was the advisor, contributing overall 
        guidance on the research as well as funding for the work. Humphrey also
        contributed to the writing of the paper and ensuring research stayed on
        track.
\end{itemize}

A frequent task of interest for generative models is the ``\emph{reversibility}
problem''(GAN inversion, etc)~\cite{Bau_2019,Goetschalckx_2019_ICCV}.
That is, determining the map from the image to pre-image
(\Cref{fig:bg-manifold}), or can be seen by mapping the model's representations
back to the data.
This is sometimes referred to as the ``inverse problem,'' but an inverse does
not always uniquely exist, so we avoid such nomenclature unless one does.
Of particular concern are Tractable Density Models
(\Cref{fig:goodfellow_taxonomy}), which allow for a formal, mathematical, 
description of the image's probability density function.
These models are of special interest to many scientists as the formalization
allows for better
interpretability~\cite{Kennamer_Walton_Ihler_2023,chen_residual_2020,fuhr_monotone_1992,hasenclever_variational_2017,hyvarinen_estimation_2005}.
Reversible models increase utility by allowing manipulation of the data
generating process, while invertible models extend this further as manipulation
on the image corresponds to a unique modification in the pre-image (and vice
versa).

\section{Model Distillation}\label{ch:flow-model-distillation}
Unfortunately these models are not as easily trained due to their more
restrictive architectures.
As discussed in \Cref{sec:bg-size}, larger models allow for more smoother
solution spaces, and thus can reduce difficulties in optimization.
Fortunately, there usually exist multiple trajectories that provide a mapping
from the domain to range, and any such mappings are equivalent. 
This encourages the training of large models, but their size makes their usage
cumbersome.
Deployment may be limited, as they may require greater system resources than
available on many systems, as is common with 
LLMs~\cite{brown2020languagemodelsfewshotlearners,devlin2019bert}.
Methods like 
quantization~\cite{gholami2021surveyquantizationmethodsefficient} and reduced 
precision can help reduce the computational burdens but may themselves require 
specialty hardware or instruction sets.
Early works by Bucilu$\breve{a}$~\etal\cite{bucilua2006model} showed that an
ensemble of models~\cite{dietterich2000ensemble} could be compressed 
into a single model.
Further work by Hinton~\etal\cite{hinton2015distilling} showed that smaller
``student'' models could reduce test error by matching the logits of a larger
``teacher'' and more accurate model, effectively \emph{distilling} the 
large model's knowledge.
These ideas expanded, demonstrating the effectiveness of to other
architectures, studying what kinds of information transfers, and how to optimize
such knowledge transfer.
\section{Distilling Normalizing Flows}
While knowledge distillation has been widely studied, these efforts have not
extended to the architectures of Compositional Normalizing Flows.
There only exists limited studies of knowledge distillation for Normalizing
Flows, such as Baranchuk~\etal\cite{baranchuk2021distilling}, the work used a 
conditional normalizing flow for the teacher but removed constraints of 
invertibility and thus the student network is no longer a normalizing flow.
Such works do not take advantage of the unique properties that these
architectures have, which similarly limits their capabilities.
Flow models are naturally invertible, learning compositional diffeomorphisms to
produce their final mappings.
Given a network, $f$, they may be broken down into $k$ sub-networks that are
each diffeomorphic themselves:
\begin{equation}\label{eq:flow-composition}
    f = f_1 \circ \cdots \circ f_k
\end{equation}
The common formulation of these architectures is to use the Change of Variables
formula, where each subnetwork contains the same information but in a different
coordinate system.
For probability distributions, we can specify such a coordinate change as follows:
\begin{equation}\label{eq:cov}
    p_x(\mathbf{x}) = p_u(\mathbf{u})\left|\text{det}\;J_f(\mathbf{u})\right|^{-1}
\end{equation}
where we are mapping from density $p_u(\textbf{u})\mapsto p_x(\textbf{x})$.
Here $\text{det}\;J_f(\textbf{u})$ denotes the absolute value of the determinant 
of the Jacobian of $\textbf{u}$.
Given the compositional nature of these flows we may similarly calculate the
final Jacobian determinant through the product of those in each transform:
\begin{equation}
    \text{det}\;J_{\mathbf{f}}(\mathbf{x}) = 
        \prod_{i=1}^n\text{det}\;J_{\mathbf{f}_i}(\mathbf{x}_i)
\end{equation}
Unfortunately the Jacobian determinant is often computationally expensive, and
much research has been dedicated to finding expressive architectures with more
computationally efficient determinant
\cite{kingma_glow_2018,dinh2017densityestimationusingreal,dolatabadi_invertible_2020,durkan_cubic-spline_2019,NEURIPS2020_d3f06eef,berg_sylvester_2019,grathwohl2018scalable}
calculations.

Due to the unique construction of these models, there are unique opportunities
for transfer between a teacher and student model.
We seek to formalize these relationships and encourage further studying.
We show that there are three main categories in which we may transfer knowledge
between teacher and student and formalize these relationships.

\subsection{Categories of Flow Distillations}
We present these categories of knowledge transfer in a general sense, noting
that arbitrary loss functions, $\mathcal{L}$, may be used between them.
We note that since each layer in a Compositional Flow represents a probability
distribution, that this presents unique conditions that may not be present
within other networks.
Given the comparison between two distributions it is often natural to use a
Kullback-Leibler (KL) divergence.

\begin{align}\label{eq:flow-kldiv}
    \mathcal{L}(\theta) &= D_{KL}\left[p_x(\mathbf{x}) 
        || p_u(\mathbf{x};\mathbf{\theta})\right]\\
        &= \sum p_x\left(\mathbf{x}\right) \log
            \left(
                \frac{p_x(\mathbf{x})}{p_u(\mathbf{x})}
            \right)
\end{align}
Though it is not necessary to make such restrictions and any divergence or
metric may be used for minimization.
One may have different interests in what actually is desired to be minimized,
potentially more interested in probabilistic constraints or geometric.

\subsubsection{Latent Knowledge Distillation}
We define Latent Knowledge Distillation, $\mathcal{L}_{LKD}$, to be the
distillation between the final learned distributions of the teacher and student.
This may be thought of knowledge distillation in the traditional sense, similar
to that of Hinton~\etal.
Specifically, we define this as knowledge transfer when data processing in the
normalizing direction.

\begin{equation}\label{eq:flow-lkd}
    \mathcal{L}_{LKD}(t,s,x) = \mathcal{L}_r(t(x),s(x))
\end{equation}

\subsubsection{Intermediate Latent Knowledge Distillation}
Due to the compositional nature of these flows, there forms more natural
relationships between intermediate layers.
For example, if we wish our student to be half the size of the teacher network
we may view every two flow layers in the teacher as equivalent to a single flow
layer in the student.
In this manner we would compress two teacher layers into a single layer in the
student.
This framing may not be work similarly with other architectures, and may require
significantly more complex maps to be found which have no guarantees of
invertibility.

\begin{equation}\label{eq:flow-ilkd}
    \mathcal{L}_{ILKD}(t,s,x) = \sum_i\mathcal{L}_r(t_i(x),s_i(x))
\end{equation}

Due to the bidirectional nature of these flows, such intermediate knowledge may
be transferred when data processing in either direction.
We believe this form of distillation is deceptively simple but may provide rich
areas of study, especially in the domain of Optimal Transport.

\subsubsection{Synthesized Knowledge Distillation}
The invertible nature of flows allows for symmetry in our models.
While LKD performs knowledge transfer in the normalizing direction, SKD is
performed in the generating direction.
We can view SKD as the inverse of LKD. 

\begin{equation}\label{eq:flow-skd}
    \mathcal{L}_{SKD} (t,s,z) = \mathcal{L}_r(t^{-1}(z), s^{-1}(z))
\end{equation}

\begin{figure}[H]
    \centering
    \includegraphics[width=0.85\textwidth]{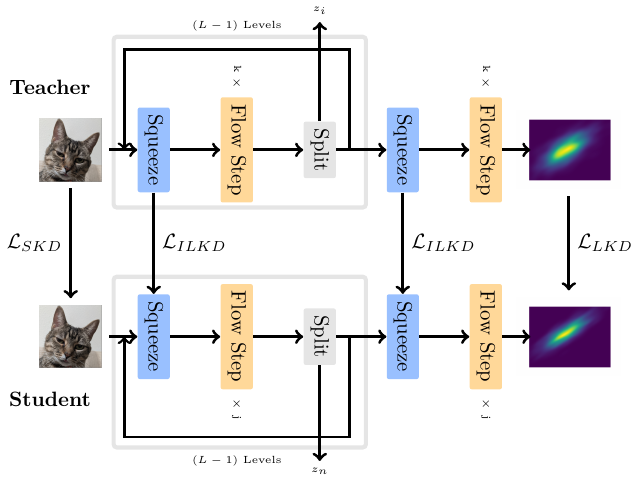}
    \caption[Illustration of Knowledge Transfer for Normalizing Flows]{%
        Illustration of knowledge transfer between two
        Glow~\cite{kingma_glow_2018} based models. $\mathcal{L}_{LKD}$
        represents the Latent Knowledge transfer between the learned
        representations. $\mathcal{L}_{ILKD}$ is the knowledge transfer between
        intermediate representations. $\mathcal{L}_{SKD}$ is the knowledge
        transfer via synthesized data.%
    }\label{fig:distilling-flow}
\end{figure}

Importantly, this form of knowledge transfer need not be performed via
conditional generation.
This means, unlike cycle loss~\cite{CycleGAN2017} or similar styles, we do not
need a ground truth label.
Instead, we can simply ensure that given the same sample from the learned
distribution, the generative outputs are aligned.
In this manner we should treat both the teacher and student as if having the
same learned distribution.

Currently, the generative capabilities of Normalizing Flows makes this form of
distillation more difficult to analyze.
Specifically, image generation has not matched the performance of other
architectures, such as GANs and Diffusion.
Only a few flow models have been trained with large numbers of 
parameter which has shown great promise in the capabilities.
Prior to TarFlow~\cite{zhai2024normalizingflowscapablegenerative} and
StarFlow~\cite{gu2025starflowscalinglatentnormalizing},
DenseFlow~\cite{grcic2021denselyconnectednormalizingflows} and
MaCow~\cite{NEURIPS2019_20c86a62}, were the largest trained Normalizing Flows,
having 130M and 177M parameters, respectively.
Even TarFlow, having variants at ${\approx}475$M parameters and ${\approx}820$M
parameters, is much smaller than many diffusion models which have well over a
billion
parameters~\cite{kingma2023understandingdiffusionobjectiveselbo,crowson2024scalable,hoogeboom2023simplediffusionendtoenddiffusion,Zheng2024MaskDiT,Gao_2023_ICCV}
while StarFlow is, to the best of our knowledge, the only multi-billion
parameter Normalizing Flow.
Additionally, there has been recent success with Flow
Matching~\cite{dao2023flow,lipman2024flowmatchingguidecode,liu2022,chen2024flowmatchinggeneralgeometries,lipman2022}
has presented promising results in this area but are restricted to continuous
flows operating on conditional velocity fields. 
The principles should similarly apply but in this work we focus on more general
approaches.
With this in mind, we should expect some complications with unconditional
generation. 

\subsubsection{All Together}

We can combine all these distillations together to create a stronger and
unifying distillation method.
We may provide weights to each distillation type, using hyperparamter
$\lambda_i$.
Combining with our standard flow loss, we can write our final loss as:
\begin{align}\label{eq:nfd-loss}
    \begin{split}
        \mathcal{L}(t,s,x,z) 
            &= \lambda_0 \log(p_s(x)) \\
            &+ \lambda_1 \mathcal{L}_{LKD}(t,s,x)\\
            &+ \lambda_2 \mathcal{L}_{ILKD}(t,s,x)\\
            &+ \lambda_3 \mathcal{L}_{SKD}(t,s,z)
    \end{split}
\end{align}

We may compress this format by writing $\mathcal{L}_{(I)LKD}$ recognizing that
the LKD loss may be viewed as another step.
We write explicitly due to its importance as a boundary condition.

\section{Distillation Experiments}
Given these classes of knowledge transfer we can see that we can use both
directions of data processing with these networks to better align their mapping
trajectories.
Our goals are to determine the capabilities of these differing distillation
methods and better understand their strengths and weaknesses. 
Given this framework there is a large search space.
We do not intend to provide a complete search, but focus on demonstration and
forming the foundations.
For simplicity, we will use Lasso Regression for losses within this work,
$\mathcal{L} = L_1$.
For hardware, all experiments were performed using a single NVIDIA Tesla V100 
GPU.

To evidence our hypotheses and the utility of our framework we use two classes
of data.
Our first will focus on density estimation (\Cref{ch:dnf-densityestimation}) 
and then focus on synthetic image generation (\Cref{ch:dnf-imagegen}).
The latter of which is a significantly more challenging task for these
architectures.
For our models, we use the Masked Autoregressive Flow 
(MAF)~\cite{papamakarios2017masked} and Generative Flow with Invertible
$1\times1$ Convolutions (GLOW)~\cite{kingma_glow_2018}.
For our GLOW models, we use the affine coupling setting.
Unfortunately, the autoregressive nature of MAF makes sampling intractable, so
we do not perform SKD distillation with this mode.

These two models are commonly used and have been much more thoroughly studied.
These models also significantly differ in architectures, which will help us
determine the capabilities of these distillation methods.
For our GLOW models we perform our ILKD distillation between flow levels,
matching the diagram in \Cref{fig:distilling-flow}.
For MAF we match across depths, pushing every two depths from the teacher into
the student.
These are not necessary choices but we believe present natural points for
communication between teacher and student.

In all studies we seek to make large reductions in model parameters, as this
will best demonstrate our ability to compress knowledge into our student
networks.
Our MAF students are half the size of their teachers while we do not let our
GLOW student contain more than $30\%$ as many parameters as their teachers.

\subsection{Density Estimation}\label{ch:dnf-densityestimation}
\begin{wraptable}[12]{l}{0.45\textwidth}
    \begin{subtable}{\linewidth}
        \centering
        \begin{tabular}{lcc}
\toprule
\textbf{GLOW}
& Level (L)
& Hidden
\\
\midrule
Student
    & $3$   
    & $32$\\
Teacher
    & $3$
    & $64$\\
\midrule
\textbf{MAF} & Depth (K) & Hidden\\
\hline
Student 
    & $3$ 
    & $32$\\
Teacher 
    & $6$ 
    & $32$\\
\bottomrule
\end{tabular}

    \end{subtable}
        \caption[GLOW and MAF Model Configurations]{%
    Model configurations for generation of density estimation. Provided for 
    GLOW and MAF architectures. Number of levels (L) is equal to 1. Notation is 
    taken from the original paper \cite{kingma_glow_2018}. 
    }\label{tab:dnf-tabular-config}
\end{wraptable}

For our tabular data experiments we perform density estimation on five common
datasets. 
We use \emph{Metric}, \emph{POWER}, \emph{GAS}, \emph{HEPMASS}, and 
\emph{MINIBOONE} from the UCI Machine Learning Repository~\cite{UCI}
and BSDS300 from the Berkeley Segmentation Dataset and 
Benchmark~\cite{MartinFTM01}.

\noindent%
Our model configurations are presented in \Cref{tab:dnf-tabular-config}.
For our MAF model we focus on expanding the teacher's depth, letting the teacher
model have twice the depth.
For GLOW we let our models have the same number of levels and flow steps but
double the number of hidden neurons in the teacher. 
For MAF this results in the teacher having approximately double the number of
model parameters as the student.
For GLOW, this results in the teacher having approximately five times the number
of parameters as the student.
GLOW is a much more powerful model than MAF and thus we expect the ability to
greatly reduce model parameters.

\begin{table}[htpb]
    \centering
    \resizebox{1.00\linewidth}{!}{%
\begin{tabular}{llccccc}
  \hline
 Architecture
     & Model
     & POWER
     & GAS
     & HEPMASS
     & MINIBOONE
     & BSDS300
     \\
\midrule
 \multirow{5}{*}{GLOW} 
    & Student
        & $-0.228$
        & $5.967$
        & $-22.668$ 
        & $-17.251$
        & $147.298$  \\
    & LKD Student
        & $-0.132$    
        & $6.008$   
        & $-22.332$  
        & $-17.136$  
        & $162.103$  \\
    & ILKD Student
        & $-0.133$    
        & $6.191$    
        & $-22.187$  
        & $-17.008$  
        & $163.148$  \\
    & SKD Student
        & $\mathbf{-0.078}$    
        & $\mathbf{6.515}$ 
        & $\mathbf{-21.852}$  
        & $\mathbf{-16.130}$  
        & $\mathbf{163.953}$  \\
    \cline{2-7}
    & Teacher
         & \phantom{-}$0.143$    
         & $6.604$    
         & $-19.938$  
         & $-13.597$  
         & $165.702$  \\
\bottomrule
 \multirow{5}{*}{MAF} 
    & Student 
        & $-0.152$    
        & $4.385$    
        & $-21.904$  
        & $-15.314$  
        & $155.463$  \\
     & LKD Student 
        & $-0.149$    
        & $4.473$    
        & $-21.389$  
        & $-15.217$  
        & $155.629$  \\
     & ILKD Student 
        & $\mathbf{-0.145}$    
        & $\mathbf{4.502}$    
        & $\mathbf{-21.223}$  
        & $\mathbf{-15.184}$  
        & $\mathbf{155.785}$  \\
     & SKD Student 
        & \multicolumn{4}{c}{Intractable}\\
    \cline{2-7}
    & Teacher
    & $0.133$    
    & $5.887$    
    & $-20.662$  
    & $-13.488$  
    & $159.442$  \\
\bottomrule
\end{tabular}


    }
    \caption[Distilling Normalizing Flow Density Estimation Metrics]{%
    Averaged test log-likelihood (in nats) for unconditional density
    estimation (higher is better) across multiple runs.
    }\label{tab:dnf-tabular-metrics}
\end{table}

For each model we train the flow for a fixed $10^4$ iterations with a batch size
of $65,536$ ($2^{16}$).
We use a learning rate of $5\times10^{-5}$, applied to the AdamW
optimizer~\cite{kingma2017adam,loshchilov2019decoupledweightdecayregularization}.
The results of these runs can be found in \Cref{tab:dnf-tabular-metrics}.
Rows labeled ``Student'' and ``Teacher'' contain no distillation and are the
baseline values of or models.
For LKD and ILKD we set $\lambda_0=0.9$ and $\lambda_1=\lambda_2=0.1$.
For SKD we again decrease $\lambda_0=0.85$ and set the rest to $0.075$.
These weights were chosen to set control the weights as percentages of the whole
loss, and are likely non-optimal.
We find that most performance comes from the introduction of the LKD student, 
with $+10\%$ for GLOW but only $+1\%$ for MAF on BSDS300.
We get continued improvements with ILKD, $+6\%$ for GLOW and $+1\%$ for MAF.
With GLOW we can cleanly sample from our distribution and find an additional
$+5\%$ gain, for a total improvement of $+12.7\%$ improvement above our student
model.

On BSDS300, our final student GLOW model is has ${\approx}25.5$ as many 
parameters as the teacher model while achieving $98.94\%$ the accuracy.
We also compare the computational performances differences of our teacher and
student models in \Cref{tab:dnf-tabular-performance}, directly comparing the
number of model parameters and their throughput.
This result suggest strong distillation capabilities, with the teacher passing
nearly all its ``knowledge'' to its student.
We do not believe our parameters are near optimal, but this provides significant
evidence to our theory that suggests it may be possible to fully distill the
teacher's knowledge into the student, under the assumption that the student's
latent representation is at least as large as the latent data manifold.

\begin{table}[htpb]
    \centering
    \resizebox{1.00\linewidth}{!}{%
        \begin{tabular}{llcccccc}
\toprule
Arch
    & Model
    & Metric
    & POWER
    & GAS
    & HEPMASS
    & MINIBOONE
    & BSDS300
    \\
\midrule
\multirow{4}{*}{GLOW} & \multirow{2}{*}{Student} & Time (ms)
    & $2.32 \pm 0.16$    
    & $2.46 \pm 0.10$   
    & $2.55 \pm 0.35$  
    & $2.47 \pm 0.07$  
    & $2.45 \pm 0.07$  \\
&  & Params (K)
    & $13.8$    
    & $14.20$   
    & $17.4$  
    & $24.9$  
    & $34.4$  \\
\cline{2-8}
& \multirow{2}{*}{Teacher} & Time (ms)
    & $3.65 \pm 0.26$    
    & $3.88 \pm 0.09$    
    & $4.41 \pm 0.28$  
    & $3.95 \pm 0.11$  
    & $3.89 \pm 0.14$  \\
&  & Params (K)
    & $\phantom{0}86.7$    
    & $\phantom{0}87.8$    
    & $\phantom{0}96.3$  
    & $114.2$  
    & $134.7$  \\
\midrule

\multirow{4}{*}{MAF} & \multirow{2}{*}{Student} & Time (ms)
    & $2.00 \pm 0.21$    
    & $1.98 \pm 0.19$    
    & $1.82 \pm 0.05$  
    & $1.82 \pm 0.05$  
    & $1.91 \pm 0.22$  \\
&  & Params (K)
    & $\phantom{0}5.0$    
    & $\phantom{0}5.6$    
    & $\phantom{0}9.4$  
    & $15.9$  
    & $21.8$  \\
\cline{2-8}
& \multirow{2}{*}{Teacher} & Time (ms)
    & $3.34 \pm 0.22$    
    & $3.22 \pm 0.18$    
    & $3.34 \pm 0.23$  
    & $3.36 \pm 0.26$  
    & $3.45 \pm 0.20$  \\
&  & Params (K)
    & $10.1$    
    & $11.2$    
    & $18.9$  
    & $31.8$  
    & $43.6$  \\
\bottomrule
\end{tabular}

    }
    \caption[Distilling Normalizing Flow Performance Metrics]{%
    Time consumption for a single batch inference averaged across multiple
    batches and the number of parameters (in thousands). Average time (ms)
    and number of parameters (in thousands) are reported.
    }\label{tab:dnf-tabular-performance}
\end{table}

\begin{table}[hbtp]
    \begin{subtable}{0.51\linewidth}
        \centering
\begin{tabular}{lccc}
\toprule
\multicolumn{4}{c}{CIFAR-10}\\
\toprule
& Levels (L)
& Hidden
& Params\\
\hline
Student
    & $8$
    & $512$
    & $11.0$M  \\
Teacher
    & $32$
    & $512$
    & $44.2$M \\
\bottomrule
\end{tabular}

    \end{subtable}
    \begin{subtable}{0.47\linewidth}
        \centering
\begin{tabular}{ccc}
\toprule
\multicolumn{3}{c}{CelebA}\\
\toprule
Levels (L)
& Hidden
& Params\\
%
%
\hline
    $16$
    & $256$
    & $\phantom{0}7.9$M  \\
    $32$
    & $512$
    & $61.2$M  \\
\bottomrule
\end{tabular}

    \end{subtable}
    \caption[Distilling Normalizing Flow Model Configurations for Image Generation]{%
    Model configurations image generation tasks (GLOW). 
    Notation is taken from the original paper \cite{kingma_glow_2018}.
    All models have a depth (K) of 3.
    }\label{tab:dnf-configs}
\end{table}

\subsection{Image Generation}\label{ch:dnf-imagegen}
To demonstrate capacity in image synthesis we demonstrate our methods on using
the CelebA~\cite{liu2015faceattributes} and CIFAR-10~\cite{4531741} datasets.
Due to MAF's low performance we only perform these experiments using the GLOW
model.
We use the same settings as before, except reduce the batch size to 32.
Model configurations and sizes can be found in \Cref{tab:dnf-configs}. 
All models have a constant depth of 3, like before.
In the CIFAR-10 student we only quarter the number of levels but in the CelebA
student we halve the levels and halve the number of hidden parameters.

In our experiments, we found that SKD appeared to be helping with model
distillation but that the results were unstable, and we were unable to complete
training.
We suspect that this was due to the poor image generation quality, as can be
seen in \Cref{fig:dnf-celeba,fig:dnf-cifar}.
For CIFAR-10 our teacher model obtained a Bits per Dimension (bpd) of 3.423,
which is slightly worse than Kingma and Dhariwal's work, while our CelebA model
reached 2.474 bpd.

The results of our experiments can be found in \Cref{tab:dnf-image-metrics}.
In terms of bpd, the ILKD student showed an improvement of $0.5\%$, but note
that the teacher is only $2\%$ better.
On CelebA we see a similar $0.16\%$ improvement, while the teacher is only
$0.2\%$ better.
When looking in terms of FID we instead see a $2.5\%$ (of $3.8\%$) on CIFAR-10
and $20\%$ (of $45\%$).
We note that our CIFAR-10 student is ${\approx}25\%$ the size of its teacher and
the CelebA is ${\approx}13\%$.

\begin{table}[htpb]
    \centering
    \begin{tabular}{lcccc}
\toprule
    & \multicolumn{2}{c}{CIFAR-10}
    & \multicolumn{2}{c}{CelebA} \\
\cline{2-5}
    & bpd
    & \multicolumn{1}{c}{FID}
    & bpd
    & FID \\
\midrule
Student 
    & $3.498$ 
    & \multicolumn{1}{c}{$71.177$}  
    & $2.479$  
    & $68.127$  \\
ILKD Student 
    & \textbf{3.481}  
    & \multicolumn{1}{c}{\textbf{69.371}}  
    & \textbf{2.475}  
    & \textbf{54.480}  \\
\midrule
Teacher 
    & $3.423$  
    & \multicolumn{1}{c}{$68.503$}  
    & $2.474$  
    & $37.460$  \\
\bottomrule
\end{tabular}

    \caption[Distilling Normalizing Flow Image Generation Metrics]{%
        Metrics for the image generation task for the GLOW architecture using 
        ILKD on the test set: bits per dimension and FID (lower is better).
    }\label{tab:dnf-image-metrics}
\end{table}

To ensure the knowledge distillation does not corrupt the hidden space, we need
to ensure that random samples from the students still maintain similar quality
images.
With high dimensional information, it is possible for Normalizing Flows,
and other models, to have a small KL-Divergence but also have poor sampling
quality. 
This appears to be the case in our results, where our teacher and student have
similar bpds but very different FIDs.
Similar to \Cref{ch:stylenat-amaps}, we need to be careful in how we analyze our
results given the biases of our metrics.

\begin{figure}[htpb]
    \centering
    \input{Includes/Figures/DNF/cifar-fig.tex}
    \caption[Distilling Normalizing Flow CIFAR-10 Samples]{%
    CIFAR-10 samples from teacher model (\cref{fig:dnf-cifar-Teacher}), student
    model (\cref{fig:dnf-cifar-wKD}), and student model with no knowledge
    distillation (\cref{fig:dnf-cifar-woKD}). All images are generated at
    $32\times32$ resolution and with a temperature of 0.7.
    }\label{fig:dnf-cifar}
\end{figure}

Considering the biases of FID, high quality samples can only happen if there is a
sufficiently good enough cover, determined by the Inception Network, within 
the learned latent space. 
Thus, we propose to measure the quality of the inferred samples for randomly 
chosen images
$\boldsymbol{u}, \boldsymbol{v}$ and an $\alpha \in [0,1]$, where $\alpha$ is
the interpolation fraction.
The preserved norm of the latent vector can be defined as:

\begin{equation}
    \boldsymbol{f(u, v, \alpha)} = 
        ((1 - \alpha) \boldsymbol{f(u)} 
            + \alpha \boldsymbol{f(v)}) 
            \cdot \frac{(1 - \alpha) ||\boldsymbol{f(u)}|| 
                + \alpha ||\boldsymbol{f(v)}||}
                {|| (1 - \alpha) \boldsymbol{f(u)} 
                  + \alpha \boldsymbol{f(v)} ||}
\label{eq:norm_interpolation}
\end{equation}

The results of this method are provided for CelebA dataset in
\Cref{tab:dnf-interpolation-metrics}.
This table shows that the ILKD Student performs significantly better than the 
student without knowledge distillation, independent of temperature.
Both the student and ILKD student show substantially larger improvements in FID
with reduced temperatures showing their improperly configured latent
representations.
In both cases, the ILKD student shows a $30\%$ improvement over the student.

\begin{table}[htpb]
    \centering
\begin{tabular}{lcc}
  \toprule
     & $T=1.0$
     & $T=0.7$
 \\
 \midrule
 Student 
    & $40.159$
    & $28.432$  \\
 ILKD Student 
    & \textbf{28.413}  
    & \textbf{19.688}  \\
 \midrule
 Teacher 
    & $19.062$  
    & $16.382$  \\
 \bottomrule
\end{tabular}

    \caption[Distilling Normalizing Flow CelebA FID]{%
    CelebA FID values of images obtained by interpolation in the latent space 
    of trained models.
    }\label{tab:dnf-interpolation-metrics}
\end{table}

While the CIFAR-10 (\Cref{fig:dnf-cifar}) samples are more difficult to 
differentiate, it is clear that in the CelebA generation (\Cref{fig:dnf-celeba}) 
that our distilled images are significantly better than those in the original
student.

\begin{figure}[htpb]
    \centering
    \input{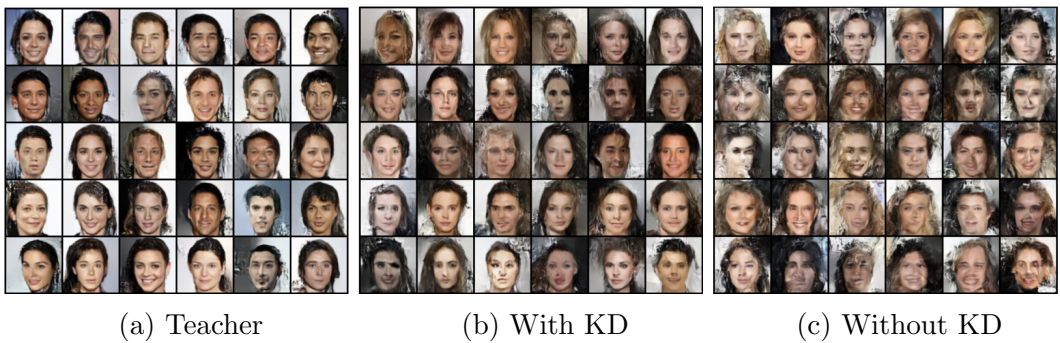}
    \caption[Distilling Normalizing Flow CelebA Samples]{%
    CelebA samples from teacher model (\cref{fig:dnf-celeba-teacher}), student
    model (\cref{fig:dnf-celeba-withkd}), and student model with no knowledge
    distillation (\cref{fig:dnf-celeba-nokd}). All images are generated at
    64$\times$64 resolution and with temperature=$0.7$.
    }\label{fig:dnf-celeba}
\end{figure}

\section{Conclusion}
Our work we sought to build foundations for investigating the capabilities of
knowledge distillation in Normalizing Flows.
Our work is not intended to be comprehensive, but to demonstrate how effective
these methods are and motivate further study.
These architectures are underrepresented, but the theory and practice shows that
these models may offer unique capabilities to the field of machine learning.
The success of large Normalizing Flows like
TarFlow~\cite{zhai2024normalizingflowscapablegenerative,liu2025acceleratetarflowsamplinggsjacobi}
and StarFlow~\cite{gu2025starflowscalinglatentnormalizing} show promise in their
capabilities, and that similar scaling success may be found here too.

With a large variety of flow types~\cite{walton2023isomorphism}, there are many
avenues open to build upon this work.
With many flow architectures often presenting computational challenges, this may
provide an avenue to resolve some of them.
We believe our foundation will extend ensembles of models, letting students
benefit from the training advantages of differing flow types.
We also believe that our process will extend to distillation between differing
architectures.
This may offer unique capabilities, such as replacing computationally difficult
flow steps with simpler ones.
While these models are often overlooked due to their mathematical formulations,
we believe that continued study will show these models to be highly capable and
researchers will highly benefit through their greater interpretability.

\chapter{Conclusion and Future Directions}\label{ch:conclusion}
\chapterquote{
People think of education as something they can finish. 
}{Isaac Asimov}

\section{Summary}
This work explored the importance of neural architectures and how they
influence not only a machine learning model's capacity to learn, but also how to
do so in computationally constrained environments.
Even as compute infrastructure grows, there exists strong pressures to use
what we have more efficiently.
If we are able to do so, then our progress can outpace our growth in compute.

\Cref{ch:background} gave an overview of the subject matter necessary
understand to influence neural designs.
This serves not to only help the reader understand the problems that need be
addressed, as well as illustrates the many pitfalls and subtleties that
exist. 
The remains many challenges when scaling models (\cref{sec:bg-SINAYN}) and
data (\cref{sec:bg-scale-data}), highlighting the importance of algorithmic or
architectural improvements (\cref{sec:bg-architectures}).
\Cref{sec:bg-measures} also discussed the difficulties when defining 
objectives and developing adequate measures, known as ``The Alignment
Problem,'' and the critical relationship which influences neural designs.

\Cref{ch:escaping} focuses on the importance of data encoding and decoding.
Specifically in how to improve these designs for Vision Transformers, allowing
them to better automate discovery of underlying data structures.
Our work demonstrates that without efficient encoding and decoding, we may
inadvertently hinder the performance of these models.
These inefficiencies, ideally, may be overcome through brute force scaling, but
through careful design we may reduce our costs, allowing us to do more with
less.

\Cref{ch:stylenat} shifts focus to modifying the core architectural designs.
Through understanding the ways in which architectures operate and how
leveraging structures within the data allows for more informative decisions in
the design of core processing units.
Neighborhood Attention modifies the Vision Transformer architecture to increase
computational and memory performances, leveraging the natural localization
biases of the data while still being able to recover global structures.
Through our improved design, allowing attention heads to operate over
independent receptive fields we are able to reduce the sacrifices made and our
models can uncover structures they previously could not.

\Cref{ch:flows} focuses on architectures with structurally focused designs.
This studies the way in which information is processed through Normalizing Flows
and how this can be used to create efficient knowledge distillation.
Understanding the mathematical structures within architectures allows for
better design and efficient methods which reduce model size and required
computation.

Putting this all together, this dissertation positively answers the
question: 
\emph{Can we design neural architectures to be smaller, faster, and cheaper 
without sacrificing performance?}

\section{Future Directions}
While this dissertation affirms that our models \emph{can} be more efficient, 
we are unable to provide a complete answer as to \emph{how}.
Despite the significant strides the field has made in recent decades, we
are only at the beginning.
With the rapid development of machine learning, it is easy to lose sight of the
larger goals. 
Therefore, we briefly discuss our core goals to ensure our future work
remains aligned.

\subsection{Core Challenges}
A core challenge still must be solved in order to efficiently design our 
neural architectures.
Largely, machine learning deals with the problem of alignment.
Without a strong mathematical foundation we are unable to verify how well our
models are aligned to our intended goals.
To draw an analogy, in \Cref{ch:stylenat} we discussed the limitations in our 
ability to determine the realism of the generated imagery.
This hinges on our inability to describe the ``realness'' of our images in a
rigorous way.
Our Generative Adversarial Network cleverly trains an adversarial detector as a
means to bypass this formalism.
While this proxy has allowed us to dramatically improve the quality of the
generated imagery, it is not uncommon for the generator to become misaligned.
Instead of generating high quality imagery, it may instead produce incoherent
images that have the right statistics to deceive the generator.
Without formalism we must take great care to ensure that we do not fall for the
same trap as the detector.
Our metrics play a critical role in driving our research and designs, but
we must not blindly follow a map that may drive us off a cliff instead of to our
intended destination.
Until such rigorous formalism is developed we must remain skeptical of
ourselves. 
We cannot forget this fundamental problem while addressing more specific
challenges.

Some practical advice may be offered by Donald Knuth:
\emph{If you find that you're spending almost all your time on theory, start 
turning some attention to practical things; it will improve your theories. If 
you find that you're spending almost all your time on practice, start turning 
some attention to theoretical things; it will improve your practice.}

\subsection{Scaling}
The subject of this thesis would be incomplete without revisiting the issue of
scaling.
The question remains: \emph{Do these methods work as data increases and as model
size increases?}
This is, after all, the fundamental question addressed in
\Cref{ch:introduction}.
This research has been carefully designed to ensure that the answer will be
\emph{yes}.
While we cannot confirm this conclusion without access to large compute
infrastructures to verify these beliefs, we have strong evidence for this
belief.

In \Cref{ch:escaping} we made only minor changes to the network, so this should
be surprising if it were to make the much larger networks substantially
unstable.
These ViT networks have shown success at scale and we believe our modifications
should make difference, with respect to scalability.
In this work, we showed that CCT has strong performance across small to medium
scales, \emph{strictly dominating} ViT at every step of the way.
The saliency maps suggest we are removing a fundamental flaw found in ViTs,
which should only lead to greater stability.

In \Cref{ch:stylenat} we similarly suspect that these methods are highly
scalable.
We trained our single $1024\times1024$ run in an effort to show this, and
\Cref{fig:ffhq-fiditer} suggests that neither our small resolution nor our
large resolution training achieved peak performance.
The design of the Hydra-Neighborhood attention specifically allows more
configurations, and thus adaptability.
The work of StyleGAN-XL mostly saw success to its own scaling, and we suspect
the same here.
While the LSUN church results did not perform as well, all results suggest that
this is likely an embedding problem, with the attention dimension being too
small.
If this is a correct assumption, then scale should yield substantial
improvements to this experiment.

In \Cref{ch:flows} we again see no blockers.
The theory behind the methods suggest that training is better performed with
large scale models but that these can also be reduced in size through
distillation. 
Here, the question is not a matter of if the procedure can continue to scale,
but by how much can we compress these large flows.
With works like TarFlow and StarFlow resenting huge state of the art models,
this only makes our methods more valuable.

At this time we do not have the computational resources to prove that these
methods are scalable, but there is nothing that suggests that they won't be just
as effective, if not more, at scale.

\subsection{Ingress and Egress of Data}
In \Cref{ch:escaping} our work focused on making the most of our data, allowing
our core architecture to make better utilize available data.
It is key that we provide our models with data in the formats that best suit
them.
Similarly, we must be careful in how we extract the data from them, ensuring we
do not lose useful relationships they have uncovered.

\subsubsection{Parameterization}
Our Compact Transformers improved upon the patching and embedding method of the
original Vision Transformer by recognizing how non-overlapping patches removed
structure from the data.
By using small kernels and overlapping patches our embedding is able to better
preserve the structure within our data.
The size of these kernels, strides, and other parameters were determined through
directed search optimizing a validation set.
These specific relationships may not hold for data that has other inherent
biases, and this process may need be done again.
Similarly, there is no reason to believe that those we found are optimal.
By revisiting former models and architectures many researchers have demonstrated
that their performances can be improved in many
ways~\cite{wightman2021resnetstrikesbackimproved,liu2022convnet,Woo_2023_CVPR}.
The method itself does not prevent these hyper-parameters from being learned.
More optimal parameters may be found through HyperNets~\cite{6792316}, or
other optimization methods.

\subsubsection{Automated Preprocessing}
The CNN based structure itself causes some of the image structure to be lost and
new methods should be investigated which can better embed these.
Most importantly, the CNN places greater importance on pixels that are local
spatially.
While we expect this relationship to be strong it may not always be true, nor
should we assume that in some cases we may wish to place greater importance on
more global features.
This is, after all, the same reasoning that led to the development of the 
transformer architecture.
Given these problems attention needs to be given to develop ingestion methods
that flexibly adapt to the data.
Modern machine learning methods are becoming multi-modal, processing language,
vision, and other manners of data.
This necessitates new forms of embedding that can recognize and adapt to the
data, performing the preprocessing for us.

Another benefit of the CNN architecture is that it is flexible to the data
shape.
CNNs mainly rely on a single dimension of our data, channels, allowing us to
more easily accommodate images of varying dimensions.
As we seek to make our models multi-modal this demands that we develop
architectures to ingest data of differing types and dimensionality.
We may inefficiently provide patchwork by padding or replicating data, but these
may provide more hindrances than utility.
It becomes important to investigate means of arbitrary data ingestion, that can
embed our data without the loss of structures within the data.

\subsubsection{Making The Most of it}
On the other side our SeqPool method demonstrated that our network had learned
useful relationships that could help in image classification but were
unavailable to our classifier sub-network.
This demonstrates the ease in which we may underutilize our networks.
Certainly our SeqPool method has not extracted all available information from
our network.
The method allows for the importance of each of our tokens but asks that a lot
be done within this simple step.
This constrains our core network which almost certainly adapts to this
bottleneck.
Other extraction methods should be explored which allow greater flexibility.
While the classification network can, in some ways, act similar to the linear
layers of the transformer, they do not have the same expansion layers that allow
data untangling as the traditional transformer.
Finally, while the classification sub-network remains the de-facto solution for
converting our multi-dimensional relationships into linear ones, there likely
exists better methods for this.
While being seemingly less exciting, finding such architecture may lead to 
transformative impacts in the field.

Further improvement needs to be made to disentangle data.
We use batching and pooling within our network and while this can speed up
and even benefit training it can also entangle data, causing our networks to
over aggregate.
Our work used a learnable class token to constrain our network to disentangle
these data, but further study to better understand and improve this
disentanglement remains.

\subsection{Core Processing Architectures}
Our work in \Cref{ch:stylenat} demonstrated that even seemingly simple changes
to the core neural architectures can have tremendous impacts on our performance.
An idea that was seemingly simple, yet non-obvious due to the full attention
mechanism naturally having the ability do this and other restricted
attention methods, not necessarily sharing the flexibility of Neighborhood
Attention.
Our work placed greater focus on demonstrating our hypothesis around integrating
local and global features than on ensuring the method was computationally 
efficient.
While significant advances have been made to the \natten kernel 
\cite{hassani2024fasterneighborhoodattentionreducing,hassani2025generalizedneighborhoodattentionmultidimensional}, 
these improvements have not been made to optimize the independent head approach.
The algorithm developed for the paper was not intended to be optimized for speed
or memory and instead for better readability. 
Simple modifications can be made to better parallelize the data processing and
further improve performance.

\subsubsection{Flexible Learning}
The number of configurations available made our method flexible, but it was not
possible to exhaustively test these within our limited computational budget.
Further work should be performed to tune these parameters.
Initial testing has shown that we are able to modify these kernel and dilation
parameters during the training of our networks.
This suggests that these parameters may be optimized during training, allowing
them to be adapted to the data being modeled.
This may resolve the need to manually craft these hyper-parameters but research
is still needed to determine the stability and effectiveness of this process.

The method has also not been tested on other architectures, such as diffusion
model.
Our focus was on demonstrating the utility of the method rather than developing
the best image generator.
Works like Hourglass Diffusion~\cite{crowson2024scalable} have demonstrated
substantial improvements due to integrating Neighborhood Attention but remain
outside our computational budgets to integrate Hydra-NA.
We believe that our results will apply much more broadly than just to GANs, but
this has yet to be demonstrated.

\subsubsection{Is Beauty in the Eye of the Beholder?}
The work also highlighted the challenges of metrics and alignment.
Our previous CCT model focused on classification tasks, where there exists a
much clearer objective and measure.
In classification, the labels are not always accurate as some noise exists 
around identification, but there is at least a clear metric to define if
the model produced some expected output or not.\footnote{We should still take
care to recognize that even a classification metric is not foolproof.}
In the case of generative imagery, no such metric exists and even may not exist.
For millennial scholars have attempted to create formal definitions of beauty,
but have yet to find success.
Therefore, those studying generative modeling must then take great care 
concerning the alignment problem and we must be creative in determining how we 
may achieve better measures and importantly, optimization methods.
Na\"{i}vely, we may unintentionally develop models which can only produce content
that is appealing to limited groups.
This requires us to be suspicious of ourselves, recognizing our own cognitive
biases.
Works like Stein \etal~\cite{stein2023exposing} help by using large human
studies, but even these studies have population biases.

While diffusion models optimize towards a probability distribution function they
often still make certain assumptions about the data, such as being \emph{i.i.d}.
This method limits our generation to only match distributions similar to those
we train on but so far have been unable to demonstrate the ability to capture
the depth and nuance that art is known for.
Without incorporating seemingly small details our generators may be unable to
escape an uncanny valley which is blatancy apparent to some but invisible to
others.
Through studying other forms of measure and optimization we most certainly will
find better neural architectures and make steps to reducing these discrepancies
in human preference.

\subsection{Structurally Aware Architectures}
In \Cref{ch:flows} we moved from transformer architectures to study Normalizing
Flow architectures.
These have seen significant progress, especially with recent works in Flow
Matching~\cite{lipman2024flowmatchingguidecode}.
While our work may not directly apply to these methods due to their continuous
nature, we show how our compositional flows layers can be compressed.
This presents the question to determine if these compositional flows can be
efficiently transformed into continuous ones.
This may allow for more flexible relationships to be found or by turning our
continuous flows into compositional ones we may gain more interpretability.
Understanding how our models operates remains a core challenge, relating to our
issues of alignment. 

For our specific work there is quite a lot of potential future work we see here.
None of our models were optimized and we were focusing on demonstration due to
computational limitations.
The SKD distillation should promise in the density estimation experiments but
was unstable in the more difficult image generation tasks.
As flow architectures become more powerful, this too is likely to increase in
capacity.
One may also wish to add schedulers to the weights of our distillation methods,
and particularly with SKD.

More broadly, with the larger discussion of this thesis, the work demonstrates
the importance of understanding the structures our neural architectures do and
do not preserve.
Normalizing Flows provide nice mathematical structures that are easier to study
than other neural architectures, but this only highlights the need to better
study the limits and capabilities of other architectures, which are
significantly more difficult than Normalizing Flows.

\section{Conclusion}
This thesis presents work that demonstrates the potential for making machine
learning models more efficient through careful design of our neural
architectures, specifically applied to Computer Vision problems.
We show that we can train smaller models, from scratch, while greatly reducing 
compute, memory, and the costs of gathering and labeling data.
We also show that seemingly trivial modifications may be made that have
significant impacts on performance.
While many of these changes may seem obvious post-hoc our work only highlights 
how hidden such simple modifications are.
Like our education, the pursuit of more efficient models is something that can
never be finished, and many grand challenges still lay waiting to be discovered.

    }{
}

\IfFileExists{main.bib}{%
    \small
    \bibliographystyle{ieeenat_fullname}
    \bibliography{main}
    }{%
}

\end{document}